# LEARNING TO SEE ANALOGIES: A CONNECTIONIST EXPLORATION

**Douglas S. Blank**

Submitted to the faculty of the Graduate School
in partial fulfillment of the requirements
for the joint degree of Doctor of Philosophy
in the Departments of Computer Science
and
Cognitive Science
Indiana University

December 1997

Accepted by the Graduate Faculty, Indiana University, in partial fulfillment of the requirements of the degree of Doctor of Philosophy.

Doctoral
Committee

______________________________
Michael Gasser, Ph.D.
(Principal Advisor)

______________________________
Robert Goldstone, Ph.D.

______________________________
Robert F. Port, Ph.D.

______________________________
Dirk Van Gucht, Ph.D.

Bloomington, Indiana
December 1997.

For Laura and Thaddeus.

# Abstract

This dissertation explores the integration of learning and analogy-making through the development of a computer program, called Analogator, that learns to make analogies by example. By "seeing" many different analogy problems, along with possible solutions, Analogator gradually develops an ability to make new analogies. That is, it learns to make analogies by analogy. This approach stands in contrast to most existing research on analogy-making, in which typically the *a priori* existence of analogical mechanisms within a model is assumed.

The present research extends standard connectionist methodologies by developing a specialized associative training procedure for a recurrent network architecture. The network is trained to divide input scenes (or situations) into appropriate *figure* and *ground* components. Seeing one scene in terms of a particular figure and ground provides the context for seeing another in an analogous fashion. After training, the model is able to make new analogies between novel situations.

Analogator has much in common with lower-level perceptual models of categorization and recognition; it thus serves as a unifying framework encompassing both high-level analogical learning and low-level perception. This approach is compared and contrasted with other computational models of analogy-making. The model's training and generalization performance is examined, and limitations are discussed.

# Acknowledgments

I would like to thank first my advisor, Mike Gasser. In 1990 I came to him with barely an idea and convinced him that it was worth pursuing. We began to meet regularly to discuss the issues surrounding this basic idea. However, after a couple of years, I seemed no closer to a thesis than on the first day. It must have seemed like I was going in circles, yet he patiently listened and suggested. Only years later did I realize that I wasn't going in circles, but rather was actually traveling a slow spiral gravitating toward the thesis described in this dissertation. I am very happy with the way that the project turned out; if the path that I took were the only way to get here then I would surely do it again. For this I owe Mike.

The rest of my committee has, over the years, also been of great inspiration to me. Bob Port, one of the founding fathers of cognitive science at Indiana University, introduced me to many of the topics that have affected my view of minds. Dirk van Gucht was my first instructor as I entered grad school and has done an excellent job of showing me what it is like to be a first-class researcher and teacher. Rob Goldstone has help show me what it is like to be an excellent cognitive scientist.

During my stay in the computer science department and cognitive science program, I have had the opportunity of working with many teachers and researchers that have inspired me. They were David Leake, Ming Kao, Gregory Rawlins, John Kruschke, Jonathan Mills, Lorilee Sadler, and Suzanne Menzel. Doug Hofstadter has, in many ways, paved the road for this dissertation; to say that my life would be very different if I hadn’t met him would be a very large understatement.

Before life in grad school, there were many people that helped shape my young mind. They were Nick Toth, Bob Meier, and Bonnie Kendall, all inspiring people in the Department of Anthropology at Indiana.

The people in the AI Lab (a.k.a., CRANIUM) were always (and I do mean *always*) there for interesting conversation and assistance. They were Sven Anderson, Fred Cummins, Doug Eck, Susan Fox, Paul Kienzle, Andy Kinley, Devin McAuley, John Nienart, Pantelis Papadapolous, Cathy Rogers, Raja Sooriamurthi, Jung Suh, Keiichi Tajima, and Dave Wilson.

My friends were always supportive – either with words, a frisbee, a beer, or a guitar. They were Eric Jeschke, Amy Baum, Judy Augsburger, Ethan Goffman, John Blair, Michael Chui, Charles Daffinger, Liane Gabora, Eric Wernert, Bill Dueber, Brian Ridgely, Terry Jones, and Rupa Das. Some other friends had to work overtime to help me keep my sanity. They were Amy Barley, Dave Chalmers, Jim Marshall, Gary McGraw, Lisa Meeden, Naz Miller, Chris-man Pickenpaugh, Kirsten Pickenpaugh, and Jon Rossie.

Many administrative friends helped me from the computer science and cognitive science staffs. They were Ann O. Kamman, Grace Armour, Julia Fisher, Pam Larson, Karen Laughlin, and Linda McCloskey. The Indiana University Department of Computer Science has one of the best systems staffs that one could imagine. I would especially like to thank Rob Henderson, Steve Kinzler, and Dave “dxp” Plaisier.

Since I left Bloomington, an entirely different computer science department has helped me; everyone at the University of Arkansas has been very supportive. I would like to especially thank Oliver Ross, Dennis Brewer, Antonio Badia, Hal Berghel, and George Holmes.

Over the years, my family has been far too supportive; if this had taken another decade or two they would have still cheered for me all the while. They are Scott Blank, Norma Blank, David Blank, Julia Blank, Laura Blank, and the rest of the Blanks, Blankenships, Baxters, and Bamfords.

Also, I would like to thank Yoshiro Miyata and Andreas Stolcke for their work on the clustering software that I used extensively in my analysis.

Finally, and most of all, I would like to thank my wife, Laura Blankenship, and my son, Thaddeus, for their love, support, and patience.

# Contents

# List of Figures

# List of Tables

# 1 Introduction

*The scientific mind does not so much provide the right answers as ask the right questions.*

-Claude Levi-Strauss

The goal of this dissertation is to integrate learning and analogy-making. Although learning and analogy-making both have long histories as active areas of research in cognitive science, not enough attention has been given to the ways in which they may interact. To that end, this project focuses on developing a computer program, called Analogator, that learns to make analogies by seeing examples of many different analogy problems and their solutions. That is, *it learns to make analogies by analogy*. This approach stands in contrast to most existing computational models of analogy in which particular analogical mechanisms are assumed *a priori* to exist. Rather than assuming certain principles about analogy-making mechanisms, the goal of the Analogator project is to learn what it *means* to make an analogy. This unique notion is the focus of this dissertation.

In the following section, I discuss the assumptions that I have made – and not made – in the Analogator project.

## 1.1 Traditional Assumptions

Although the history of modeling analogy-making is only a few decades old, some well-entrenched traditions have developed. The traditional view of modeling analogy-making appeals to the intuition. A common assumption is that analogy-making begins with two structures – one representing a well-known *source* situation, and the other a lesser known *target* situation (see, for example, Gentner, 1983; Holyoak and Thagard, 1989a). These two structures are imagined to be similar to Minsky's frames (1975), or Schank and Abelson's scripts (1977). Analogy-making is assumed to be a search through the structures, matching analogous components. Data structures are assumed to be composed of structured parts, namely *objects*, *attributes*, and *relations*. There are assumptions about what types of parts can match, which parts to consider first, and which parts should be weighted most significant. An analogy is considered to have been made when a complete set of "mappings" between source and target components has been found.

Much research into analogy-making has focused on creating efficient searches through the data structures. Because these data structures are seen as being a core part of analogy-making, and analogy-making is viewed as a search through them, analogy-making (like much of high-level cognition) has been considered better handled by traditional symbolic artificial intelligence (AI) techniques. Briefly, here are the major assumptions made by most of the traditional models of analogy-making:

1. *Analogy-making begins with two structures.*

2. *Analogy-making is a search through the structures in an attempt to find analogous parts.*

3. *Syntax alone determines the similarity between any two objects, attributes, or relations.* For example, `LARGER-THAN` and `BIGGER-THAN` are not seen as any more similar than `BLACKER-THAN` and `WHITER-THAN`.

4. *For any two relations to be seen as analogous, they must exactly match in terms of their number of arguments, and types of arguments*. For example, `LARGER-THAN(radius, circle-1, circle-2)` would be seen as having nothing in common with `LARGER-THAN(circle-1, diameter, circle-2, meters)` even though they express similar relations.

5. *Relations, attributes, and objects are distinctly different things*. Because of this assumption a relation, such as `CIRCLE-OF(obj1, obj2, obj3, obj4, obj5)`, cannot be seen as analogous to a circle-shaped object.

6. *Context plays no part in making the analogy*. This is a common simplifying assumption.

7. *The result of making an analogy is the creation of a mapping between corresponding pieces of the two structures*.

We will examine these assumptions in detail in Chapter 6. We now turn to examine the assumptions made in the Analogator project.

## 1.2 Analogator's Assumptions

My goal was to construct a computational model with few assumptions so that analogy-making could "fall out" as naturally as possible. Although the Analogator model is not an implementation of assumed analogical mechanisms *per se*, it is not without assumptions. Analogator's assumptions are few and quite general. They are:

1. *Analogy-making should be the natural by-product of perceiving the similarity between two things.*

Some researchers have postulated that analogy-making is a special mode that we enter into when our reasoning system hits an impasse (see, for instance, Burstein, 1988). Yet, analogy-making has been shown to occur spontaneously (French, 1992). In addition, it has also been argued that analogy-making is a perceptual process (Mitchell, 1993). Therefore, it is a simpler explanation to posit analogy-making as a process of perception that is constantly in operation rather than appealing to more complex theories.[1] This assumption places analogy-making deep into our cognition, down at the core of our intelligence.

2. *Analogy-making should be linked to lower-level processes*.

Mitchell (1993) has shown that analogy-making is related to categorization and recognition. Likewise, I believe that a computational model of analogy-making should be connected to those same lower-level processes. A unification of low-level and high-level processes is a highly desirable trait of any cognitive model.

[1] Hofstadter (1995) makes a similar point.

3. *Perception should be seen as naturally breaking things into the* figure *and the* ground.

Things that we perceive, such as scenes, sentences, and sounds, fall into two pieces: the "figure" and the "ground" (see early gestalt psychologists Rubin, 1915, and Wertheimer, 1923, or more recent linguists Talmy, 1983, and Langacker, 1987). The figure of an image is defined to be the area that has our focus of attention, and the ground is everything else. Being able to attend to a voice at a noisy cocktail party is a related "figure-ground segregation" ability (at which most humans do quite well). Descriptive sentences are often described in figure-ground terms. In fact, everything that we perceive seems to be subjected to this division (See Figure 1-1).

I believe that seeing the figure and ground of a situation is also related to the notion of "gist extraction" (Hofstadter, 1995). Hofstadter believes that gist extraction should be considered a part of the analogy-making problem description; however, it is rarely included in a model.

4. *Analogy-making should always occur in context*.

All perception occurs in a context and affects the way in which we see things. In the analogical literature, specialized contexts are sometimes referred to as "goals" or "pragmatics" (for example, see Holyoak and Thagard, 1989a).

One may notice that these primary premises do not mention many notions that most researchers would insist be included in a description of a model of analogy-making. For instance: there is no mention of "relations", "objects" or "attributes"; no mention of "rules of mapping"; no mention of "search", "matches", or "slippage."

The goal can be stated succinctly: create a learning model that can perceive the figure and ground of a novel situation, given the context of another. Due in part to the principles' generality, the resulting model, as we will see, is also quite general, and many analogical problems can be expressed in a manner suitable for testing in Analogator's framework. The following section outlines specific types of analogy-making problems explored in this dissertation.

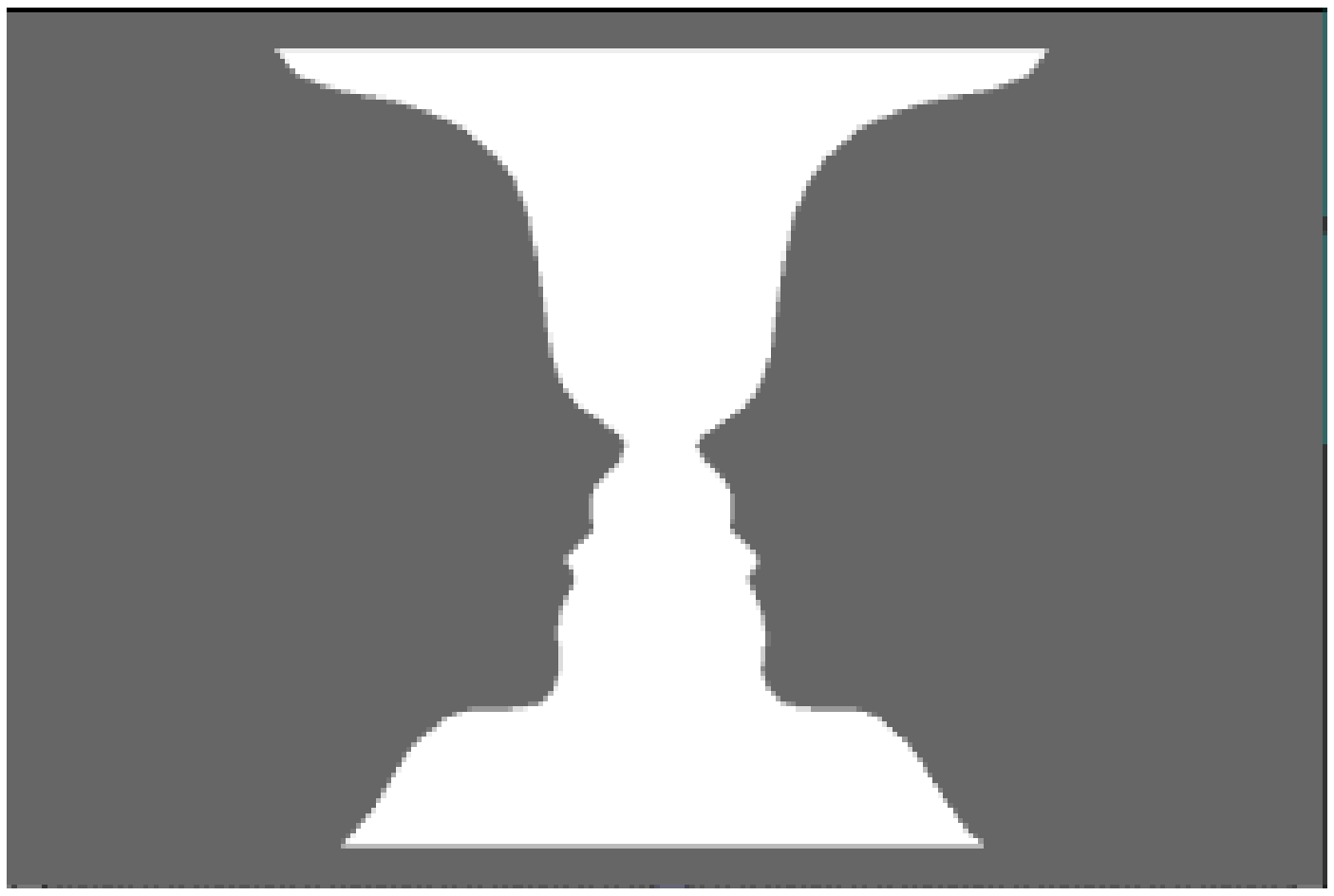

Figure 1-1. A classic example of the battle between figure and ground segmentation. Is this a picture of two faces or a vase?

## 1.3 Analogator's domains

The idea for Analogator's domain is based on Hofstadter and French's Tabletop project (Hofstadter *et al.*, 1995; French, 1992). To illustrate the basic idea, consider the two scenes in Figure 1-2. Quite simply, the goal is to examine the source scene and the element selected in it (as indicated by the pointing hand), and identify the "same" element in the target scene. The problem, of course, is determining what "same" means in a given situation. In any particular problem, there is no "right" answer. A discussion of possible answers is provided in the next section.

### 1.3.1 Type #1: Geometric-Spatial Analogies

The scenes of geometric shapes in Figure 1-2 represent the first domain explored. We will restrict problems in the geometric-spatial domain to be analogies composed of squares, circles, and triangles. In these samples, the colors will be restricted to black and white.

At this point, the reader is encouraged to examine the analogical problem of Figure 1-2 and consider the possible issues that come into play. In attempting to make an analogy between the two scenes in Figure 1-2, one might realize that there is not a white triangle in the target scene like that selected in the source scene. Therefore, a completely literal-minded approach will not work in this particular case. Also, there is not an object in the upper-right corner of the target (like that in the source), so considering just *absolute location* will not solve this problem. One possible solution is to realize that the selected object in the source differs from the other two objects in the source on the dimension of shape. Applying the rule "the object that differs from the other objects on the dimension of shape" would lead one to choose the black square in the target scene. Although that is not the only way to conceptualize the problem, it is one way that captures the basic elements of abstract analogy-making. Notice that in order to make the analogy, one must see how the selected object differs from the other objects. “Differs on the dimension of shape” is an abstract relation that is not directly represented by the scene, but is a perceived category created on-the-fly.

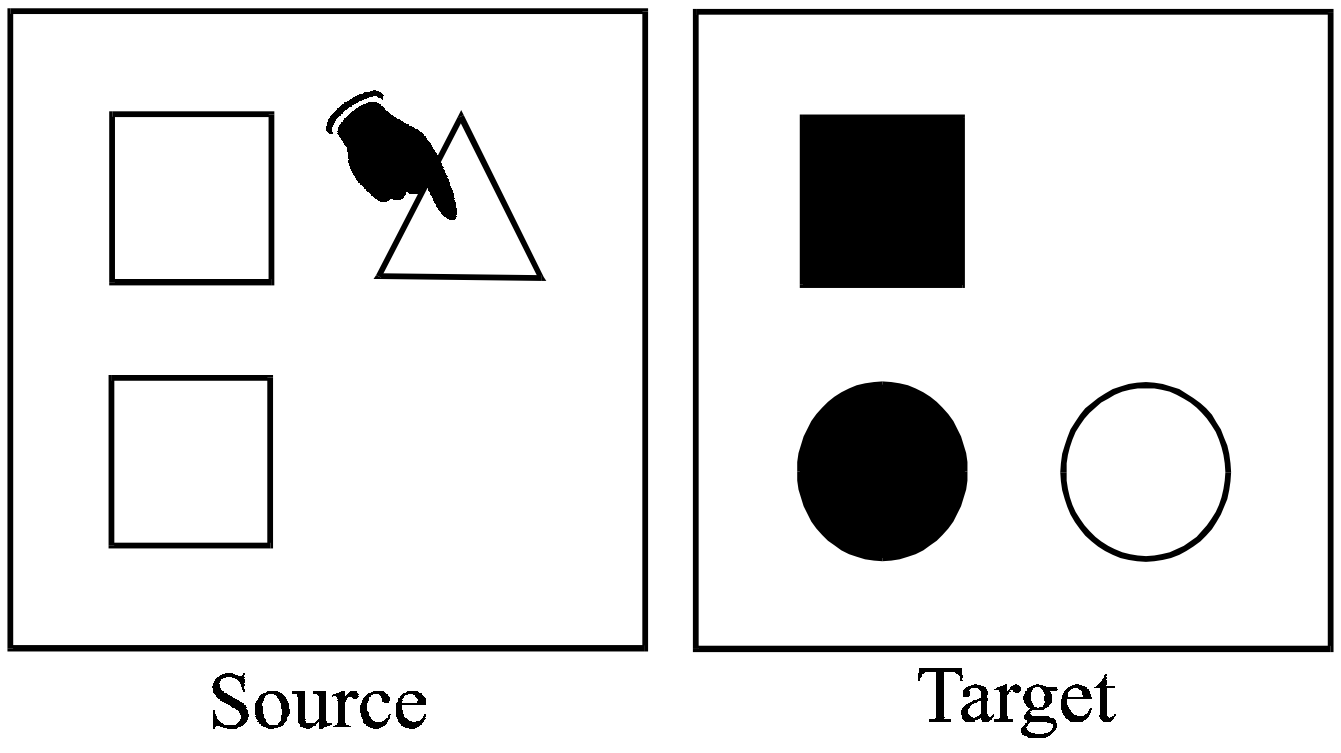

Figure 1-2. Sample #1: A geometric-spatial analogy problem. Which object in the target is the "same" as the object being pointed to in the source? (After French, 1992).

To further illustrate, consider Sample #2 (Figure 1-3). Notice the similarity between Sample #1 and Sample #2. In Sample #2, the white triangle in the source scene has been replaced with a black square. All of the other objects’ shapes, colors, and positions have remained the same. In trying to make an analogy with the scenes in Sample #2, one might recognize that there is a black square in the target scene, just like the one being pointed to in the source. If one put oneself in “superficial mode” this would be the preferred object to choose. But if one resists the superficial treatment of the problem, one might notice that the selected object differs from the other objects in the source on the dimension of color. Again, this is not a category represented directly in the source scene, but is created on-the-fly due to the specifics of this particular problem. Following this line of thinking, one might choose the white circle, as that is the object in the target scene that “differs in the same way” (i.e., “the object that differs on the dimension of color.”) After seeing this relationship, the mapping seems most natural to many people, even though black becomes white, and white becomes black.

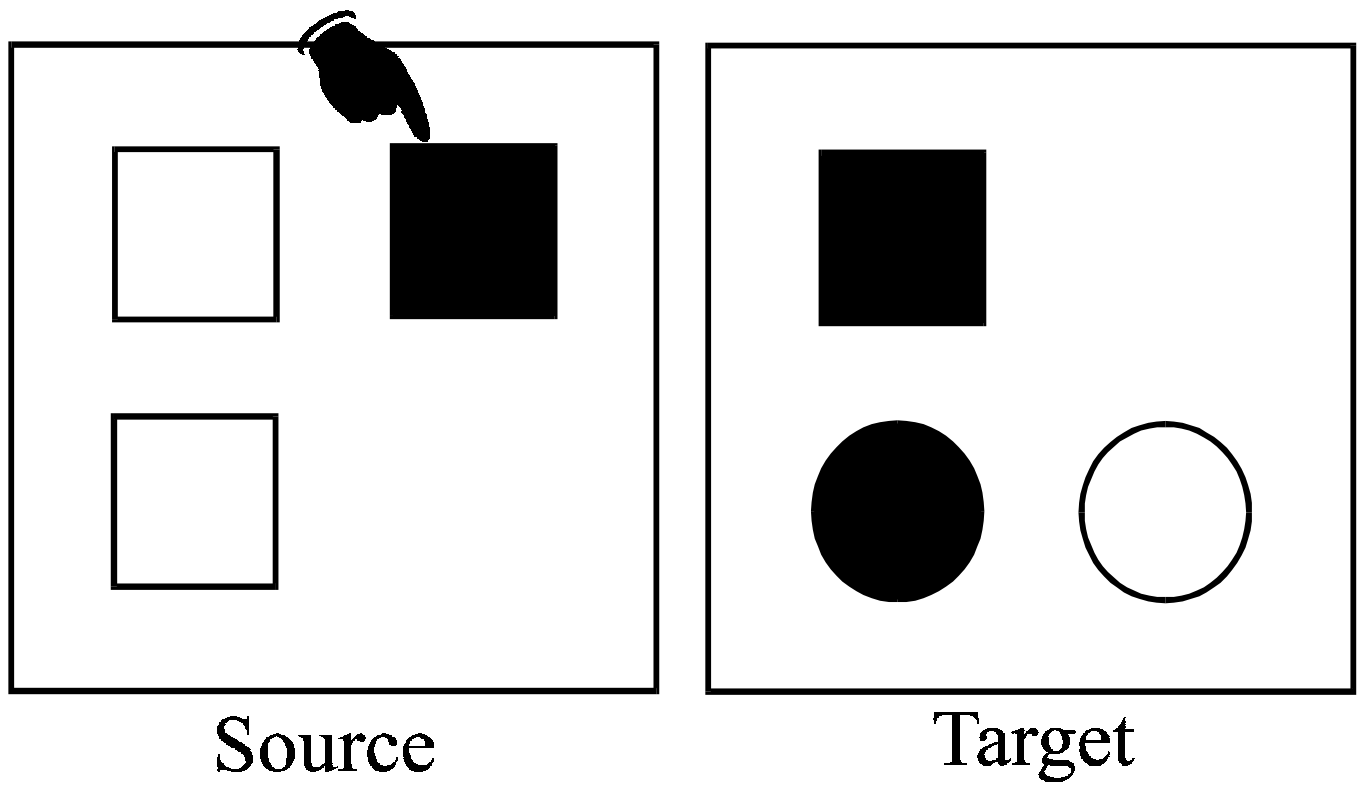

Figure 1-3. Sample #2: A variation of Sample #1. In this version, the black triangle has been changed into a black square. Notice how this change affects the choice of an analogous object.

As demonstrated, by altering an analogy problem just slightly, one might perceive it completely differently. That is, two analogy problems may be superficially very similar, but are perceived very differently. This distinction, related to *categorical perception* (Harnad, 1987), is the flip side of analogy-making (e.g., things may be superficially very different but are perceived the same).

In Samples #1 and #2, we did not consider any object's position in the way we perceived the problem or selected an analogous object. However, one might wish to do so. Consider Sample #3 (Figure 1-4). One is immediately struck by the similarity between the two sets of objects. However, imagine ignoring all object *attributes* (e.g., color and shape) for the moment, and focusing only on an object's position. The selected object is located in the bottom left-hand corner of the source scene. There is an object in the bottom left-hand corner of the target scene, and that could be considered to be the "analogous" object, albeit through a pretty superficial analogy. Thinking more abstractly, one might see the source and the target as mirror images of each other. In that way, the source scene is a backwards L, and the target scene is a normal L (which is a view supported by the objects' attributes). However, one might also see the target scene as being in the same configuration, rotated 90° counterclockwise. Perceiving the problem in this manner, one might choose the black square in the target scene.

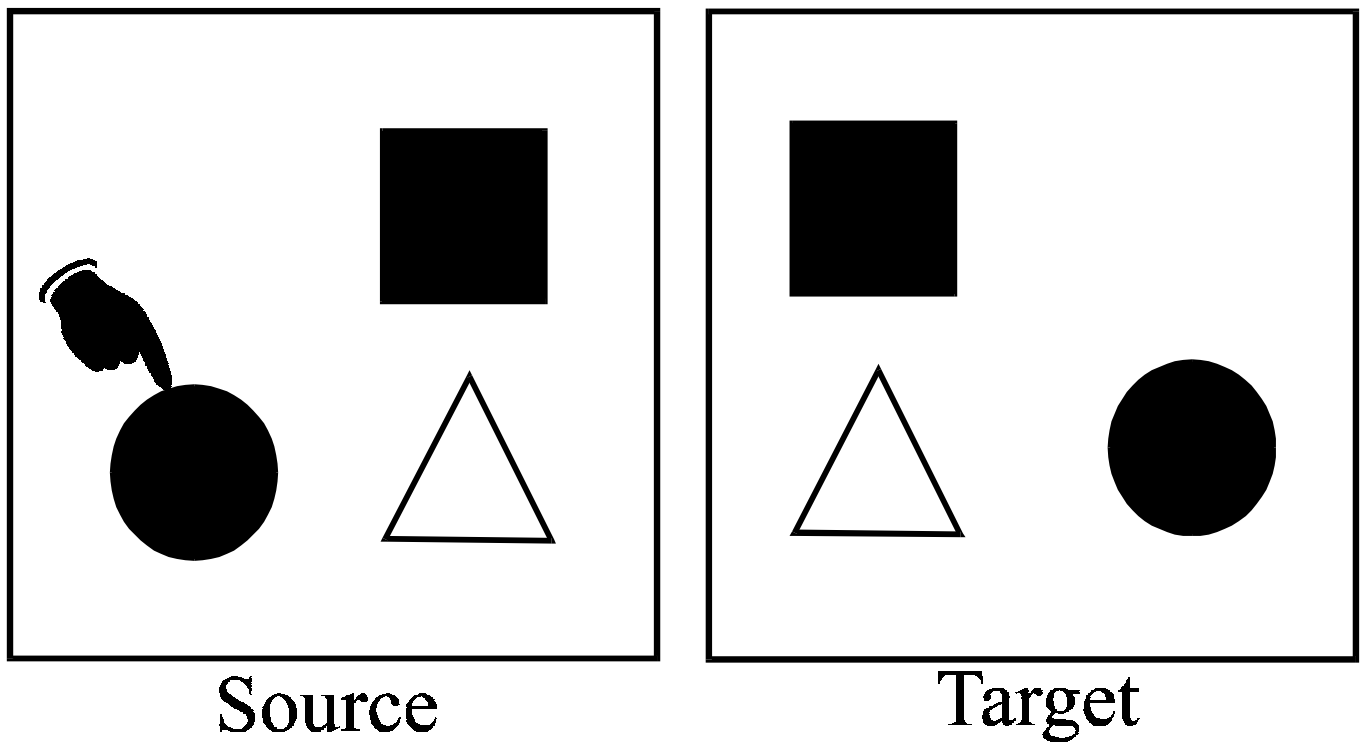

Figure 1-4. Sample #3: Position of objects in a scene may effect perceived similarity.

Finally, one could take into account both position and attributes when making an analogy. Consider Sample #4 (Figure 1-5). In this problem, one might notice that the selected object is one of a pair of same-shape objects. Likewise, the target also has two objects that have the same shape, and a third that does not. However, this does not uniquely identify which is the analogous object in the target scene. If one imagines the target as a mirror image of the source scene and sees the white squares as mapping onto the black circles, then one could pick between the black circles, choosing the one in the bottom right-hand corner.

With geometric-spatial analogy problems, we have seen that this simple domain exhibits many subtleties, requiring concepts such as "mirroring" and "rotation", dimensions such as "color", "shape", and "position", and more complex, on-the-fly categories such as "the object that differs on the dimension of shape." We will now examine another spatial domain.

### 1.3.2 Type #2: Letter-Part Analogies

This domain has also been adapted from Hofstadter and colleagues, specifically from their Letter Spirit domain (McGraw, 1995; Hofstadter and McGraw, 1995). These problems are treated in a similar manner to those in the geometric-spatial domain. However, rather than being arbitrary objects in a spatial arrangement each source and target scene consists of a letter ‘a’. For example, consider the two a’s in Sample #5 (Figure 1-6.) As before, consider the selected portion as indicated by the pointing hand (the selected portion has also been colored gray). The question is: what is the corresponding part in the target scene?

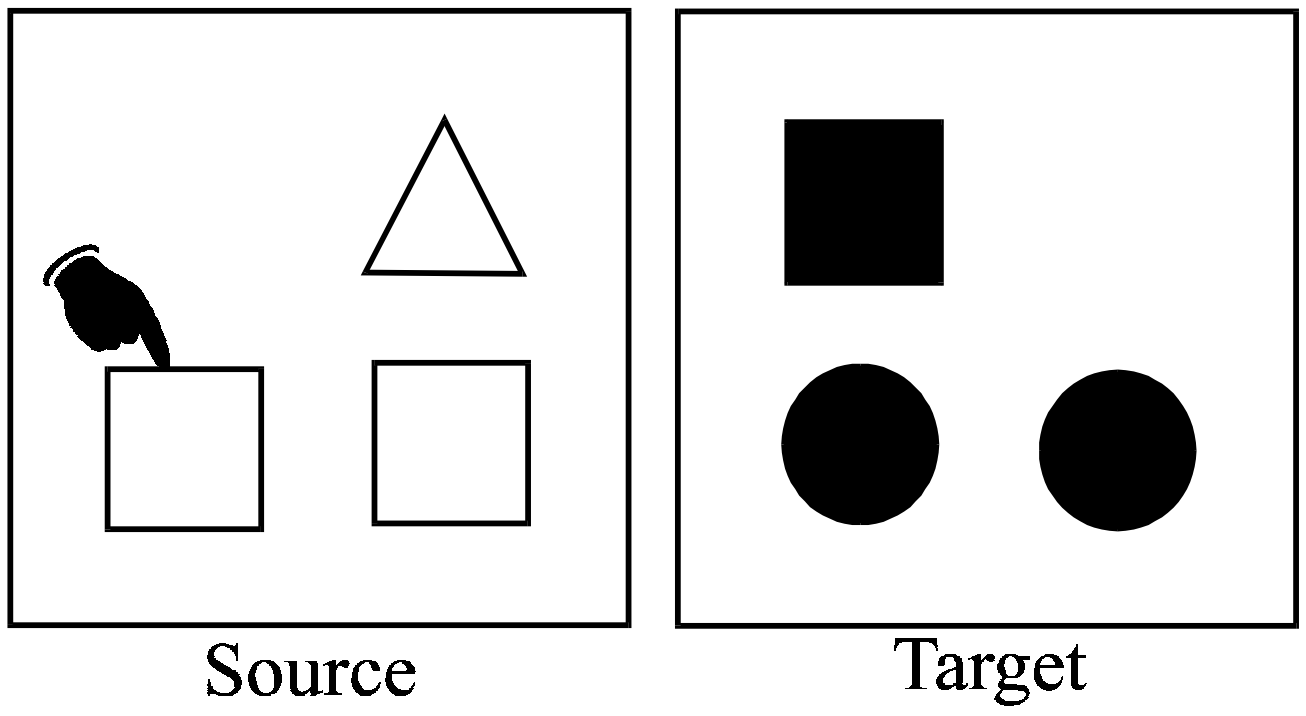

Figure 1-5. Sample #4: Creating on-the-fly categories allows objects to be grouped together in substructures.

In this example, one naturally sees the a’s in the two images, and finds the corresponding parts. In this example, most people would probably consider the topmost

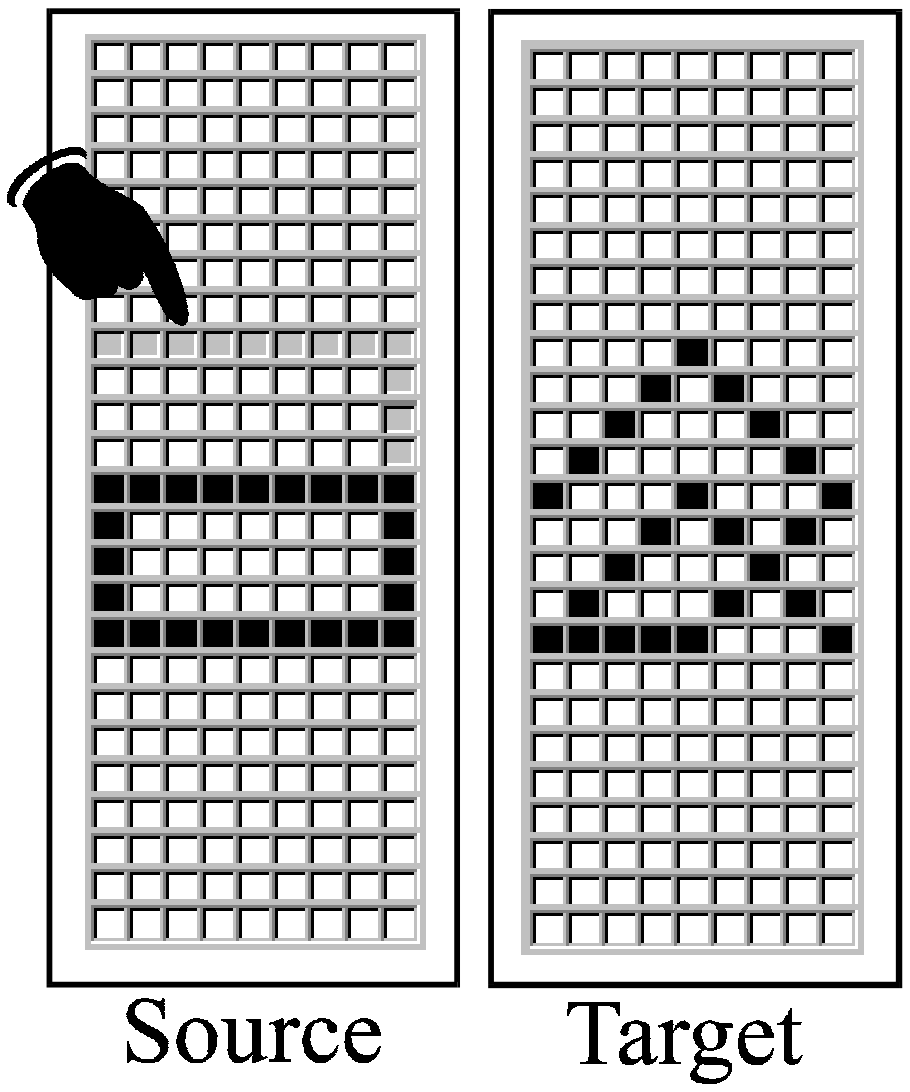

Figure 1-6. Sample #5: A letter-part analogy. Shown are two letter a’s. Considering the selected part in the source scene (pointed to and gray), what is the analogous part in the target scene?

upside-down V in the target letter to be the analogous part.

Consider the similar problem shown in Figure 1-7. The top of the target ‘a’ is quite a bit higher than that of Figure 1-6. Notice that the black squares (or *pixels* as I shall refer to them) that formed the analogous part of the target ‘a’ of Figure 1-6 are also a part of the target ‘a’ of Figure 1-7 (this is illustrated in Figure 1-8). This demonstrates that the same exact pixels may be used for different parts of different a’s. That is, analogous parts cannot be determined by local properties of a small set of pixels, but involve global relationships of the entire letter.

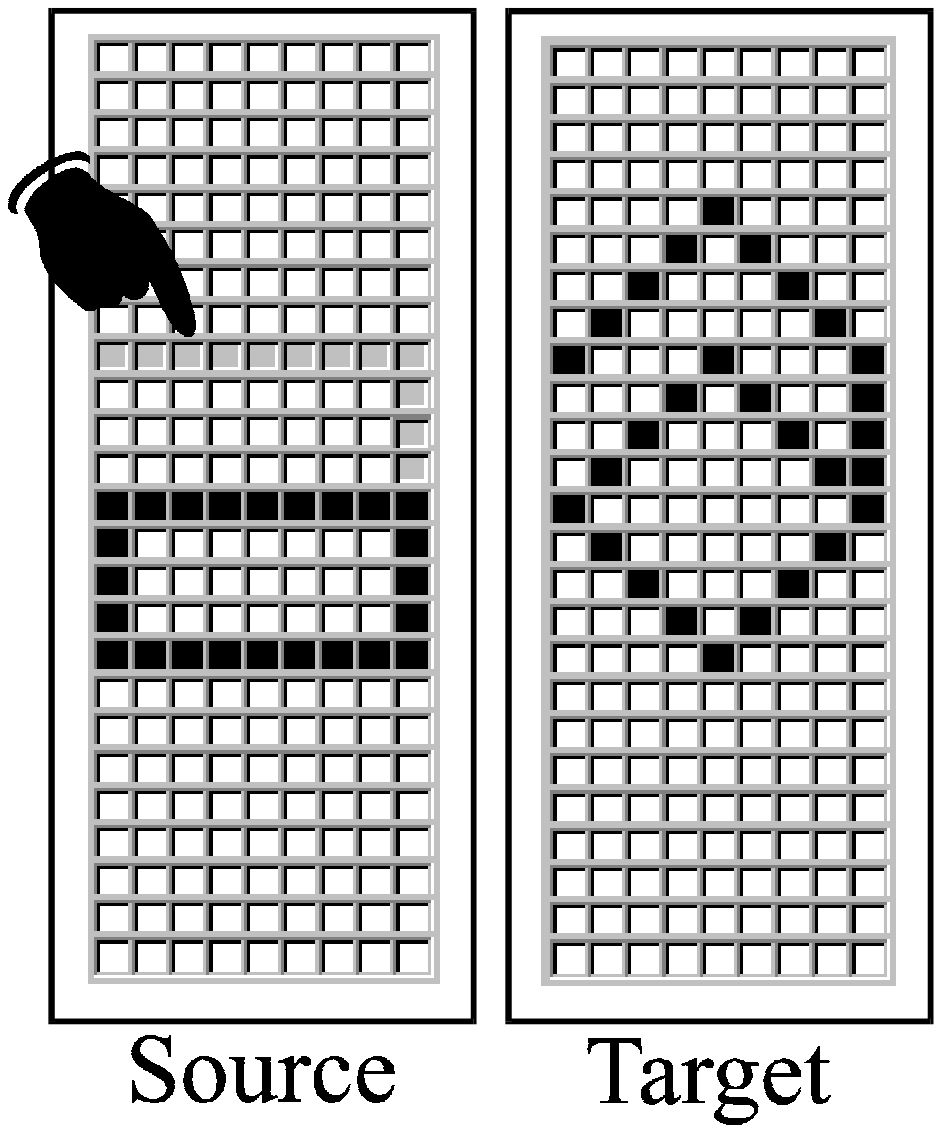

Figure 1-7. Sample #6: A similar letter-part analogy to that of Figure 1-6.

Like the geometric-spatial analogies, the letter-part analogies involve spatial relations. In this domain, there are just a couple of parts with specific (if not well-defined) relationships, but many instances of a single letterform. We will now examine the final domain.

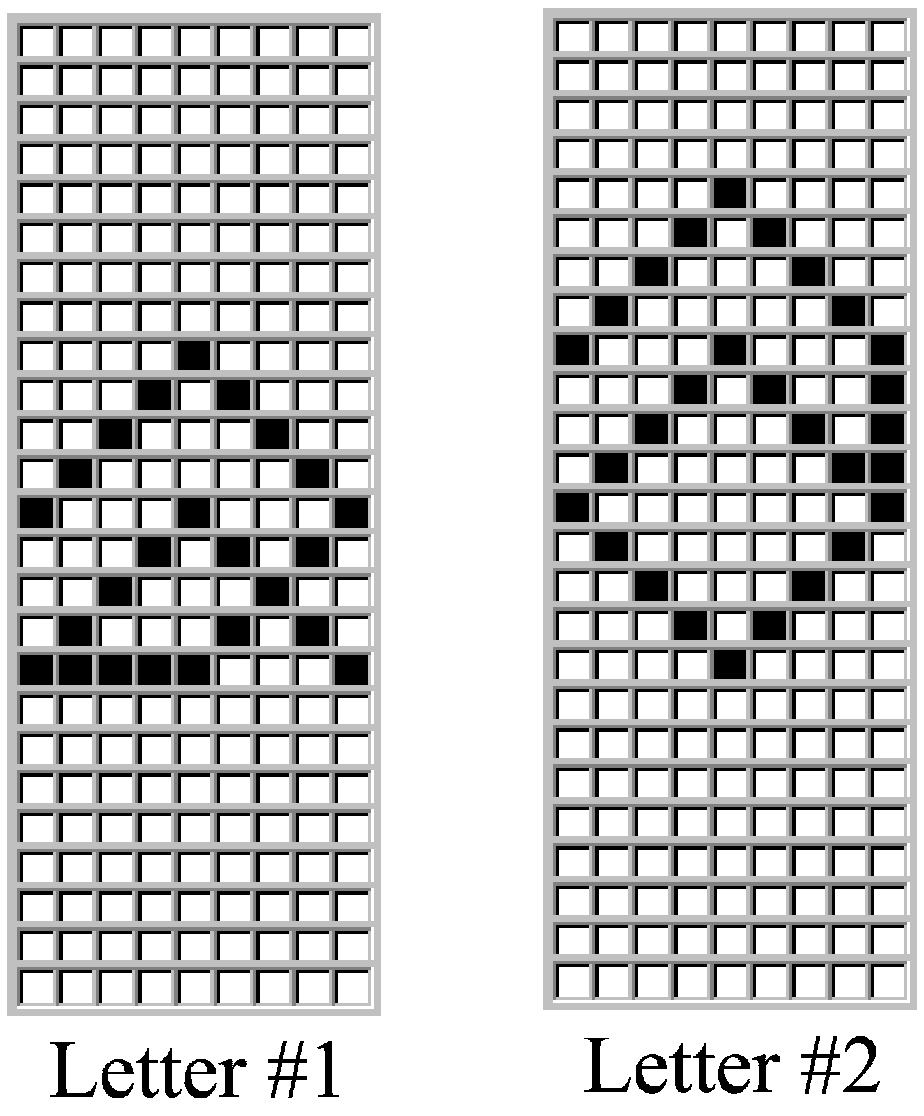

Figure 1-8. Comparison of two a's. Although Letter #1 and Letter #2 have many pixels in common, the parts that those pixels compose are different. A model based purely on "pixel statistics" would have trouble correctly identify the different letter-parts.

### 1.3.3 Type #3: Family Tree Relationship Analogies

The last domain considered in this dissertation is that of family trees, based on a domain created by Hinton (1986). Consider the two families in Sample #6 (Figure 1-9). The question in this domain is: Who in the target family is the analogous person to the person selected in the source? Although the representation of Figure 1-9 is given in a spatial format, I introduce this domain to test Analogator in purely syntactic domains. Therefore, this domain will be presented in terms of syntactic "facts", such as "Christopher is to Penelope as Roberto is to Maria". Given all of the relations represented schematically in Figure 1-9 as facts, can one still identify the corresponding person? Using the graphical version, it is easy to actually see which person is in the "same" place. However, simply using a set of facts to find the analogous person is a much harder task, even for humans. However, by aligning the structures from the two families, one can correctly identify corresponding people.

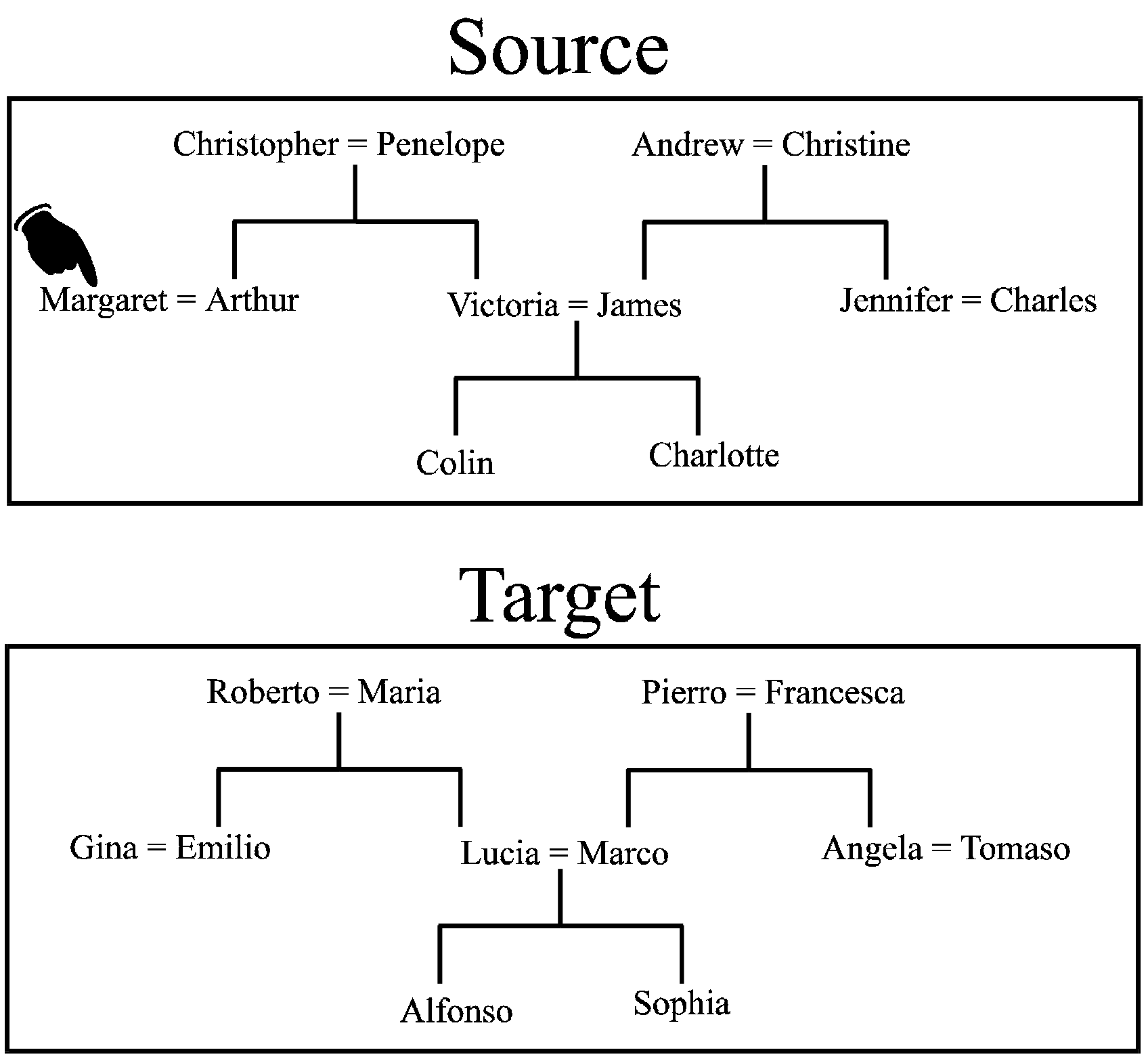

Figure 1-9. Sample #6: two isomorphic families. Who, in the lower family, can be seen as analogous to Margaret? (After Hinton, 1986).

As demonstrated, the Analogator domains contain many of the subtleties of real-world, abstract analogies, yet are distilled down into just a few dimensions. Having examined Analogator's domains, we now turn to a brief overview of the Analogator model.

## 1.4 The Analogator model

Analogator has been implemented as a connectionist network. A connectionist network is a mathematical model based loosely on neurons. By sending simulated activation between simulated neurons and adjusting connections between them, a set of these neurons can learn to associate a given *input pattern* with a desired *output pattern*. Basically, Analogator takes analogy problems as input, and produces an appropriate answer as output. In addition, by generalizing, it can make appropriate analogies to novel problems.

Analogator learns to make analogies by seeing many example analogies; it is trained by having many instances of analogy problems given as input, and requiring the network to produce the desired output. During training, Analogator is told what the appropriate answer is for a given problem. The output is interpreted to be an indication of the object in the target scene that Analogator "computes" to be the most analogous.

In order to perform the types of analogies depicted in the previous section, a new network training method was created, called the *recurrent figure-ground associating procedure*. Specifically, it is an implementation of the assumptions described in Section 1.1. The recurrent figure-ground associating procedure will be described in detail in Chapter 4.

## 1.5 Summary

The research described in this dissertation is fundamentally different from other research on analogy-making; it asks very different questions, and explores very different issues. Unlike most computational models of analogy-making, my goal was not to build a model capable of making sophisticated analogies between abstract scenarios. Rather, my goal was to design a system capable of learning to make analogies by beginning with only low-level perceptual information – those data generally restricted for use in “simpler” tasks, such as categorization and recognition. Although most researchers would not deny that abstract analogy-making is related to these lower-lever processes, little has been done to attempt to bridge the gap between them.

The methods used to model these two types of processes – the categorization of low-level sensory data on the one hand, and analogy-making based on abstract, structured representations on the other – have traditionally been very different. Connectionist models have proved superior in low-level tasks (like categorization), while more abstract, cognitive processes (like analogy-making) have been tied to the classical techniques in artificial intelligence. Analogator was designed to explore the gap between these processes by attempting to explain both within a single framework.

## 1.6 Overview

Chapter 2 examines analogy-making, learning, and generalization. Chapter 3 provides a gentle introduction to the connectionist mechanisms used throughout this dissertation. Chapter 4 discusses the Analogator representation, network architecture and training procedure in depth. Chapter 5 describes a series of analogy-making experiments performed with the Analogator system. Chapter 6 compares and contrasts other computer models of analogy-making, and Chapter 7 summarizes and concludes.

# 2 Analogy-Making, Learning, and Generalization

*Consider that I laboured not for myself only, but for all them that seek learning.*

-Ecclesiastics, Old Testament

This chapter defines and examines the concepts of analogy-making, learning, and generalization and their relationship to each other.

## 2.1 Analogy-Making

To see similarity between two things that might not appear to have much in common on the surface is to make an analogy. In ordinary, everyday thought, analogy-making is a very natural and spontaneous process. When asked to think of a bird and an airplane, one naturally associates the head of the bird with the cockpit, the tail feathers with the rudders, the bird's wings with the plane's wings, the bird's legs with the plane's landing gear, and so on. The human penchant for analogy-making has fascinated philosophers of mind for millennia, psychologists for a century, and cognitive scientists working in the field of artificial intelligence (AI) since the first fledgling analogy programs were developed back in the 1960's. Analogy-making has been recognized as being of central importance in such diverse areas as linguistics (Harris, 1994), creativity and design (Hofstadter *et al.*, 1995), problem solving (Hellman, 1988), and scientific discovery (Holyoak and Thagard, 1995), to name just a few.

For the artificial intelligence researcher, the potential rewards of creating flexible computer programs capable of making analogies are enormous. How much easier, say, programming would be if one could just edit a section of code and then instruct the computer to "do the same thing" to another section, and have the computer adapt the operations in very clever, analogous ways. Unfortunately, current computer programs, even those that learn, tend to be very "literal minded." The hope is that computer programs can be made more intelligent and useful by incorporating an ability to make analogies.

To the cognitive scientist, the prospect of creating programs capable of making analogies is tantamount to understanding the cognitive phenomena of analogy-making itself. Analogy-making takes many forms, variously termed by cognitive scientists: recognition, categorization, induction, generalization, analogical reminding, analogy-making, analogical problem solving, and analogical reasoning. Each form of analogy-making is slightly different from the others, however all involve the *perception of similarity*. Humans are incredibly flexible at perceiving similarity in a multitude of ways. For example, consider the following vignettes:

1. A woman goes to a potluck picnic with a frisbee and some potato salad. After throwing the disc for awhile, she gets in line to eat dinner. She soon realizes that there aren't any plates left. Hungrily, she looks down and sees the frisbee, but now she sees it as a plate. She scoops the food onto her frisbee-plate and enjoys her meal.

2. Two grandparents walking along a trail get tired. They wish they had a couple of chairs, but then they see two large, flat boulders. They wearily walk over to the rocks, sit down, and have a nice rest.

3. Two graduate students are complaining about how much paperwork they are required to fill out and turn in to the administration when one student remembers that he forgot to send his evil step-mom a mother's day card.

4. John Rogers, a comedian, likens graduate school to "the snooze button on the clock radio of life," and his audience laughs.

5. A physics student begins to understand the properties of light by comparing and contrasting the behavior of water waves and marbles.

6. A little girl points to the moon and says "ball."

7. A doctor recognizes a set of symptoms and diagnoses a disease.

8. A presidential hopeful attempting to remind voters of past events, refers to another candidate’s activities as “Whitewatergate”, “Travelgate”, and “Filegate”.

Although some of these examples are more abstract than others, I believe that each of these represents an act of analogy-making. It is a wide spectrum ranging from mundane, low-level recognition and categorization to sophisticated, abstract analogy and metaphor. As the above examples illustrate, analogy-making appears both spontaneously and deliberately in the thought processes of people of all ages in many different kinds of situations. Understanding this wonderful, pervasive ability we call analogy-making would represent a giant leap forward in understanding the full breadth of human thinking.

## 2.2 Learning

Although everyone seems to know what it is, learning is actually very difficult to precisely define. Roughly, any system that improves its performance in response to internal changes caused by experience can be said to learn. However, this brief definition is too broad and too vague to be of much use. For instance, when we oil a squeaky door it may have had an “internal change”, but we do not mean that it has learned anything about being quiet. Wine and cheese in the cellar may be improving their “performance” due to “experience”, but we would not say that they are learning. Tempering a sword could be described as “improving its performance in response to internal changes caused by experience,” and yet we do not mean that it has learned anything. Flowers interactively adjust themselves to face the sun, but, again, no one would claim that they have learned about the position of the sun. Then, what *do* we mean we mean when we say a system has learned? To better illustrate the learning phenomenon, consider these vignettes:

1. Within a few months, a baby has acquired the ability to recognize his mother.

2. After just a couple of years, a young toddler discovers how to get the undivided attention of her parents in many ways.

3. A young boy burns his hand on the stove. It never happens again.

4. A man trusts a stranger, and the stranger takes his wallet. It never happens again.

5. After years of practicing, a young woman develops an ability to play the tuba.

6. A man goes to a potluck picnic with a frisbee and some coleslaw. He soon realizes that there aren't any plates left. Hungrily, he watches a woman scoop food onto her frisbee-plate and enjoy her meal. He makes a mental note, and does the same.

7. A physics teacher explains the properties of light to a student by comparing and contrasting the behavior of water waves and marbles, and the student passes a test.

8. A robot's sonar sensor registers 0.85, it moves forward, and then its bumper sensor is activated. Later, after some internal adjustments, its sonar sensor again registers 0.85, it moves *backward*, and its bumper sensor remains inactive.

9. A man goes to medical school and develops the ability to diagnose a disease by examining many patients with the disease.

10. After studying a math textbook and practicing for many days, a boy is able to recite from memory the first 100 digits of $\pi$.

11. After watching Jeopardy, a woman in New Jersey now knows which mammal runs the fastest.

12. A man knows that if he receives a promotion then he will get a bonus. Later, he finds out that he will be promoted. Therefore, he deduces that he will get a bonus.

These examples demonstrate the wide range of phenomenon that we call learning. Cognitive scientists have given these forms of learning various names: learning by deduction, learning by abduction, learning by induction (also called learning from examples), learning by memorization, learning from a single example (also called one-shot learning), learning by analogy, rote learning, learning by instruction (also called learning by being told), and learning by observation, to name just a few too many. Although each of these forms of learning emphasizes a different aspect of learning, they all involve a change to an internal, persistent *memory* of the system. Other than this single fact, the above examples have little else in common.

As can be seen, 'systems that learn' is a diverse category. It also appears that the category is getting larger. Carbonell, Michalski, and Mitchell (1983) define a type of learning that involves the "direct implanting of new knowledge":

> …Variants of this knowledge acquisition method include: Learning by being programmed, constructed, or modified by an external entity, requiring no effort on the part of the learner (for example, the usual style of programming).
>
> (Carbonell, Michalski, and Mitchell, 1983)

This type of 'learning' is much too inclusive, encompassing not only sword tempering, and door oiling, but program writing, and building construction. For this dissertation, we will limit our discussion to the category 'learning by example'. Furthermore, we will limit that category by considering only learning methods capable of *generalization*.

## 2.3 Generalization

The word 'generalization', as we normally use it, refers to the ability to infer the general from the particulars. For instance, if a robot with a visual system successfully learns to avoid bumping into a trash can, and can avoid the trash can even though it might be in a position never before seen, then we can say that the robot has learned to generalize from previous particular instances of trash can images. Generalizations of this type are often based on statistical regularities found directly in the sensory stimuli.

However, generalizations need not be tied to low-level perception. Recall the man from example 12 from above who got a promotion and then a bonus. One could say that he is simply using *modus ponens* to deduce the previously-unknown fact that he will get a bonus.[2] However, one could also argue that he is generalizing from examples he has seen before. Imagine that, instead of using *modus ponens*, he had made a generalization based on the following story he had read:

> A woman knows that if she gets caught dating a fellow employee then she will get fired. Later, she finds out she has been caught. Therefore, she predicts that she will get fired, and she does.

In effect, the man can be said to have learned *modus ponens* by example. Having generalized the previous story, he concludes that he will get a bonus. Notice that this type of generalization appears quite different from the generalization made by the robot viewing trash cans. The robot was able to generalize based on common properties found directly in its low-level visual system. Although, the generalization of the notion of *modus ponens* may be suggested by low-level perception (i.e., the examples had to enter the man's head through his senses), it is a generalization of abstractions, such as things known to be true, consequences of actions, etc.

---

[2] Recall that *modus ponens* is the logical rule of inference that states "if 'A implies B' is true, and 'A' is true, then 'B' necessarily follows."

Abstractions, such as 'truth' and 'consequences', are concepts. The word 'concept' has probably been used in more varieties of ways than 'learning'. Informally, a concept is a general 'idea' inferred from specific instances or occurrences. So, to the robot, 'trash can' is also a concept. In effect, concepts are themselves generalizations.

Therefore, the robot made generalizations based on similarity found directly (or nearly directly) in its low-level sensors, while the notion of *modus ponens* was a generalization based on similarity found in higher-level descriptions. Following Hofstadter's terminology, I shall call the process of perceiving from sensory stimuli *low-level perception*, and the process of perceiving from abstractions *high-level perception* (Hofstadter *et al.*, 1995).

We have seen that analogy-making is the perception of similarity between two things that could be very dissimilar on the surface, and learning by example is the process of altering an internal memory so that a system can make generalizations. Also, we have seen that generalizations are inferences formed from perceived similarity. Therefore, analogy-making and learning-by-example are intimately connected to the perception of similarity via the process of generalization.

## 2.4 Analogy-Making and Learning

Considering the amount of overlap between analogy-making and learning, it is ironic that the interaction between these two areas of research has been so limited. That is not to say that there has not been much research associating these two processes, but that research has traditionally been restricted to two categories of models: those that learn low-level perceptual processes, and those that learn by making analogies between high-level descriptions. Analogator defines a third category of models – those that learn what it means to make an analogy. The Analogator model will be examine in detail in Chapter 4. These first two categories will be explored in more detail here.

### 2.4.1 Low-level Perceptual Processing

Recognition and categorization based on sensory stimuli are two tasks studied via low-level perceptual processing models. In the last fifteen years, programs designed to learn low-level perceptual processes have flourished. Artificial neural networks (ANNs) have become the mechanism of choice for studying low-level perceptual processes. In the mid-1980's, research in ANNs, also called connectionist networks, was revitalized (after a long period of inactivity) due to the creation of many new and interesting learning algorithms. We will examine connectionist networks in detail in Chapter 3; briefly, a connectionist network is a model that is based loosely on neurons and can learn by being exposed to examples. Many such connectionist learning procedures do quite well with low-level perceptual tasks, such as the categorization of printed or hand-written characters. Also in the 1980's, the field of machine learning exploded with many other types of learning schemes designed for other generalization tasks (see, for instance, Marr, 1982; Holland, 1975; and Holland *et al.*, 1986). However, connectionist systems by their very nature are better suited to learning low-level perceptual processes than are many other machine learning techniques. This is due to the fact that connectionist networks can be directly connected to low-level stimuli, and can learn to re-represent a problem.

For example, connectionist-based optical character recognition (OCR) systems can view digital photographs (or similar input) and produce category labels after learning is complete. For instance, many large cities in the United States now have connectionist OCR systems capable of reading hand-written ZIP codes on envelopes to help with the automatic routing of mail. Of course, people's hand-writing varies widely, yet these systems work very well. Typically, the programs in this category have very narrowly defined tasks; they are designed to take low-level stimuli, such as a picture of the number 4, and produce a category, such as "four". Currently, these models are only capable of solving very basic problems, and, therefore, have had little connection to more high-level tasks.

### 2.4.2 High-level Conceptual Processing

The second category of models, those that learn by making analogies between high-level descriptions, has also had a recent surge of active research. Models in this category attempt to solve a problem by comparing it's highly abstracted representation to other abstracted representations of problems that the system has previously seen (and solved).[3] The idea is designed to work as follows. Imagine a robot attempting to vacuum a carpet. The robot plugs the vacuum cleaner in and the vacuum cleaner works for a moment, but then it stops. Not knowing much about how vacuum cleaners work, the robot searches its memory and recalls a similar event that happened before when it was attempting to, say, make some toast. It remembers that the toaster stopped working, but was fixed by jiggling the toaster's lever. The robot recalls this toaster-episode from memory, *maps* it to the current situation, and, in a clever fashion, jiggles the vacuum cleaner's *on/off switch* (the analogous part to the toaster's lever). The vacuum cleaner comes back on, and the robot is able to complete its chores. As this story illustrates, there are two main goals of such an analogical model: 1) the retrieval from memory of similar episodes, and 2) the application of the recalled solution to the current problem in an analogous fashion. Notice that both goals completely rely on the perception of similarity: the first must see the similarity (or not) between the current problem and all other episodes stored in memory, and the second must match similar pieces of the retrieved memory episode with the current problem.

[3] 'Representation' will be formally defined in Chapter 3.

The basic idea of using analogy as a computational tool to solve new problems is relatively old, dating back to at least the early 1960's (Minsky, 1963). Many researchers throughout the last three decades have attempted to construct AI programs in this spirit. One of the most successful AI research paradigms ever is based on these basic ideas: Schank's Case-Based Reasoning (CBR) approach (Schank, 1982).

All of the models in this category posit the existence of an analogy-making 'engine'. That is, the mechanism that is responsible for seeing similarity and finding correspondences is hard-coded and never changes. Learning, if involved at all, is limited to altering and adding memory episodes rather than developing or honing analogy-making abilities. The perception of similarity in these models operates quite differently than the low-level perceptual models, as the representations used are very different from sensory stimuli. Each representation is a highly abstract description of the problem's gist. Because the representations take a very different form from those of low-level stimuli, learning is hardly ever incorporated into the modeling of the perception of similarity between highly abstracted descriptions.

In practice, a problem's representation cannot be too different from those of the solutions stored in memory or the program would be unable to see any similarity. Because of this limitation, much effort is spent attempting to get representations into the correct form. Also, these models generally suffer from a boot-strapping problem: the model needs to have a solution in memory to solve a new problem, but where do the initial memories come from? Typically, the researcher must supply them. Although the role of learning is restricted to altering memory rather than being used in the perception of similarity, the analogy-making process is of central importance. Generally, these types of models have little interaction with low-level sensory stimuli.

The goal of Analogator is to explore the large gap between those systems that learn the very process of perceiving similarity and those that are hard-coded to make analogies between abstractions. There are many issues that can be explored in this chasm: Can a single mechanism perform both low-level and high-level generalizations? Since they are generalizations themselves, can concepts be created by this same generalizing mechanism? What is the role of concepts in such a system? How could a system learn this general generalization mechanism? These are core questions in AI and cognitive science.

In order to explore these questions more fully, the traditional approach based on making analogies between high-level descriptions is described and examined in the following section.

## 2.5 A sketch of the traditional approach to analogy-making

Intuition has been the mother of invention, at least in AI. Many high-level cognitive tasks have been modeled by AI researchers implementing their intuitions. Planning, for instance, is a task that has been traditionally modeled in terms of "goals", "problem spaces", "goal states", etc. Intuitive concepts, such as goals, represent high-level descriptions that, it is hoped, mirror the actual cognitive processes and mental representations that people actually use to solve problems. Goals, problem spaces, and their like, have been suggested by introspection. That is, as people attempt to solve a problem, they report their self-perception of what is going on in their mind; people report in terms of goals and problem spaces. For instance, someone attempting to fix a car will say statements such as: “Before I can check the battery, I must open the hood. But I remember that the hood is stuck, so that idea leads to a dead end. I’m now going back to the drawing board…” In this scenario, there is a goal to check the battery, and a plan that involves opening the hood. It makes sense, it appears, to model planning using goals, goal states, etc.

Like planning, analogy-making has been traditionally modeled with similar introspective concepts. For analogy-making, the introspective concepts are items such as *correspondences, objects, attributes, relations*, and *mapping processes*. When expressed in these terms, one imagines analogy-making as creating explicit links via a “mapping process” between corresponding elements. So, if one were thinking of a bird and an airplane again, then one might model the analogy-making process by creating connections between the bird’s and the airplane’s representations. The two objects could be symbolized by high-level descriptions, such as those shown in Figure 2-1.

Mappings can now be made in a straightforward manner by matching the abstracted part names (i.e., CONTROL-SOURCE) in the two representations. So, for instance, *head* would map to the *cockpit*, the *tail-feathers* would map to *rudders*, etc. If one then asked, "What corresponds to the bird's legs?" one could search through the representation of the bird, find *legs*, and simply follow the *SUPPORT* link to the plane's representation to find *landing-gear*.

To learn to make analogies between birds and airplanes seems to require a very different mechanism that that of learning to generalize over different views of trash cans. Yet, combining these two types of generalization into a single framework is ultimately my goal. To examine this possibility further, we will need to see exactly how connectionist networks generalize.

## 2.6 Generalizing in connectionist networks

Connectionist learning networks, as we will see in Chapter 3, have the power to generalize over novel patterns. However, in order for this to happen, one of the following three conditions must be met:

```
(bird
     (CONTROL-SOURCE head)
     (LIFT-MECHANISM wings)
     (SUPPORT legs)
     (STEERING-MECHANISM tail-feathers))

(plane
     (CONTROL-SOURCE cockpit)
     (LIFT-MECHANISM wings)
     (SUPPORT landing-gear)
     (STEERING-MECHANISM rudders))
```

Figure 2-1. Abstract representations for comparing a bird with an airplane.

1. There must exist statistical regularities in the representations.

2. There must exist statistical regularities in the way that representations are used.

3. A combination of 1 and 2.

In this sense, generalization is merely *interpolation*. That is, networks must have experience with similar patterns in order to make useful generalizations with new ones. Unfortunately, Analogy-making is often seen as *extrapolation*. This implies that analogy-making is, in some way, more powerful than the generalization abilities of a network would allow. However, we shall see in Chapter 4 that there are some techniques to reduce extrapolation to interpolation. First, we must examine the connectionist mechanisms in more detail.

# 3 Connectionist Foundations

*If you have built castles in the air, your work need not be lost; that is where they should be. Now put the foundations under them.*

-Henry David Thoreau

Cognitive models fall into two broad categories: the *symbolic* and the *subsymbolic*. It is difficult to clearly define the two categories; each is usually described by its tendencies rather than any single definitive property. Symbolic processing is characterized by hard-coded, explicit rules operating on discrete, static tokens. Subsymbolic processing is associated with learned, fuzzy constraints affecting continuous, distributed representations. These two categories roughly define the ends of a spectrum of models.

The Analogator project is firmly rooted in the subsymbolic part of the spectrum; Analogator is implemented as a connectionist network operating on distributed representations. The goal of this chapter is to provide background knowledge of connectionism sufficient to understand the basic terms and operations of general subsymbolic computation. Specifically, this chapter examines motivations for using connectionism, learning via back-propagation, hidden layer activations and analysis, simple recurrent networks, and tensor product algebra.

## 3.1 Why connectionism?

My foremost reason for choosing connectionism as the basis for a cognitive model is its potential for learning and generalization. Adaptive connectionist networks are often capable of detecting statistical regularities in the set of input patterns presented to them. After being suitably trained, connectionist networks are able to perform in reasonable ways when exposed to novel input patterns based on what they have learned during training. This ability, often called *generalization*, *induction*, or *interpolation*, allows a network to operate much more flexibly than a system that relies on explicit, rigid ‘rules’.

Connectionist networks also generally exhibit graceful degradation of their performance as a task increases in difficulty or in the presence of noise. This is in contrast to symbolic systems, which typically cannot handle noise, and have clear processing limits. Generally, symbolic systems either work, or they do not—there is no middle ground. Connectionist networks also operate in a parallel fashion, so processing can be quite fast once learning has been completed.

However, there are some major disadvantages to connectionism. First, there is no guarantee that a network will be able to learn a given task. In addition, even if the network is able to find a solution, it might take a very long time to reach an acceptable level of performance. Connectionist networks typically learn by gradually adjusting their weights during training, and may require many exposures to a training set before a network performs appropriately. Even if a network does learn a task, it may learn to generalize improperly, and not perform well on novel input patterns. Finally, if a network learns to perform as desired on a given problem, there is not generally a way to communicate in abstract terms *how* the network has solved the problem. That is, connectionist networks cannot provide a simple explanation of their solutions. However, in spite of their problems, connectionist networks have shown their usefulness on many tasks, especially low-level recognition and categorization problems, such as optical character recognition.

## 3.2 Connectionist networks

A *connectionist network*, sometimes called a *parallel distributed processing* (PDP) network, is a model which is based loosely on neural architecture. Connectionist networks attempt to capture the essence of neural computation: many small, independent units calculating very simple functions in parallel. These networks are composed of two basic building blocks: idealized neurons (often called *units*) linked via weighted connections. Each unit has an associated *activation* value, which can be passed to other units via the links with the connection weights mediating the amount of activation that is passed between units.

In a typical connectionist network, activation flows from unit to unit over the weights like electricity over wires. The units in *feed-forward* networks are organized into layers with no connections from one layer back to previous layers, so all activation flows in one direction without cycles. In most such networks, there are three types of layers: *input*, *hidden*, and *output*. If the units in a layer receive activation from outside the network, then the layer is called an *input layer*. An *output layer* produces activation representing the result of the network's computation, and a *hidden layer* is defined as a layer that is neither an input nor an output layer. Furthermore, layers can be subdivided into sections called *banks*.

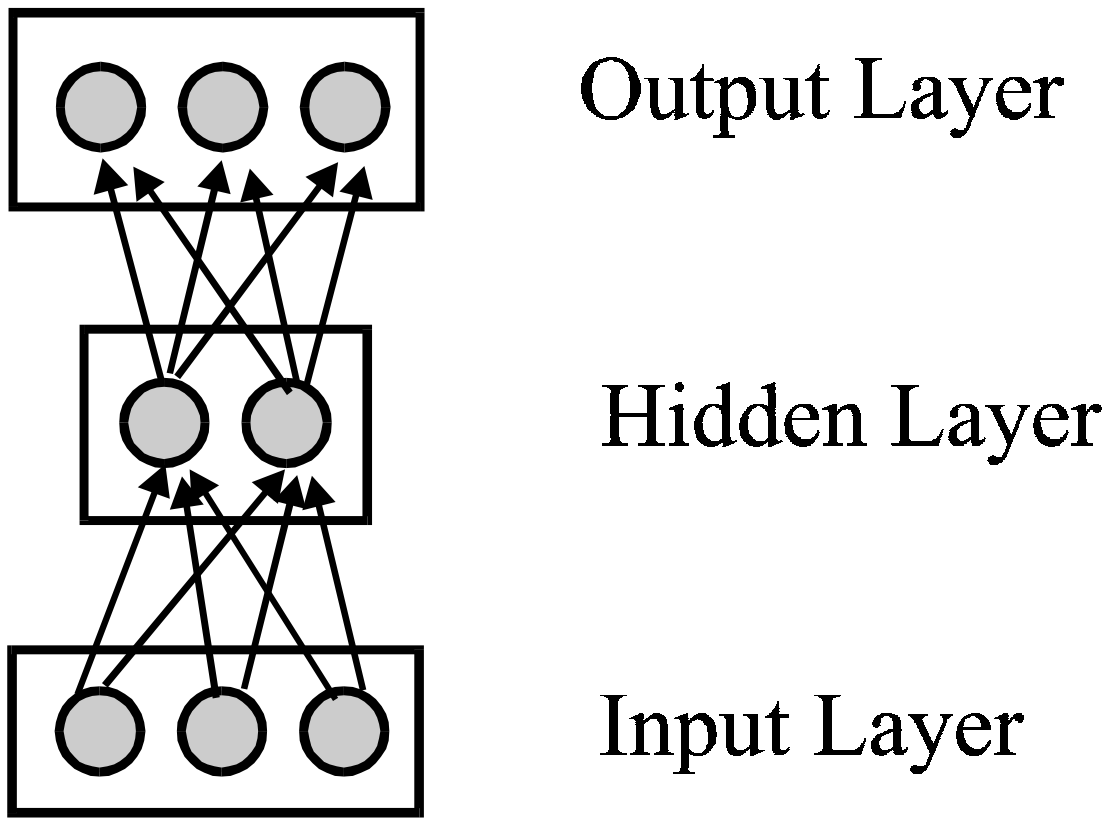

Figure 3-1. A 3-layer, feed-forward network.

Figure 3-1 shows a schematic diagram of a standard 3-layer, feed-forward network.[4] Layers are usually displayed with input layers on the bottom and output layers on the top, as shown. The gray circles represent units, the rectangles represent layers, and the arrows represent weighted connections. Usually when networks are large, not all of the weighted connections are drawn.

Briefly, processing in a network occurs as follows. The input layer is loaded with a set of activations called an *input pattern*. Activation flows across the weights between the input and the hidden layers. At each unit in the hidden layer, the incoming activations are summed and passed through an activation function. The hidden activations then flow from the hidden layer to the output layer. At each unit in the output layer, the new activation is calculated as before. The resulting activations emerge from the network as the *output pattern*. The process of propagating activation from the input layer through the hidden layers to the output layer is called the *propagation phase*.

---

[4] The number of layers in a network is sometimes calculated by not counting the input layer. This is often warranted as the input layer does not do any computing but serves only as a place to load the input values. However, I will count input layers whenever I describe a network architecture. Therefore Figure 3-1 shows a 3-layer network.

Figure 3-2 shows a single unit *m* of a connectionist network. The total activation coming into a unit is called the unit's *net input*.[5] The net input of unit *m* is calculated by summing the incoming activations multiplied by the associated weights, and is labeled

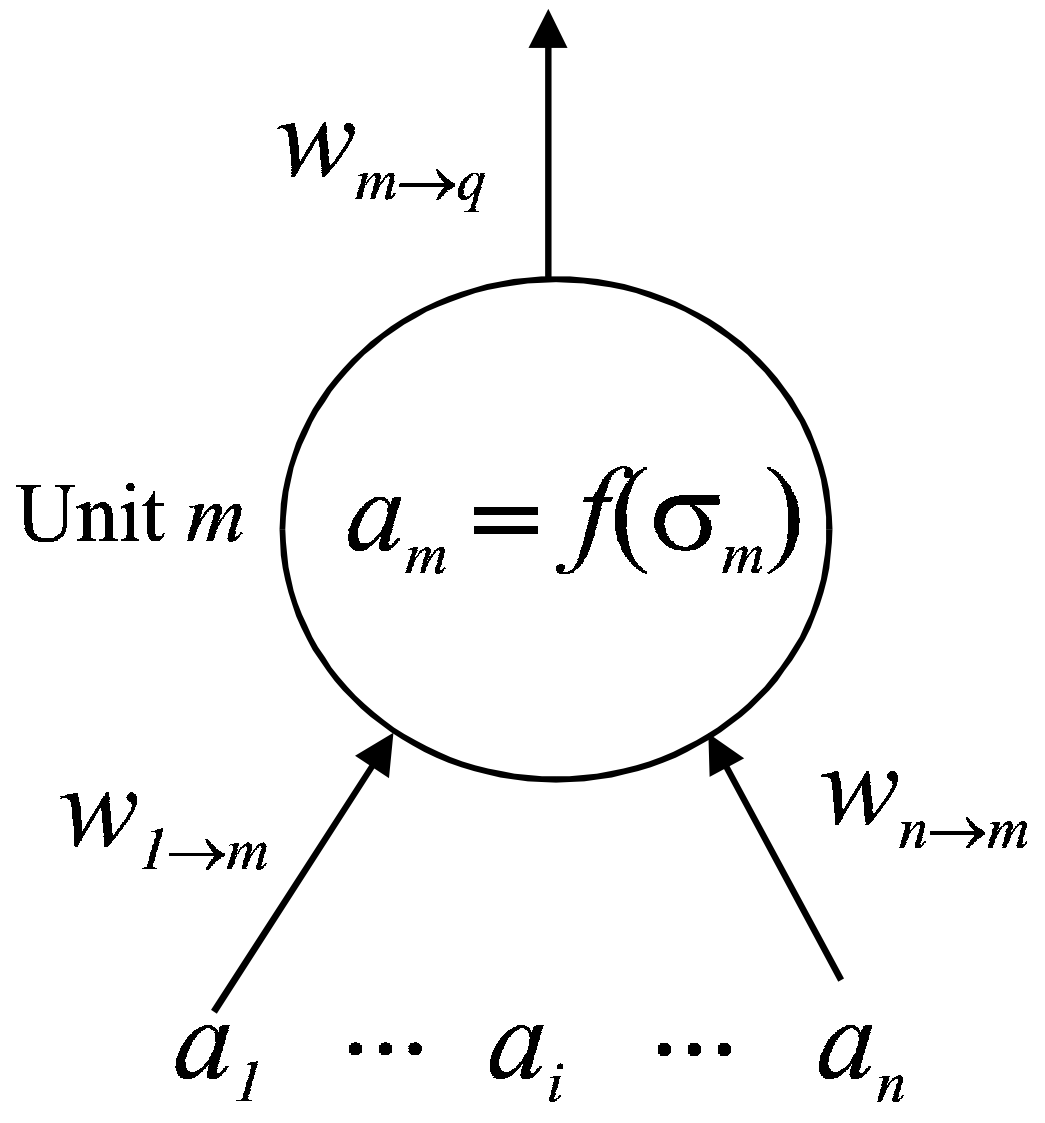

Figure 3-2. An idealized neuron. Activations come in from units 1 through *n*, and are propagated to unit *q*.

---

[5] Here, 'net' is not short for 'network' but refers to the "total left after deductions," as in *net income*.

$\sigma_m$. The net input is only computed for hidden and output layers, not input layers. For input layers, the activation is defined to come from an external *input pattern*. The net input to a unit *m* in the hidden or output layer is thus defined to be:

$$\sigma_m = \sum_{i=1}^{n} a_i w_{i \to m} \tag{3-1}$$

where *i* is the index of the units in the previous layer that are connected to unit *m*, $a_i$ represents the activation of unit $i$, and $w_{i \to m}$ represents the weight of the connection from unit *i* to unit *m*. The unit's *activation function*, *f*, is then applied to the net input, $\sigma_m$, yielding the new activation value. Typically, *f* is a logistic function of the form:

$$f(\sigma_m) = \frac{1}{1 + e^{-\sigma_m}} \tag{3-2}$$

Applying the logistic function to the net input σ squashes it down into the range from 0 to 1. Figure 3-3 shows the squashing function with σ plotted on the x-axis, and *f*(σ) on the y-axis. The squashed net input becomes the new activation for unit *m*, and is used to calculate other net inputs in subsequent layers (e.g., in Figure 3-2, the activation is passed on to unit *q*). This particular squashing function is used because it is non-linear and its first derivative has special properties that are used in the learning update procedure described below.

### 3.2.1 Linear networks

As shown in Figure 3-1, between each pair of layers is a set of weights. Each weight matrix transforms the activations from the lower layer to the activations at the next layer. In a *linear network* (a network where *f* is a linear function) there is no need for a hidden layer because any function that can be computed with two linear transformations can, of course, be computed by a single linear transformation (i.e., a single weight matrix between two layers). Such two-layer networks are limited in the functions they can compute (Minsky and Papert, 1969).

### 3.2.2 Non-linear, 3-layer networks

Although two-layer networks are restricted in their computing power, non-linear, 3-layer networks are potentially much more powerful. This type of connectionist network is capable of developing patterns of activations on the hidden units in response to a learning algorithm. These hidden layer patterns can be viewed as *recodings* or *representations* of the network's input patterns. These representations, combined with a differentiable, non-linear activation function, allow a network to solve problems which are out of reach of linear networks. The next section examines how weighted connections can be altered in a network so that it may learn.

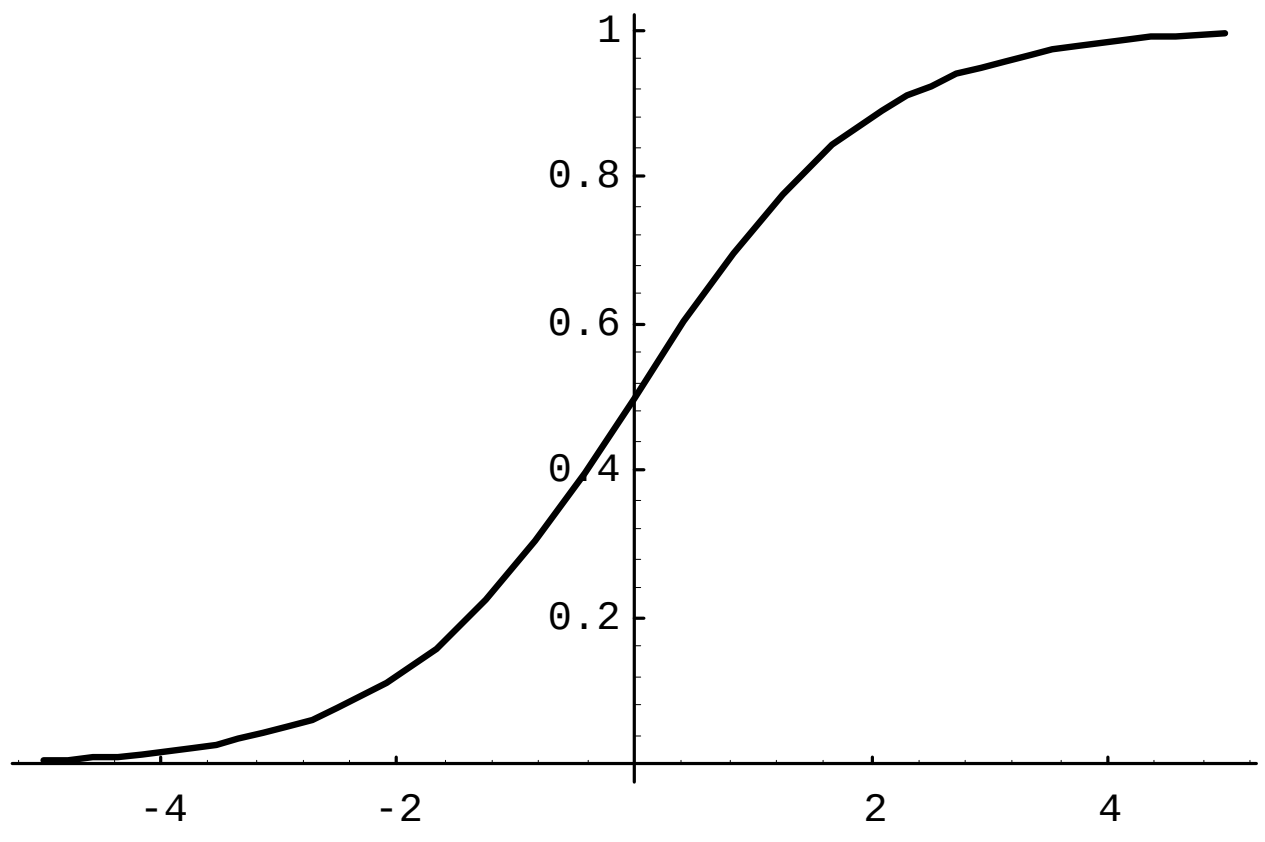

Figure 3-3. A simple logistic function. The x-axis shows the inputs into the function, and the y-axis the resulting squashed value between 0 and 1.

Table 3-1. A sample set of input/output training pairs

| | Input Pattern | Target Output Pattern |
|---|---|---|
| 1 | 0 0 0 | 0 0 1 |
| 2 | 0 0 1 | 0 1 0 |
| 3 | 0 1 0 | 0 1 1 |
| 4 | 0 1 1 | 1 0 0 |
| 5 | 1 0 0 | 1 0 1 |
| 6 | 1 0 1 | 1 1 0 |
| 7 | 1 1 0 | 1 1 1 |

## 3.3 Learning via Back-propagation

Before training, a connectionist network, such as the one described in the previous section, usually doesn't perform any useful function. The weights are generally initialized to random values, so the network produces meaningless output when activations are propagated from the input layer to the output layer. However, by using an iterative learning algorithm to adjust the weights, the network can be gradually coaxed into producing a desired output pattern for each input pattern in its training set.

There are many algorithms for adjusting weights, such as competitive learning (von der Malsburg, 1973; Fukushima, 1975; Grossberg, 1976), and genetic algorithms (Holland, 1975). Possibly the most widely used method is the *back-propagation of error* method, or simply 'backprop' (Rumelhart, Hinton, and Williams, 1986). Backprop is a gradient descent learning algorithm. In a nutshell, back-propagation calculates an error value for each output unit and adjusts the weights responsible for the error in such a way as to reduce it the next time around. The following is a more detailed sketch of the back-propagation algorithm and its use.

When training a network with backprop, each output unit has a desired output value called the *target*.[6] The difference between the target value and the actual output value produced by the propagation phase is called the *error*. The main function of the backprop algorithm is to correctly assign blame to the weights responsible for the error and to adjust them accordingly.

As an example, consider the patterns shown in Table 3-1. These patterns define the function ADD1 for 3-digit binary numbers. A network that correctly transforms the input patterns to their associated output patterns could be described as computing the binary ADD1 function. For instance, input pattern #1, the binary representation of zero, should produce the target output pattern {0 0 1}, the binary representation of one. Using the network shown in Figure 3-1, the input/output patterns shown in Table 3-1, and the backprop learning algorithm just outlined, the network can be trained to produce the desired target output patterns, thereby implementing the ADD1 function. A specific examination of this process follows.

Initially, all 12 of the weights in Figure 3-1 are set to small random values.[7] The back-propagation training procedure begins with the random selection of a pair of patterns, say pair #2. The first step is to load the input pattern onto the input layer. This is accomplished by setting each input unit's activation to the corresponding value in the input pattern. For pair #2, the activations of the input units would be set to {0 0 1}. Next, the activation is propagated through the network as previously described. At this point, there now exists a pattern of activation on the output layer, say {0.4 0.6 0.7}. As the desired target output values are {0 1 0} the associated error values are {-0.4 0.4 -0.7}, i.e., the difference between the target and the actual output. Those error values are then "back-propagated" through the network, and the weights are updated accordingly.[8] For a detailed explanation of the workings of back-propagation see (Rumelhart, Hinton, and Williams, 1986) or (Bechtel & Abrahamsen, 1991).

---

[6] Because analogies are divided into source and target components, I will refrain from calling the desired outputs of a network "targets" so as avoid confusion. In subsequent chapters I will refer to them as simple "desired outputs".

[7] Specifically, the weights should be small enough so that the net activation at any unit should initially be between 0 and 1.

The propagation of error and adjustment of weights for a single training pattern is called a *trial*. Following the first trial, another pair of patterns is selected, and the entire process is repeated. One such sweep through all training patterns is called an *epoch*. At the end of an epoch, the total error of all patterns is computed. If the total error is within an acceptable tolerance value, training stops, otherwise another epoch begins.

When the error falls within range of the stopping criterion, the network has successfully learned the training patterns. However, there are many possible ways in which the network may have learned the training patterns. The network might somehow have developed specific 'rules' for transforming each individual input pattern into its corresponding output pattern. If this were true, then the network would be unable to generalize. That is, the network would have found a way to produce the correct answer for each training pattern, but the solution would not carry over to new patterns it has not seen before. The generalizing ability of the network can be tested by placing an untrained input pattern on the input layer, propagating the activation through the network, and comparing the actual output with the expected output.

---

[8] To ensure convergence of the network to a stable, final set of weights, the updating of the weights will often be delayed until after all input/output pairs have been seen. This is called *batch-mode* training.

For instance, suppose that we had removed pair #4 from the training corpus prior to training. After training, we place the input pattern {0 1 1} (pattern #4) onto the input layer. If the output layer's activations are sufficiently close to {1 0 0} (the target output pattern), then the network can be said to have generalized, that is, to have recognized and taken advantage of regularities in the input patterns it has learned. Many variables may effect whether a network generalizes, including the number of training patterns, the number of units in the hidden layer, the input representations, and difficulty of the problem.

Sometimes, statistical regularities in the input patterns appear in a way that is difficult (or impossible) for a network to take advantage of. However, back-propagation networks with a hidden layer have the capacity to *recode* their inputs. By doing so, they are able to reformulate the problem and sidestep many of the limitations initially pointed out by Minsky and Papert (1969). Many types of problems can be solved in this manner, given that one knows (or can compute) a target value for each output unit. For example, 3-layer back-propagation networks have learned to convert text to phonemes (Sejnowski and Rosenberg, 1987), learned to form the past-tense of regular and irregular verbs (Rumelhart and McClellend, 1986), and learned to discover regularities hidden within relationships (Hinton, 1988), to name just a few.

## 3.4 Connectionist Representations

In order to examine connectionist representations, we need the foundation of some basic ideas and a set of common terms. One of the most basic concepts in connectionism is that of *pattern*. A pattern is an ordered set of activation values. In this discussion, we will use the word 'pattern' to mean a *static* set of activations, but a pattern could also be dynamic, existing through time.

A *representation* is a pattern that carries or conveys *meaning*. Meaning is a function that maps *entities* to representations. *Entities* are composed of *objects*, *attributes*, or *relations*. Objects, attributes, and relations are concepts used by the human designers to describe and perceive the world. A *representation scheme* is the inverse function of meaning; a representation scheme maps representations to entities. A *situation* is a set of entities. Although these terms may be defined circularly, together they form a consistent view of representations.

Let us now reexamine the ADD1 network using this terminology. The numbers 0 through 7 are the entities being represented. The number 2 is an entity represented by the pattern {0 1 0}, and the representation {0 1 0} means the number 2. As the patterns are constrained to consist of the activation values 0 and 1, the patterns are called *binary*.

Table 3-2. A localist representation scheme.

| Pattern # | Input Pattern | Target Output Pattern |
|---|---|---|
| 1 | 1 0 0 0 0 0 0 | 0 0 1 |
| 2 | 0 0 0 0 0 1 0 | 0 1 0 |
| 3 | 0 1 0 0 0 0 0 | 0 1 1 |
| 4 | 0 0 0 1 0 0 0 | 1 0 0 |
| 5 | 0 0 1 0 0 0 0 | 1 0 1 |
| 6 | 0 0 0 0 1 0 0 | 1 1 0 |
| 7 | 0 0 0 0 0 0 1 | 1 1 1 |

There are two schemes for encoding entities: *localist representation schemes*, and *distributed representation schemes*. A localist representation scheme is a function that maps all of the relevant entities from the world to a representation such that all of the representations of the relevant entities are orthogonal. As an example, consider a new representation scheme for the ADD1 network. Notice that the original representation scheme used to encode the numbers 0 through 6 produced non-orthogonal patterns (Table 3-1) and is, therefore, not a localist representation scheme. One possible localist representation scheme for the ADD1 network is shown in Table 3-2. A distributed representation scheme is any function that produces non-orthogonal representations for all relevant entities. There are many ways for a set of representations to be non-orthogonal; Table 3-1 is one set, and Table 3-3 is yet another quite different distributed representation scheme. Analogator will use only localist representations for input patterns.

Table 3-3. Another distributed representation scheme.

| | Input Pattern | Target Output Pattern |
|---|---|---|
| 1 | 0.1 0.5 | 0 0 1 |
| 2 | 0.4 0.3 | 0 1 0 |
| 3 | 0.3 0.7 | 0 1 1 |
| 4 | 0.6 0.9 | 1 0 0 |
| 5 | 0.7 0.8 | 1 0 1 |
| 6 | 0.5 0.4 | 1 1 0 |
| 7 | 0.2 0.1 | 1 1 1 |

## 3.5 Hidden Layer Patterns

As mentioned, a 3-layer connectionist network trained with backprop has the capability of non-linearly recoding each of the input patterns into non-arbitrary, distributed patterns of activation at each successive hidden layer. This process of learning to recode input patterns into intermediate patterns of activation spread across the hidden layers amounts to the development of distributed internal representations of the input information by the network itself. The ability of connectionist models to develop their own distributed internal representations is an extremely important property of this class of models.

## 3.6 Principal Component Analysis

Although much power is gained by having non-linear hidden layers, the networks lose perspicuity; analysis of how a network solves a particular problem can be very difficult. To get a better idea of how a network solves a particular problem, one can examine its hidden layer patterns as follows.

In a 3-layer network, each input representation produces a pattern on the hidden layer units on its way to the output layer. If there were a small number of units on the hidden layer (e.g., less than four), one could examine these activations by assigning each unit an axis in a graph. For each input representation, the hidden layer pattern would define a point in *hidden layer activation space*. Plotting each input pattern's position in hidden layer activation space can provide visualization hints as to how the network has solved a problem. For instance, one might see clusters of points. These clusters presumably reflect abstractions and categorizations discovered by the network and represented in the hidden layer patterns.

Plotting patterns in hidden layer activation space is only useful, however, when the number of the units in the hidden layer is three or less as humans are not very good at analyzing data in four or more dimensions. However, Principal Component Analysis (PCA) can provide an approximation of a multi-dimensional space in fewer dimensions. PCA works by finding dimensions that better describe the variance among a set of points in a multi-dimensional space than the original axes. For instance, consider the points on the left-hand side of Figure 1-2. When PCA is applied to the set of points, a vector is found that maximizes the differences between all points (the dashed arrow). This vector, called an *eigen-vector*, defines the first principal component. On the right-hand side of Figure 3-4, those same points are presented along a single dimension, the first principal component. Thus, the original *x*, *y* plot has been reduced to a single dimension that captures most of the variance of the original points. The PCA algorithm is repeated to find the dimension with the next most variance, and so on. Of course, the real usefulness of PCA comes from reducing a highly dimensional representation down to two or three principal components, as demonstrated in Chapter 5.

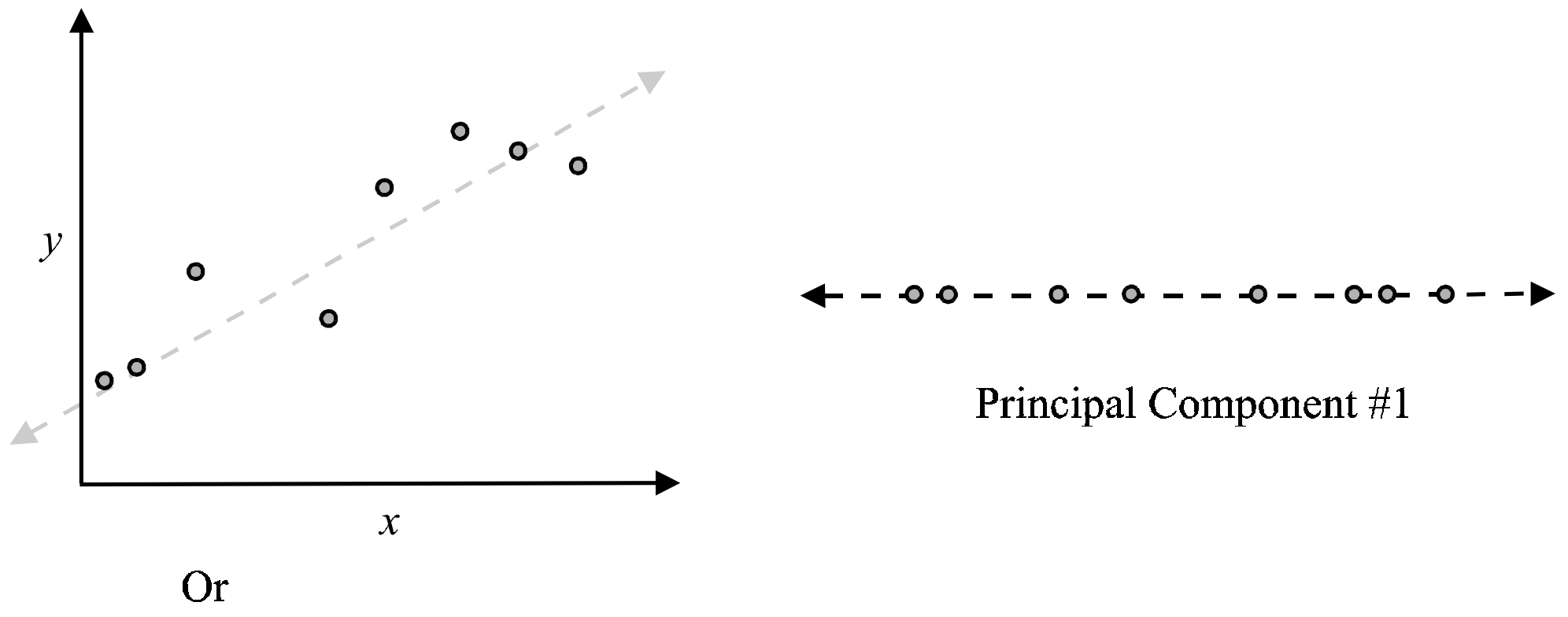

Figure 3-4. Principal Component Analysis. The dotted line accounts for almost all of the variability between the points.

## 3.7 Simple Recurrent Networks

Feed-forward networks work well for associating a single pattern to another. However, allowing cycles in the flow of activation in a network gives the network the ability to learn patterns in *sequences*. Networks that can learn patterns in sequences are useful for handling data in time, or variable length patterns. Analogator uses sequencing to allow hardware sharing between analogous components, as we will see in the next chapter.

To train a recurrent network to the true gradient using backprop, information must be kept about the flow of activation over a recurrent connection through time. Elman has proposed a short cut to the process and it has been found to work quite well (Elman, 1990). As this solution simplifies the calculations by approximating the true gradient descent, Elman calls this type of network a Simple Recurrent Network, or SRN.

SRNs eliminate the need to keep detailed histories about recurrent connections by adapting the standard feed-forward architecture to accommodate the recurrent flow of activation. Elman's method is described as follows. After the propagation and back-propagation phases, the activation patterns from the hidden layer are copied to a special bank of units on the input layer, called the *context bank*. At the next time step, input will flow from the regular input units and the context bank to be combined at the hidden layer units. Following the propagation phase and the backprop phase, the activations will again be copied back to the input layer, and so on after each step.

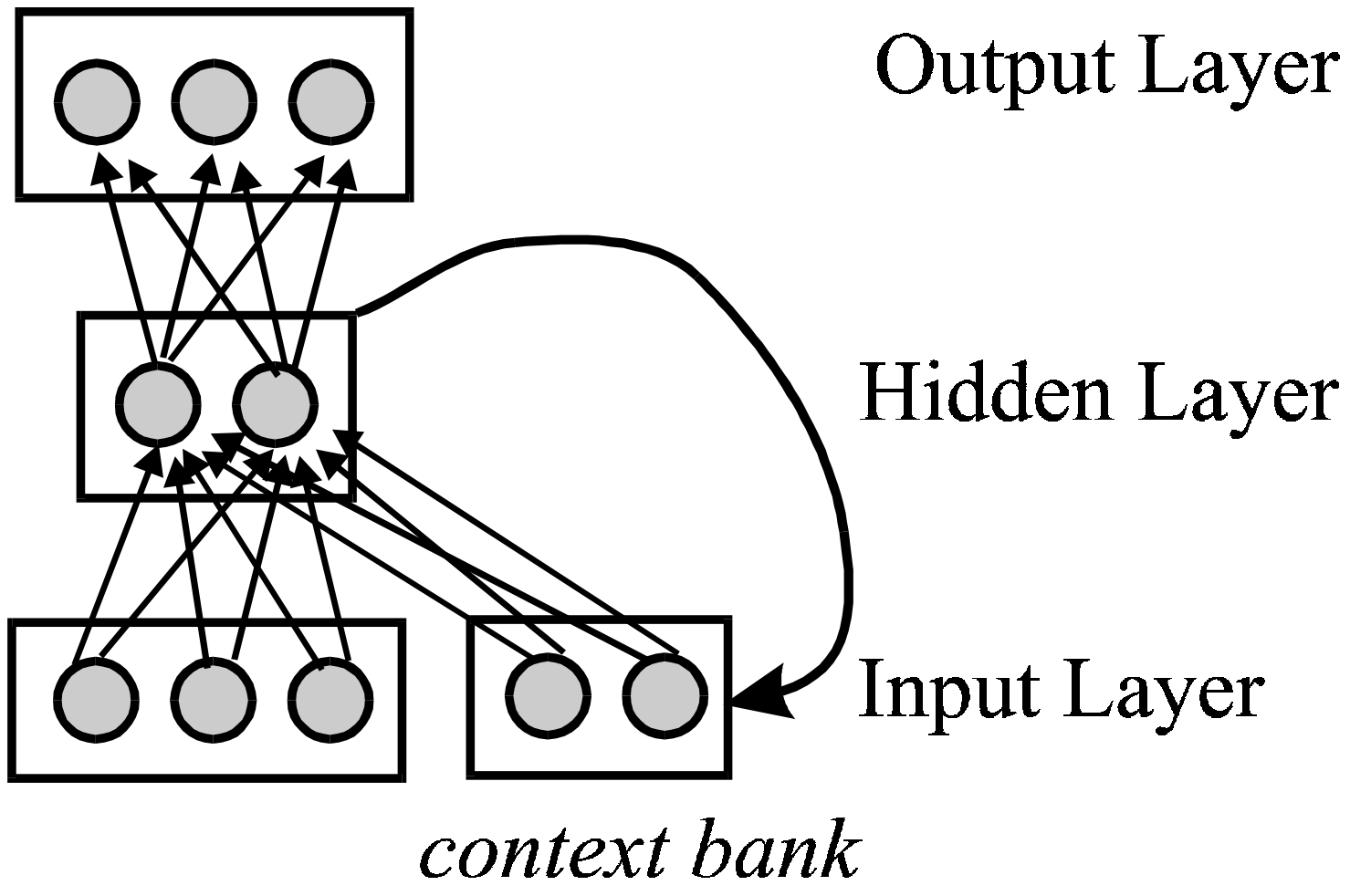

Figure 3-5. A Simple Recurrent Network (SRN).

This method has no guarantee of converging as it only approximates the true gradient descent, but it has been found to work quite well in practice. This has allowed researchers to work with feedback networks without the trouble of the more complex, and more time consuming, learning algorithms. SRNs have been applied to grammar learning (Elman, 1990), robot control (Meeden, 1994), and many other problems requiring recurrent processing. SRNs are related to more elaborate connectionist networks, such as Pollack's Recursive Auto-Associative Memory (RAAM; Pollack, 1988) (see also Blank, Meeden, and Marshall, 1992, for a thorough examination of sequential RAAMs, a close cousin of the SRN).

## 3.8 Tensor products

Many methods have been developed in the past few years for representing structured information in a connectionist network. This has, in part, been done to meet the challenges posed by Fodor and Pylyshyn (1988). They charged that connectionist models were incapable of adequately representing structure. One early answer to their challenge was Smolensky's *tensor product variable binding* (Smolensky, 1991). Let us briefly examine the problem of representing structure in connectionist networks.

We have seen how a pattern may represent an entity, but what must one do to represent two entities? One answer would be to simply assign a new pattern to represent the pair. But this quickly creates an explosion of patterns as we would need a unique one for each combination of entities. Another answer might be to concatenate the two patterns, one after another. However, this creates a pattern that is twice as big as the others and would grow longer with each additional entity. What is desired is a fixed-width representation capable of representing one or more entities and the relationships between them. Also, we would like a system with a graceful degradation in performance as the number of entities increases. Although there now exist many such techniques for representing structured representations, the simplest is probably Smolensky's tensor product method.

Smolensky's method first requires that structures be broken down into *roles* and *fillers*. A role is a named position; a filler is the specific object or thing that fulfills a particular role. For example, 'president' is a role currently filled by the filler 'Bill Clinton'. Roles and fillers are a very common form of representation used by linguists and computer scientists.[9] The tensor product method is a mathematical procedure for associating (also called *binding*) a role with its filler.

Specifically, Smolensky's tensor product representation is a distributed set of activations formed by taking the outer product of a role pattern with a filler pattern (Smolensky, 1990) (see Figure 3-6). In this methodology, a role is an entity, as is the role's filler (i.e., they both have patterns). The matrix of values formed by calculating their outer product represents the *binding* of these two entities. For instance, in Figure 3-6 the role vector is represented by the activation values {0.9 0.1 0.5 0.1} and the filler is represented by {0.1 0.5 0.9 0.1}. The matrix produced by the outer product represents the filler in that particular role.

---

[9] Computer scientists also call roles *variables*, and their fillers *values*.

Multiple role/filler pairs may be represented by adding outer product-formed matrices. In this manner, one may represent complex structures, such as trees or lists. For instance, consider the sentence “Mary gave the book to John.” One possible set of role/filler pairs is: *GIVER / mary*, *RECEIVER / john*, and *OBJECT-OF-TRANSFER / book*, where roles are given labels in uppercase letters, and their fillers are shown in lowercase italics. Each filler is bound to its role via an outer product of the two associated patterns. Finally, all three outer products are summed to produce a single matrix representing the entire sentence or structure (see Figure 3-7). The 3 by 3 matrix to the far right of Figure 3-7 represents the full sentence, “Mary gave the book to John.” The number of vectors used in creating a tensor determines its dimension, which is also called its *rank*. Therefore, Figure 3-7 shows the creation of a rank-2 tensor.

Notice that the tensor product method creates representations that are larger in rank than their original vectors (i.e., the resulting matrices are substantially larger than their originating role and filler patterns). This problem has been addressed by Plate’s Holographic Reduced Representations (HRR) and Kanerva’s spattercode (Plate, 1991; Kanerva, 1996). However, their basic methodology is not substantially different from

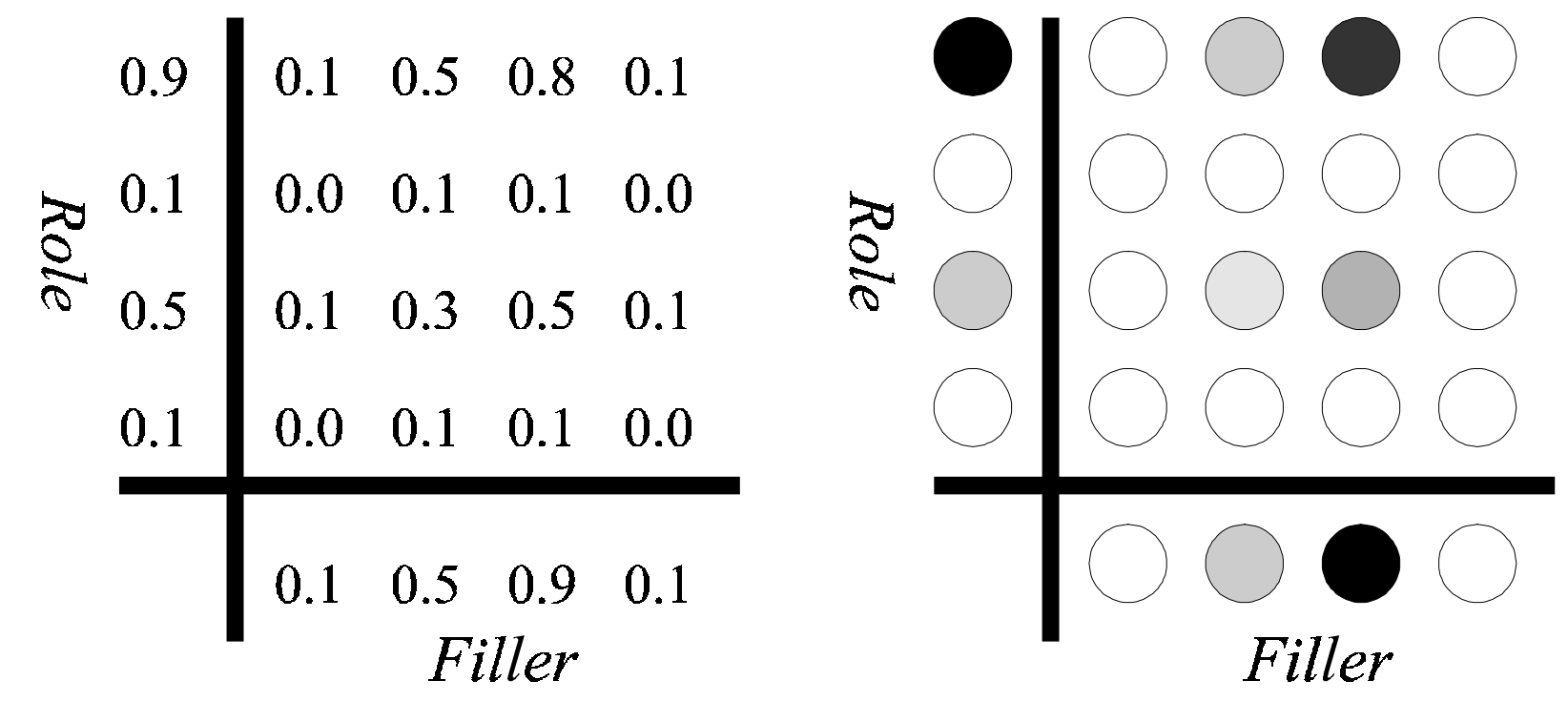

Figure 3-6. Two methods of depicting a tensor product role/filler binding.

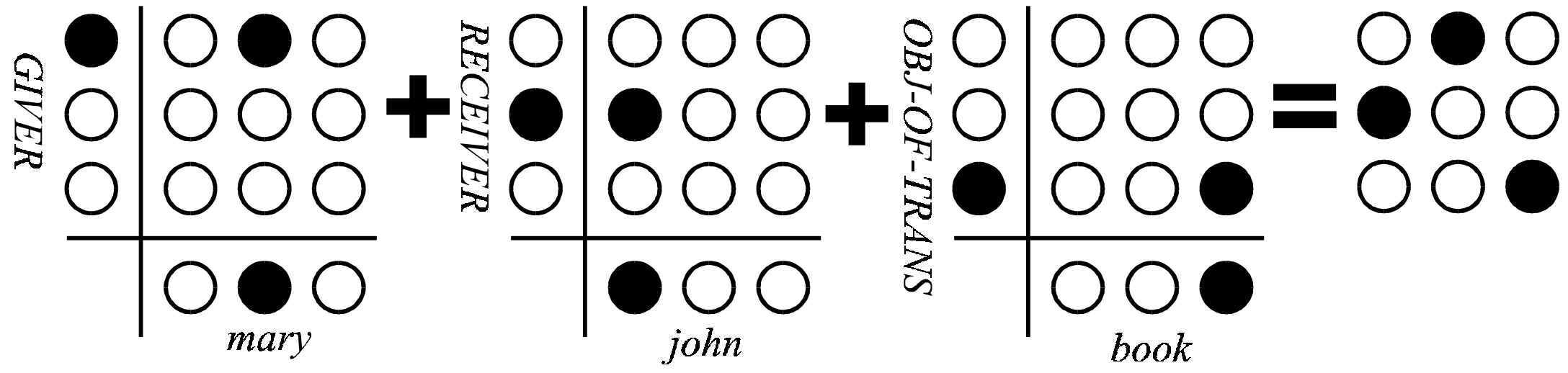

Figure 3-7. Tensor product representation of "Mary gave the book to John."

Smolensky's tensor product variable binding. We will further compare these methods in Chapter 6.

In this chapter we have examined the connectionist foundations of Analogator, namely: the basic architecture of connectionist networks, a learning algorithm called back-propagation, the development of hidden layer patterns and their analysis, a network architecture for simple recurrent networks, and role-filler binding via tensor product algebra.

We now turn to see how these basic building blocks are used to create Analogator.

# 4 The Analogator Model

*Spoon feeding in the long run teaches us nothing but the shape of the spoon.*

-E. M. Forster

As described in Chapter 1, the Analogator project was created to study the integration of learning and analogy-making in various domains. A key distinction between Analogator and other models of analogy-making that incorporate learning is that Analogator focuses on *learning how to make analogies* rather than *learning by using* *analogical reasoning*. The difference is that other models have an analogy-making mechanism built in, while Analogator learns the analogy-making mechanism itself. This chapter will examine the Analogator model and how it learns about analogies.

## 4.1 The Representation Problem

The question of how to represent things to a computer program lies at the core of many issues in AI and cognitive science. There have been many heated debates over the form representations should—and should not—take. This is an especially important topic in analogy-making where the exact form of initial representations is critical.

There are, in general, two ways entities (i.e., relations, objects, and attributes) may be represented: *implicitly* and *explicitly*. As we saw in Chapter 3, an entity may be represented by a specific pattern. Such a representational system encodes entities explicitly. An implicit representation system, on the other hand, is one in which entities do not have specific patterns; there is not a fixed set of patterns that encode the meaning. Rather, the presence or absence of an implicitly represented entity must be inferred from the interaction of other patterns. We shall now examine examples of implicit and explicit representations.

Specifically, explicit representations are highly organized, and exactly expressed. Most representations that computer scientists handle are of exactly this type. If they were not, in fact, they would be of little value. Traditional computer science programs work *because* of explicitly structured representations. However, programs that use explicitly structured representations are notoriously brittle. If, for instance, one misplaces a link in a linked-list, the remainder of the list is lost. Computer programs are designed to operate on precisely specified data; this is at once their strength and their Achilles' heel.

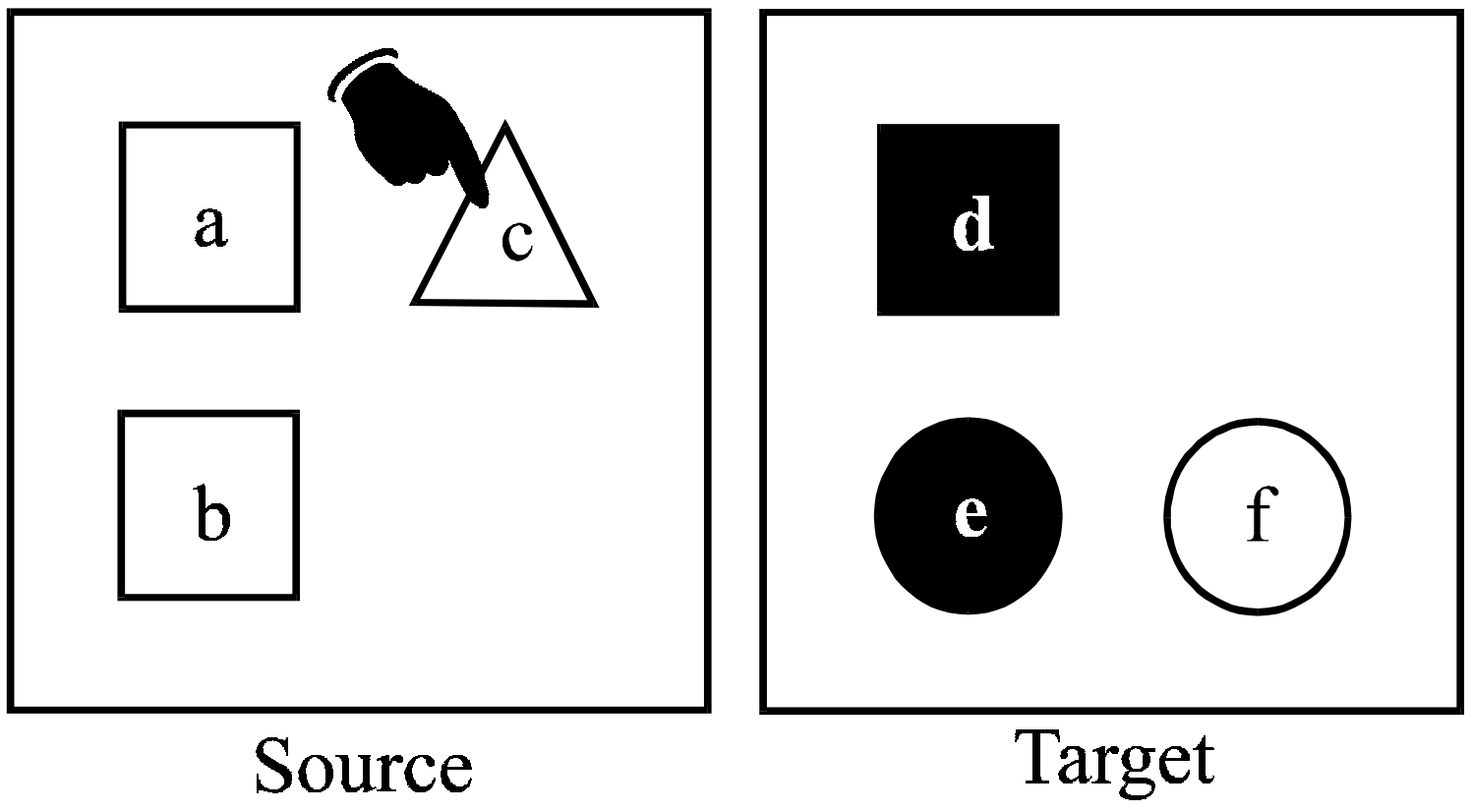

Figure 4-1. An Analogator problem with reference labels.

Traditional analogy-making programs, like those in computer science in general, have used explicitly-structured representations, and, therefore, have all of the associated strengths and weaknesses. Hofstadter and his research group have argued convincingly that analogy-making cannot be adequately modeled by using explicitly-structured representations spoon-fed to an analogy-making engine (Chalmers, French, Hofstadter, 1995; Hofstadter *et al.*, 1995). To examine their complaints, consider the Analogator problem in Figure 4-1. Recall that the goal is to find the object in the target that is the same, in some sense, as the selected object in the source.

One may notice that the source scene has a *pair of squares* (labeled **a** and **b**) next to a *triangle* (labeled **c**). Likewise, the target scene could be perceived as a *pair of black objects* (labeled **d** and **e**) next to a *white object* (labeled **f**). If one perceived the scenes in such a way, it is clear that one should choose **f** as the analogous object. On the other hand, the target scene could also be seen as a *pair of circles* (**e** and **f**) next to a *square* (**d**). This makes it equally obvious the analogous object is the *black square,* and one should select **d**. These two conflicting ways of perceiving the same target scene are shown with explicit representations in Figure 4-2.

If one were to choose one of the two forms of representing the target scene and use that as a starting point in an "analogy-making engine", it is clear the representation chosen would bias the program. In fact, by choosing a representation, Chalmers *et al.* (1995) argue that the researcher has inserted too much of their interpretation of the scene into the analogy-making system.

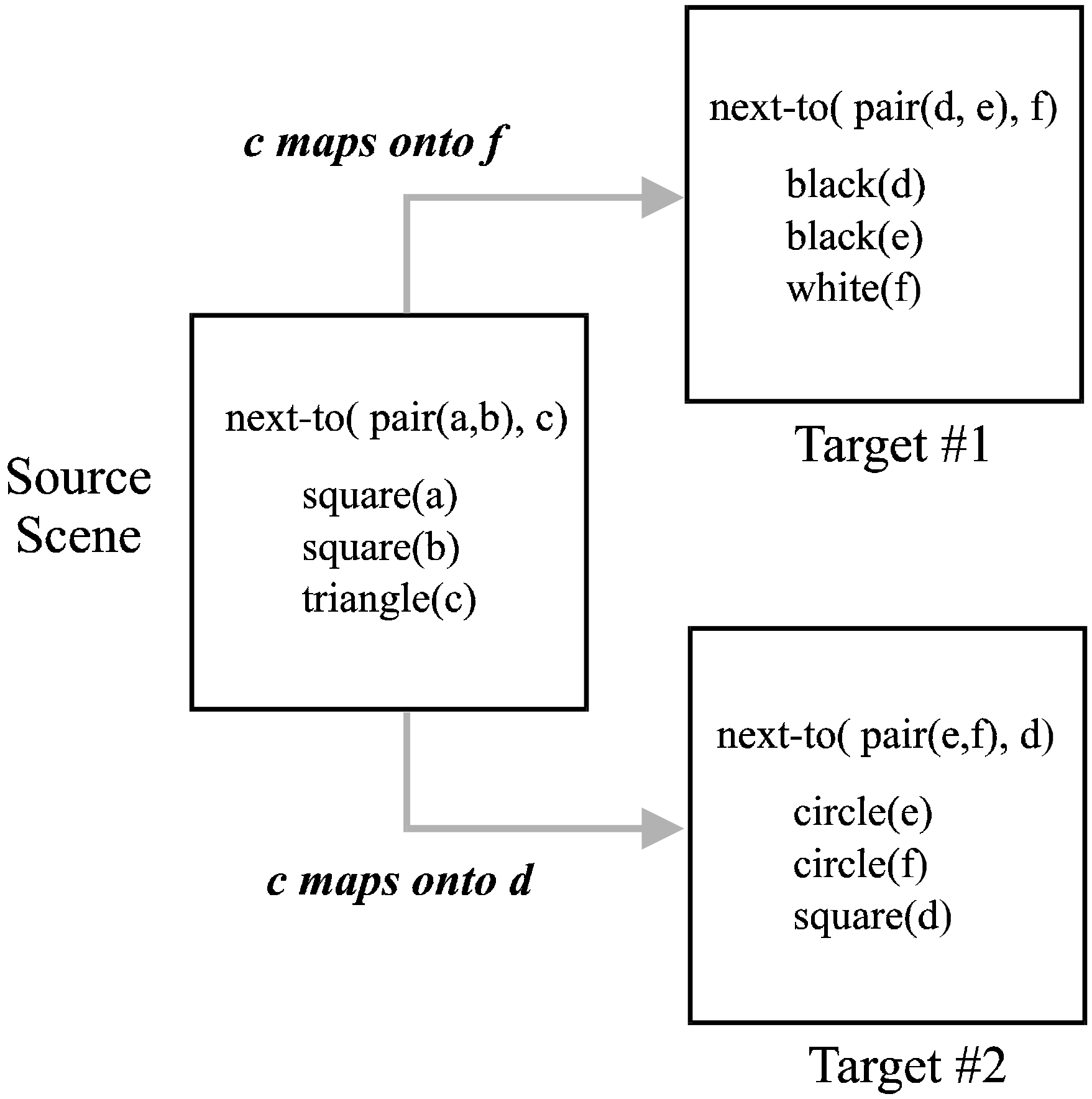

Figure 4-2. Two ways of explicitly representing the target scene of Figure 4-1. If the target scene were perceived as that of Target #1, then the analogous object would be the white object (f). If the target scene were perceived as Target #2, then the analogous object would be the square object (d).

This simple example shows how the form of representations used may control the outcome of the analogy-making process. That is, creating explicit structures dictates what kinds of correspondences can be made, and, therefore, determines what analogies will be made. Hofstadter and colleagues have argued that the *perception* of a situation is an important part of the analogy-making problem, and should be included in any analogy-making model (Mitchell, 1993; Hofstadter *et al.*, 1995). Their solution has been to design a program capable of building-up explicit representations in response to top-down and bottom-up contextual pressures guided by perception. The realization that perception is a part of analogy-making has been one of the most important contributions of cognitive science to the understanding of making analogies.

A model that builds up structures in response to perceptual and conceptual pressures could construct either of the targets of Figure 4-2. Rather than using perception to build-up explicit structures, Analogator learns to perceive relationships between implicitly structured representations. These two techniques are actually very similar in motivation and style of processing. However, they are also very different in other respects. A comparison of these two variations is presented in Chapter 6.

As learning is the focus of the Analogator project, one might ask, “Why start with a representation at all? Why not have the system learn to make analogies based on, say, low-level visual stimuli?” I admit to finding this idea attractive. Brooks has proposed a similar notion which he calls “intelligence without representations” (Brooks, 1991). One might debate whether his robots are actually ‘representation-free,’ but his point is that an artificial system can accomplish quite a bit without the traditional explicit, *knowledge-based* representations. However, starting with raw visual input is currently too tough a problem. This is a hard “chicken-or-the-egg” dilemma; low-level processing seems to require top-down guidance, yet the top-down guidance needs to be learned from bottom-up processing. The answer, of course, is that they should develop together, and that is the tough problem. For the moment, though, let us consider a photograph as a representation.

Imagine a digitized photograph of a dog sitting in its doghouse. In no sense are ‘dog’, ‘doghouse’ or ‘inside’ explicitly represented in the photograph. In fact, some might argue that a photograph cannot be considered a representation at all. Philosophical arguments aside, a photograph does not contain any information *explicitly*, except for specific values of brightness and hue at specific points. Instead, all of the information about the dog, its doghouse, and their relationships to each other, is *implicitly* contained in the pixel values. In practice, even if photographs could be considered representations, they are not very useful ones. This is because photographs do not make use of category labels. It nearly goes without saying that category labels are necessary in a representation. On the other hand, explicitly-structured representations contain nothing but category labels. Recall that explicitly structured representations are made up of labels that are created in a given context, and analogy-making requires different category labels in different contexts.

The solution used in Analogator is a combination of implicit representations and explicit category labels. Specifically, objects are represented by explicit patterns, while the relationships between them remain implicit. To elucidate this process, we will examine one implicit representational scheme.

## 4.2 Implicitly Structured Representations

Recall from Chapter 3 that to bind a role to a filler using the tensor product mechanism, the outer product between the vector representing the role and the vector representing the filler is computed. For example, to represent the sentence “Mary gave John the book,” one would bind the representation of *john* to the representation of the role *GIVER*, *mary* to the role *RECEIVER*, and *book* to the role *OBJECT-OF-TRANSFER*. Also recall that to represent the sentence “Mary gave John the book” one need merely component-wise add the three previously created tensor products. This process creates a rank-2 tensor that is a distributed representation of the entire sentence.

If one wished to represent the sentence “The square is over the circle” using tensors, it is only slightly more complicated. The method used by Halford *et al*. (1994) is to compute a rank-3 tensor as follows. First, a vector is created to represent each object and relation (one each for *square*, *circle*, and the relation *OVER*). The outer product is computed between the two vectors representing *square* and *circle*. This creates a rank-2 tensor. Next, the outer product is computed between that rank-2 tensor and a vector pattern representing *OVER*. This creates a tensor of rank-3 which represents the relationship *OVER(square, circle).* As this is a connectionist representation, it has many nice properties; however, it also has problems if used to make analogies.

Forming a representation based on a labeled relation, such as *OVER*, creates an explicitly structured representation. Much research has been directed toward building such connectionist representations capable of encoding explicit structures, such as trees (see, for instance, Pollack, 1988; Plate, 1991; Kanerva, 1995; Halford *et al*., 1994; and Hinton, 1991). These connectionist methods for encoding structure are very interesting; however, I do not believe that the representations they produce will benefit models of analogy-making because they result in explicitly labeled structures (for an alternative opinion about their usefulness in making analogies, see Gentner, 1993).

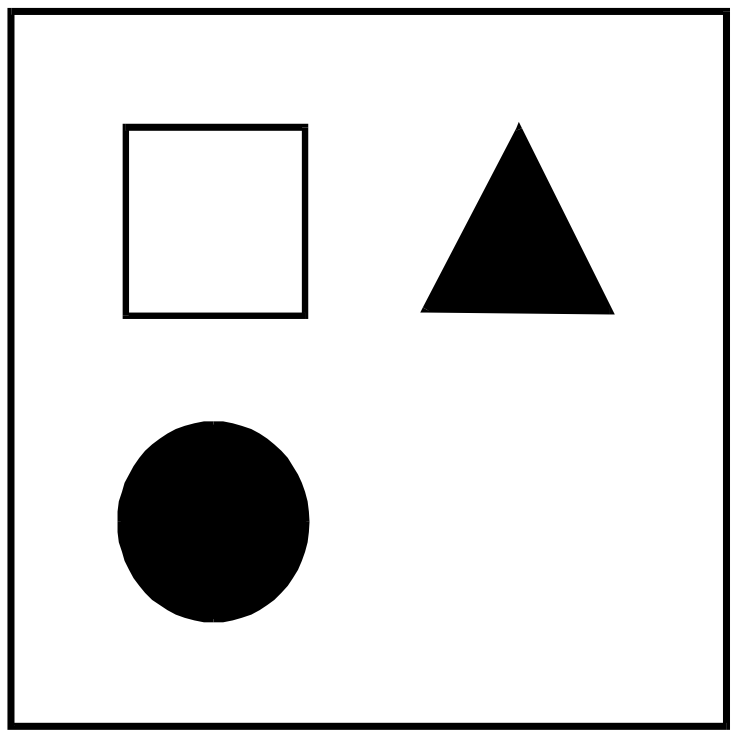

Figure 4-3. A sample Analogator scene.

In any case, the tensor product methodology can be altered to avoid the creation of an explicitly named relation. This is accomplished by replacing the notion of 'role' with a representation of 'location.' To examine this technique, consider Figure 4-3. Imagine that we wish to represent this scene as an implicitly structured representation. The first step is to break each object into two parts: the object's *attributes*, and the object's *location*. Each object's location part shall be called its *where-component*, and each object's attribute part shall be called its *what-component*. Each object's where-component represents its location in the scene, and each object's what-component represents the object's attributes (e.g., *black*, and *circle*). There is much evidence to suggest that human brains (as well as the brains of monkeys) create representations based on this "what" and "where" distinction (see, for instance, Landau and Jackendoff, 1993; Kossyln *et al.*, 1989; Ungerleider and Mishkin, 1982; Schneider, 1969). However, the details of how "what" and "where" are encoded and bound to each other in the brain is not yet known.

To represent the *white square* in Figure 4-3, we must first create the appropriate what- and where-component vectors. In this example, let us suppose that we are representing a scene by a 7 x 7 matrix. An object's where-component is composed of activation values indicating where the object is located in the scene. Furthermore, let us assume that we are representing an object's attributes with 5 values, 2 each for the color (one for *black*, one for *white*) and 3 for shape (one each for *square*, *circle* and *triangle*). An object's what-component will be a set of attributes: {*black*, *white*, *square*, *circle*, *triangle*}. Each attribute will be encoded as a localist representation where 1 means true, and 0 means false (see Chapter 3 for details of localist encoding schemes). Therefore, a *white square* is represented by {0 1 1 0 0}.

On the left side of Figure 4-4 is the where-component vector representing the location of the white square from Figure 4-3. Notice that the where-component has black-colored circles to indicate the area in the scene that the object occupies, namely in the upper left-hand corner. Running across the bottom of Figure 4-4, one will find the what-component vector representing the object's attributes. Taking the what-component and performing an outer product with the where-component produces a tensor of rank-2 as shown in Figure 4-4.[10] This representation is made up of five "planes," one for each of the five what-component units. This method of representing objects in a scene use location to bind the attributes. That is, 'location' is the dimension that ties the attributes *white* and *square* to a specific object.

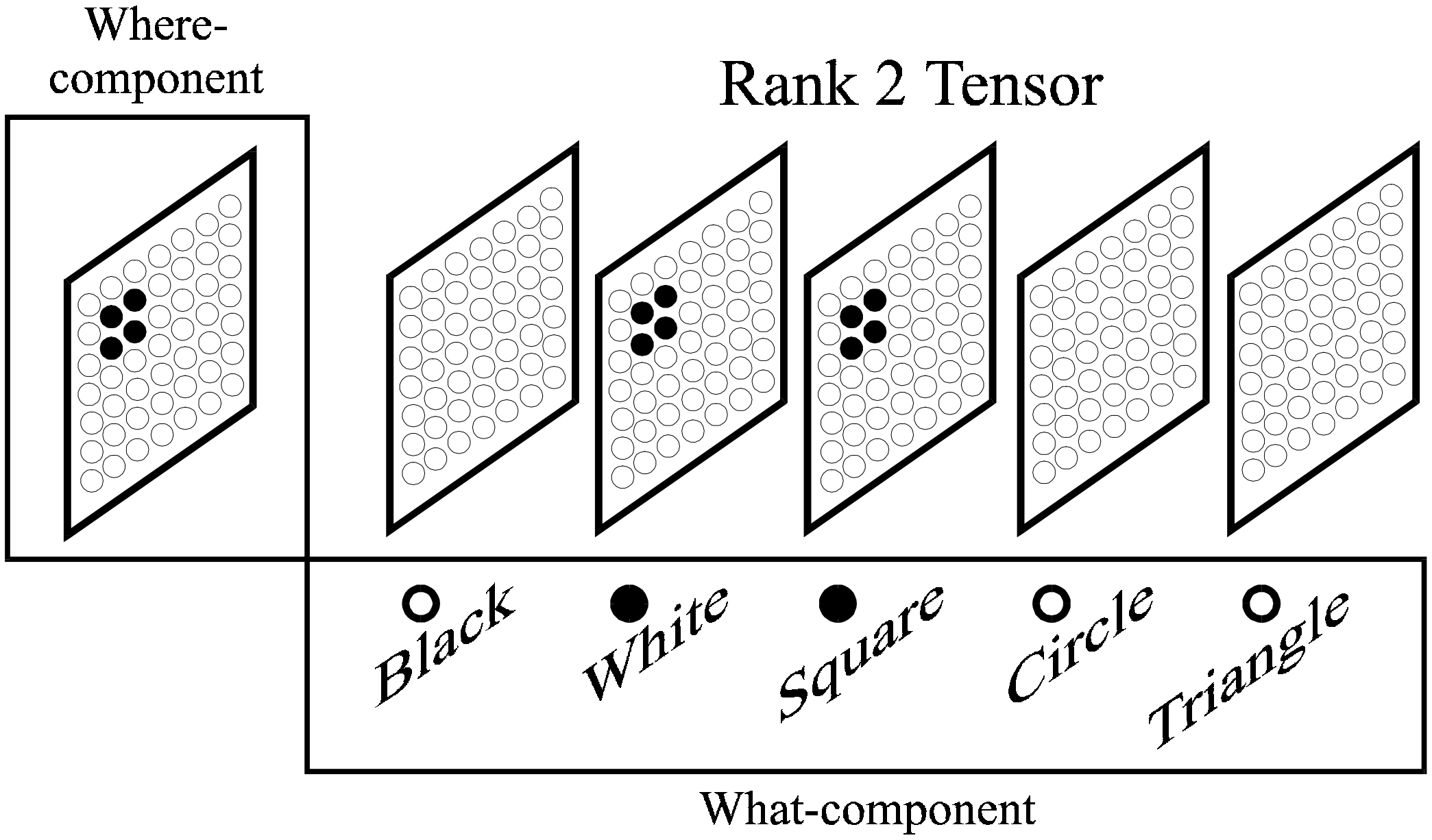

Figure 4-4. The creation of the rank-2 tensor. The resulting representation consists of a 7 x 7 x 5 matrix of values.

---

[10] The where-component vector is actually a 2-dimensional matrix representing the *x* and *y* axes of the scene; however, it will be considered a vector of rank 1 for the tensor product computations.

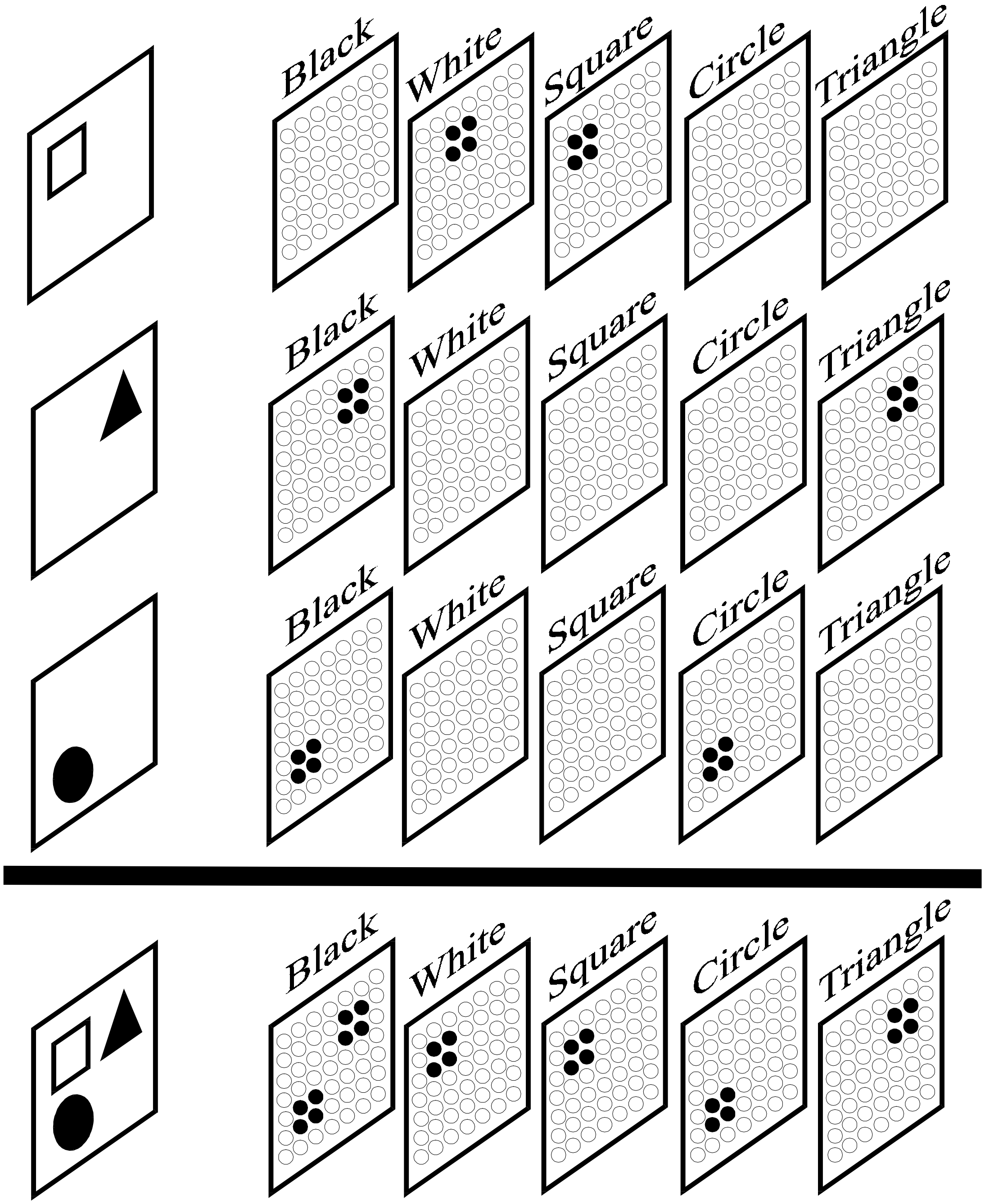

Figure 4-5. Using the tensor product method to represent attributes. The bottom row reflects a representation encoding all of the objects in their appropriate position.

Imagine that we have now created such representations for each of the objects in Figure 4-3. The final step is to create one representation that encodes the entire scene. This is done by component-wise adding the three rank-2 tensor product representations, as shown in Figure 4-5. Notice that attributes are explicitly represented. On the other hand, the relation *OVER(square, circle)* can be inferred from the representation but it is not explicitly encoded. In fact, all relations between the objects are implicit. This is similar to a representation used by Feldman (1985).

This representation scheme breaks down if the locations are non-orthogonal. However, if the scenes are confined to non-overlapping objects, then it is guaranteed that the where-component for each object's location is completely orthogonal to that of all other possible objects in the scene.

Because the notion of an *iconic representation* is appealing, researchers have used the term before, although without a specific implementation (see Harnad, 1987; also Barsalou's related notion of a "perceptual symbol," 1992). The representation described above captures many of the qualities that an iconic representation should have. It is iconic in that the values representing an object's attributes are localized to particular regions of the representation. It also preserves information about spatial relations between objects without committing to a single interpretation. It is interesting to contrast the representation system described above with Harnad's multi-representational system (1987). In order to explain same-difference judgments, as well as categorical tasks, his theory requires three separate representation schemes. Although Harnad has not specifically discussed the issue of representing location, he has discussed a multi-representational scheme that has representations which are purely iconic and others that are purely categorical, whereas I have described a single hybrid representational scheme.

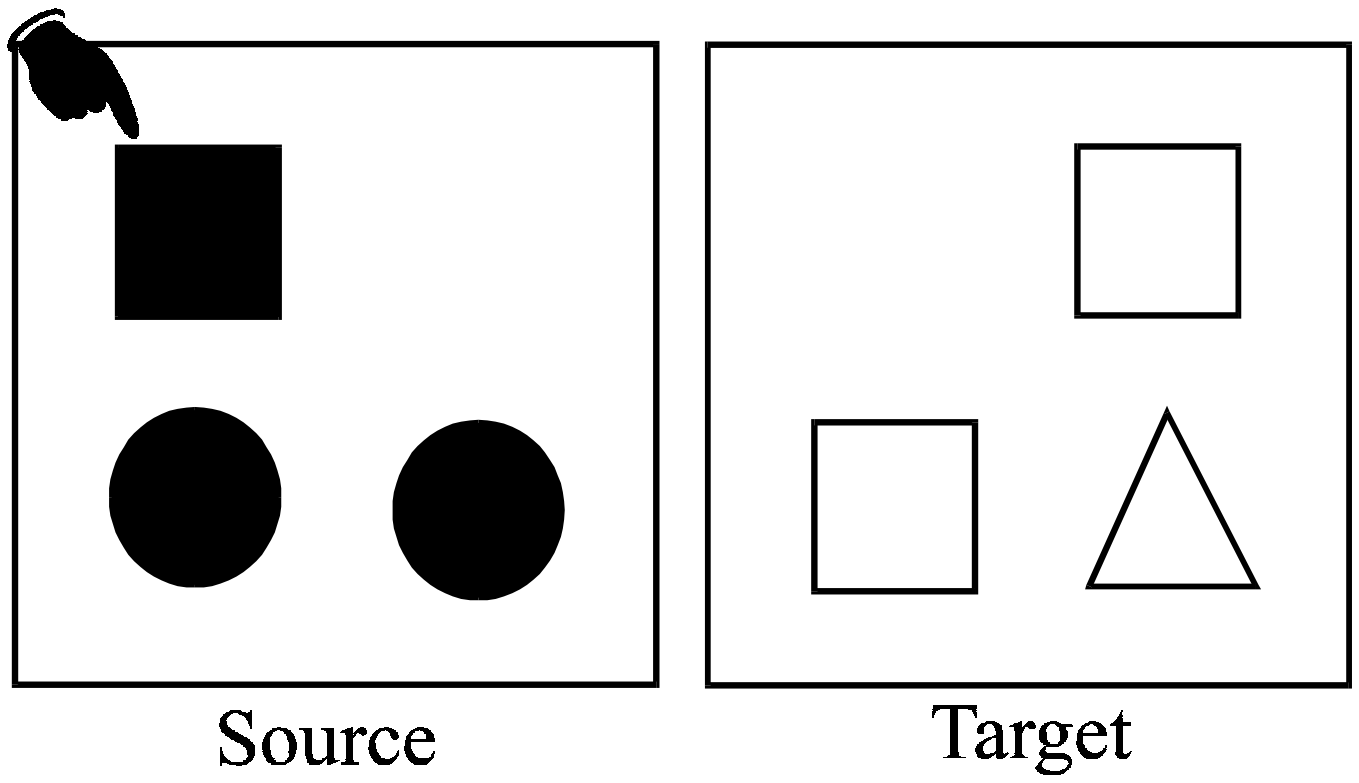

Figure 4-6. A sample Analogator problem from the geometric-spatial domain.

In summary, the representational scheme described above provides a method to encode a wide variety of scenes without explicitly representing relations. We now turn toward the task of using these representations in making analogies.

## 4.3 Network architecture and Training Procedure

Analogator's network architecture is a simple recurrent network (SRN) as described in Chapter 3. Recall that the SRN is a feed-forward network trained via backprop (or some other gradient descent mechanism) that simulates recursive connections by allowing activations to be copied from the hidden layer back to the input layer. We will now examine the details of the training process.

The *recurrent figure-ground associating procedure* uses the SRN architecture in the following 2-step manner. To make this concrete, let us use Figure 4-6 as an example problem. In the first step, a representation of the source scene (S) and the selected part are placed onto the input layer (see Figure 4-7). The selected section is that part of the source scene that is being pointed to, or, in other words, is the figure of the scene (SF). The network is then trained via backprop to segregate the figure and the ground into two banks on the output layer. In a manner similar to that of auto-associative training, this step only requires the network to reproduce the selected object on the figure bank (SF) and the rest in the ground bank (SG) of the output layer. The ground is defined to be everything but the figure in the scene. Notice that as the activations flow from the input to the output layer, activations are produced on the hidden layer ($H_t$). One way to think of what the network is doing on this step is: "in the context of a pointing finger, the network 'sees' the object that is being pointed to as the figure, and everything else as the ground."

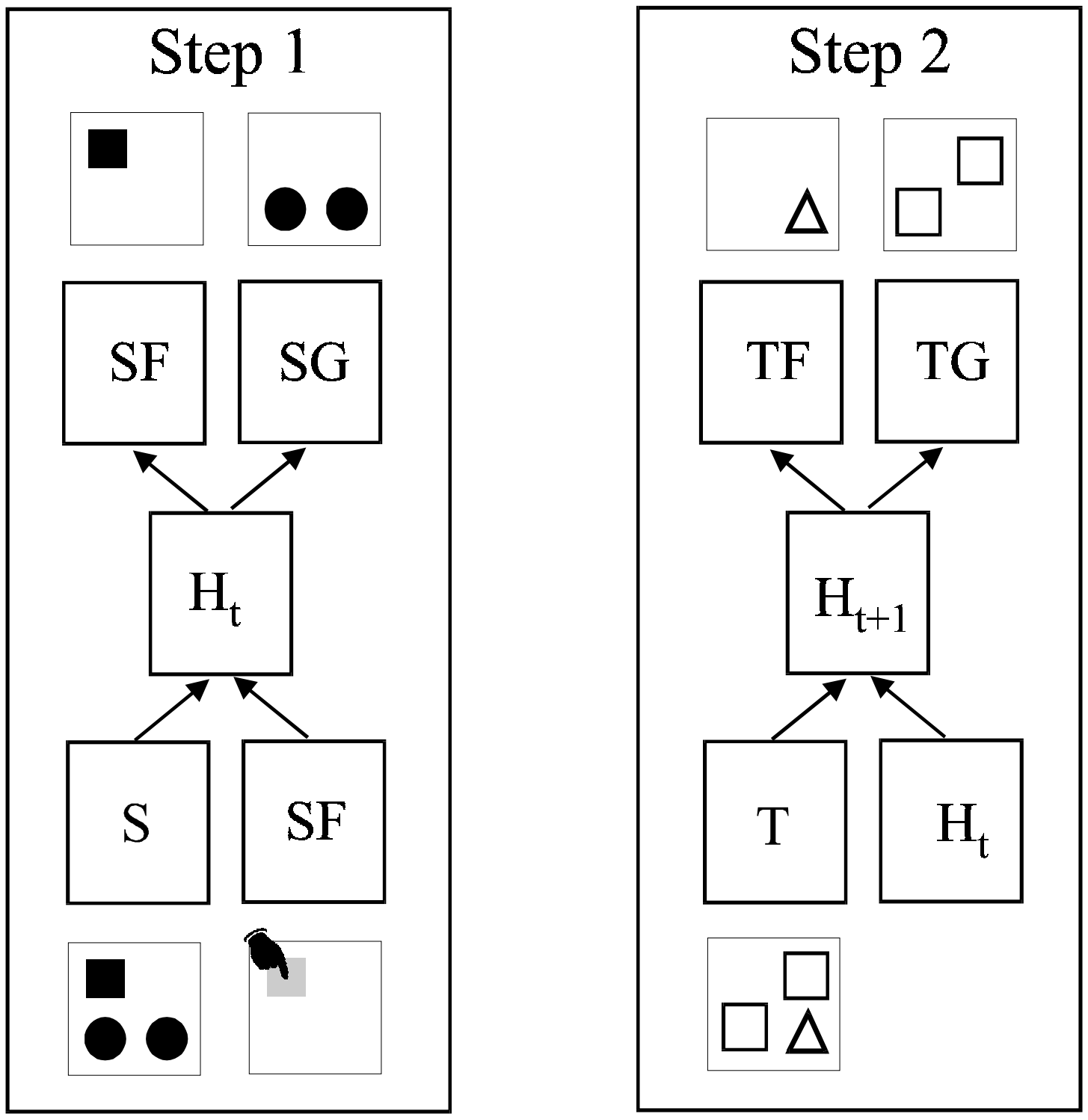

Figure 4-7. The recurrent figure-ground associating procedure.

On the second step, the entire representation of the target scene is placed onto the input layer into the same units that held the source scene from step 1 (T). There is no "pointing finger" here; instead, we will use a distributed representation of the source scene and its "context." We can accomplish this by copying the hidden layer activations from the previous time step ($H_t$) to the input layer. The network is now trained to produce the analogous figure and ground on the output layer (TF and TG). Of course, this time we wish for the target scene's figure and ground to emerge on the output layer. As this is a supervised learning task, the network is told what the correct answer should be during training. One way to think of what the network is doing on this step is: "in the context of the source scene and a figure in the source, the network 'sees' the analogous object of the target scene as the figure, and everything else as the ground."

### 4.3.1 A detailed example

To examine this process more carefully, let us go through a more detailed example using implicit representations and the training procedure. First, consider that we have already constructed implicitly structured representations of the source and target scenes of Figure 4-6. In step 1a, we simply place the activations of the source scene's representation onto the appropriate bank of the input layer. In addition, we also activate the region as indicated by the pointing hand on the context bank of the input layer (see Figure 4-8). The context bank is a 4 x 4 set of units. At this point, we will use the context bank to indicate the focus of attention, namely the area of pointing. In step 1b, the activations on the input layer are propagated through the network as described in Chapter 3. The desired output is a representation of the figure and ground of the source scene. As location uniquely identifies an object, both the figure and the ground are represented by 4 x 4 units on the output layer. In this manner, the network can identify an object as the figure on the output layer by activating the units representing the object's location. The same is true for all objects considered to be the ground. Finally, error is backpropagated through the network, and the appropriate weights are adjusted.

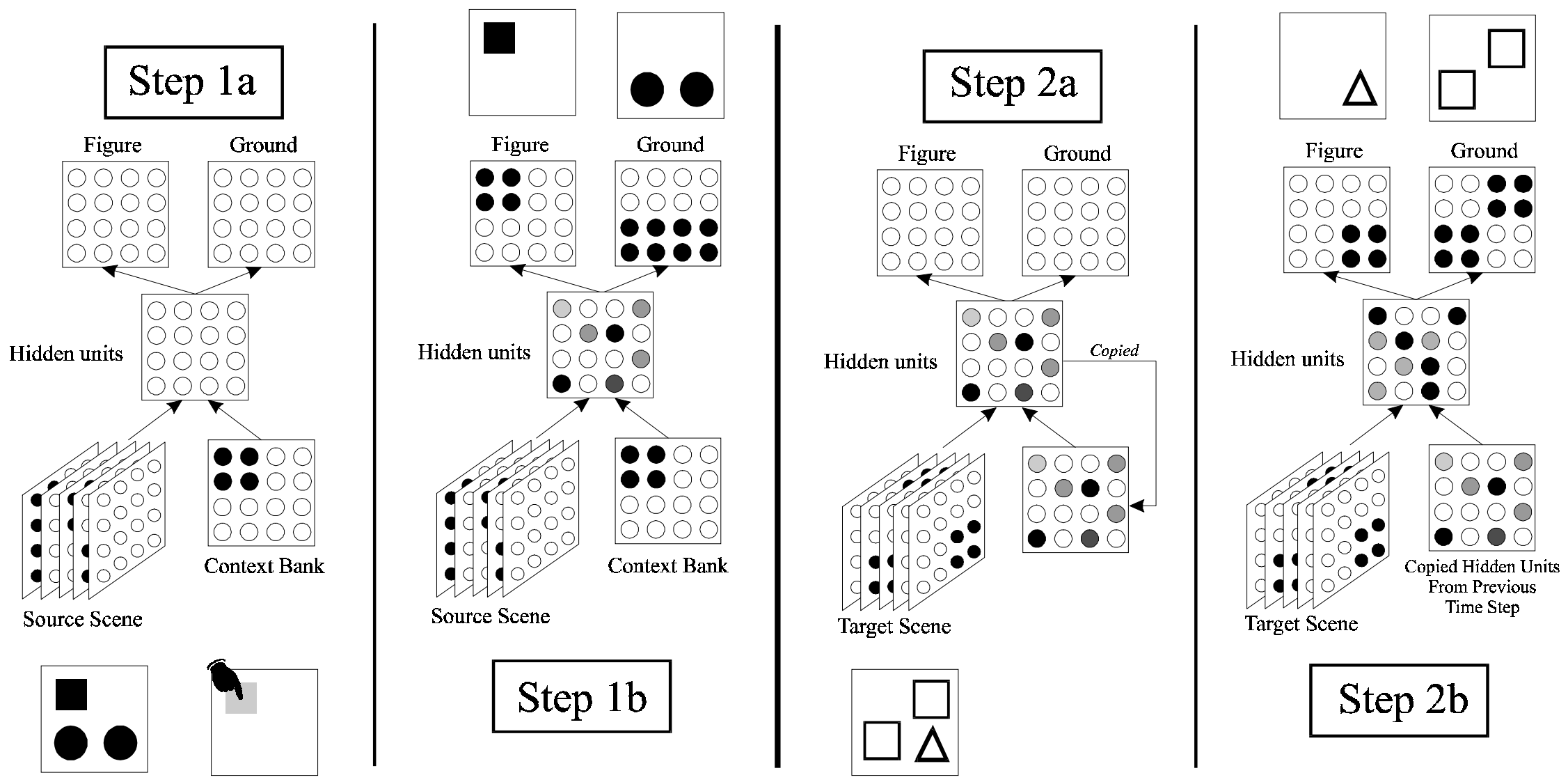

Figure 4-8. The recurrent figure-ground training procedure using iconic representations. Steps 1b and 2b require that the network produce on the output layer the locations of the figure and ground objects of the source and target scenes, respectively.

Step 2a begins with the placement of the target scene's implicitly structured representation onto the input layer. The 4 x 4 bank that held the location of the figure of the source in step 1a is loaded with the hidden layer activations from step 1b. It is hoped that the hidden layer activations contain (in distributed form) information about the source scene and its figure. The previous hidden layer activations are supplied to the input layer to provide context. Notice that for the copy to work, the hidden layer and the context layer must be the same size. Step 2b propagates the activations through the hidden layer and out to the output layer. Like step 1b, the figure and ground of the target are identified by their locations in the scene.

### 4.3.2 Discussion

This architecture and training procedure have the following desirable features. Instead of having a bank of units for the source and a separate bank for the target representation, they share the same connections, thereby making generalization easier. That is, correlations learned on one set of weights do not have to be learned on another. It makes perfect sense for the source and the target scenes to use the same weights, as they act as analogous parts of the analogy problem. Likewise, the figure of the source and the recurrent activations also share units. They, too, fill analogous roles in the training of the network: looking at the "source scene" and the "object being pointed to" is analogous to looking at a "target scene" and "remembering the source scene plus what was pointed to in it."

Forcing internal representations to share activation space with external patterns provides a unique method of *symbol grounding* (see Harnad, 1990, 1993; Gasser, 1993). This occurs in the Analogator network when hidden activations are used in the same bank that retinotopic patterns are also entered (i.e., the context bank). Pollack's RAAM (1988) is the only other architecture that I am aware of that allows internal and external representations to share hardware. However, RAAM is usually used to represent symbolic (e.g., structureless) representations rather than retinotopic patterns, like Analogator does.

The recurrent figure-ground training procedure, like its cousin, the sequential RAAM (Pollack, 1988; Blank *et al.*, 1992), has a more difficult time training than most backprop networks. This is due to the fact that the hidden layer activations that are copied back to the input layer keep changing as training progresses. This means that the network is always associating a "moving target" with the desired outputs (Pollack, 1988). That is, the network attempts to associate a given input with a given output, but, as with all SRN's, the inputs keep changing. In fact, there is no guarantee that such a network will be capable of learning the task, or generalizing to novel inputs. The next chapter explores these issues.

# 5 Experimental Results

*The test of a first-rate intelligence is the ability to hold two opposed ideas in the mind at the same time, and still retain the ability to function.*

-F. Scott Fitzgerald

This chapter will examine the performance of Analogator in the three domains introduced in Chapter 1: the letter-part, geometric-spatial, and family tree domains. All three problem types will use the SRN architecture and the recurrent figure-ground training procedure as described in Chapters 3 and 4, respectively.

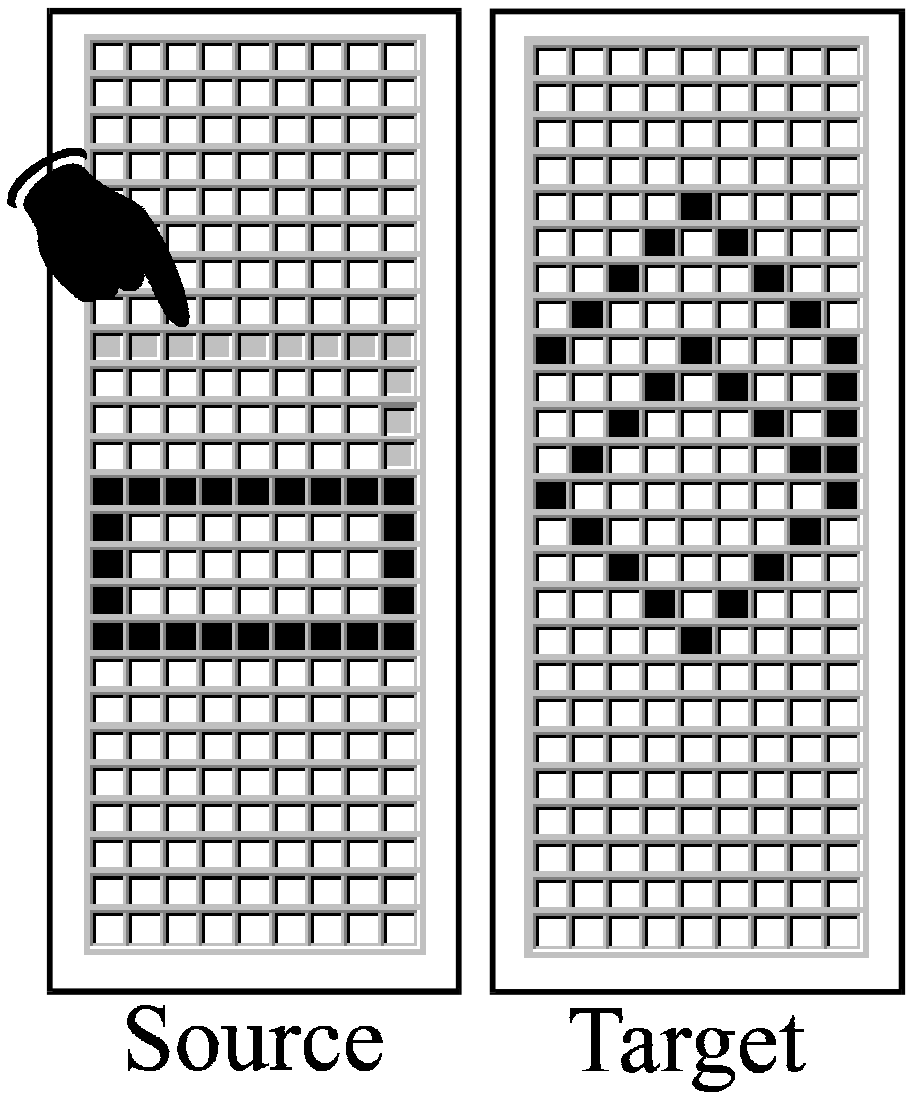

Figure 5-1. A letter-part analogy. What in the target scene is analogous to the shaded portion of the source?

## 5.1 Letter-Part Analogies

Recall that a letter-part analogy requires finding analogous parts between two letters in two different styles of font. For example, Figure 5-1 shows two letter a's. The goal is to find the part in the target letter that plays the same role as the shaded pixels in the source. In this example, most people would probably consider the topmost upside-down V in the target letter to be the analogous part.

The data for all of the letter-part analogies used in these experiments are based on those collected by McGraw (1995) for the Letter Spirit project. The Letter Spirit project is an ambitious attempt at the automatic design of an entire 'gridfont' given a few seed letters in a particular style (Hofstadter *et al.*, 1995; McGraw, 1995). The term 'gridfont' is used as all of the letters of a font must conform to the constraints of a grid of line segments (see Figure 5-2, left side). Although I have adopted some of their data, the Letter Spirit project was not designed to explicitly make the kinds of analogies described in this section. However, in light of their ambitious goals, letter-part analogies would be a trivial problem for a completed Letter Spirit program.[11]

The Letter Spirit grid is composed of 56 line segments with which to construct letters. The segments connect 21 points, stretching horizontally, vertically or diagonally between them. The grid is divided into three regions: the ascender, central, and descender zones (McGraw, 1995). Although it is a restricting grid, there are many different fonts that can be created. In fact, Hofstadter and his research team have created a database of nearly 300 fonts consisting of the lowercase letters ‘a’ through ‘z’. For this set of experiments, I have used only the letter ‘a’ from those fonts. To create the training corpus for Analogator, I selected only 229 of the original 286 fonts as some of the a’s were duplicates, and others were deemed to be too complex (see Appendix B).

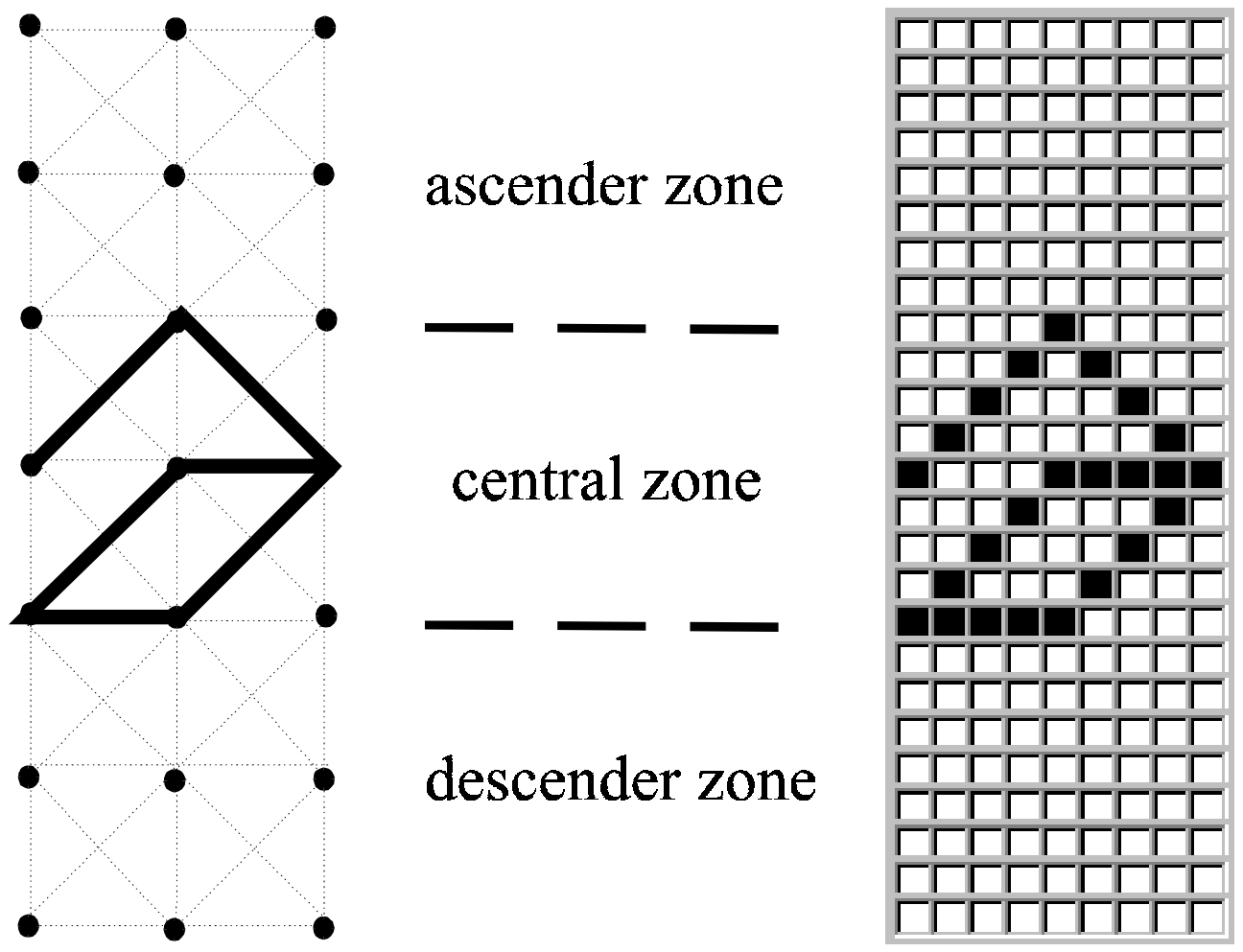

Figure 5-2. The Letter Spirit grid (after McGraw, 1995), and a pixel-based representation.

[11] At present, only the recognizer portion of the Letter Spirit has been implemented (McGraw, 1995). Letter Spirit, like its siblings Copycat and Tabletop, is hand-coded and does not learn the analogy-making mechanism (Hofstadter *et al.*, 1995). Tabletop will be examined in Chapter 6.

A method to convert the line segment representation into a format that could be input into a network was needed. There are many possible methods for translating the segments into a suitable form. The most straightforward method would be to simply let each line segment be represented by an input unit. One could then let the presence of a segment be represented by an activation of 1 on the corresponding unit, and absence be represented by 0. In fact, this localist encoding is exactly the method used by McGraw in his Netrec connectionist letter recognition model (1995) and by Greber *at al.* (1991) in their GridFont network. However, this method conveys very little of the rich perceptual image humans experience when they look at such a letterform. For example, in Netrec each diagonal line was represented by its own unit, sharing nothing with any other line segment. Of course in reality, a line segment on the grid shares much information with other lines, such as its endpoints with other segments. A localist scheme such as this can make generalization very difficult for a network; not all representations are equal.

To represent each letter so that a network could better learn the relationships between line segments, a matrix of 9 x 25 pixels was created. Each line segment was represented by 5 pixels Figure 5-2, right side). Most importantly, each segment shared some of its pixels with other line segments, creating a meaningfully distributed representation.

For the ‘a’ in each font, I identified two roles: the *brim* (as in the brim of a hat) and the *body*. I loosely defined the brim to be every pixel of the ‘overhang’ up to the normally circular area. The body was defined as everything other than the brim (i.e., the normally circular area and anything else that was not the brim). If the brim and the body touched, then that pixel was defined to be a member of both roles. In cases where it was ambiguous where the brim stopped and the body began, I arbitrarily picked a spot. The exact dataset, including annotated brims and bodies, can be found in Appendix B. Because the training for these set of experiments is supervised, each letter must be carefully divided into its parts. Figure 5-3 shows an editing session defining the parts of a letter. The left matrix shows the full letter, the center matrix shows the brim, and the right matrix shows the body.

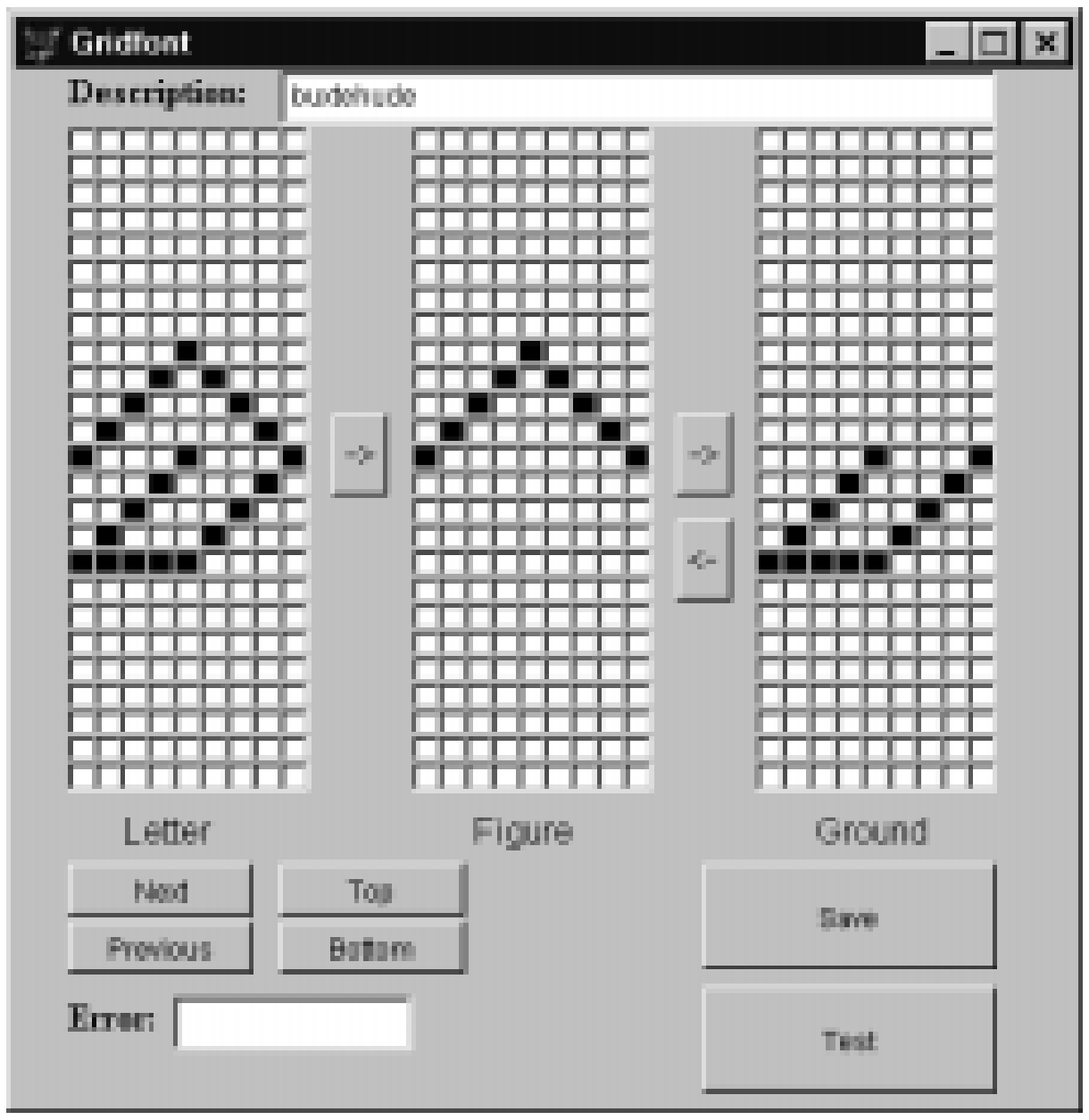

Figure 5-3. The design of a letter ‘a’, and its figure and ground components.

### 5.1.1 Experiment #1a: Letter-Part Identification

For each analogy problem in this experiment, a source letter and a target letter were needed. To use backpropagation as described in Chapter 3, specific desired outputs were also required. Therefore, a single training set consisted of: 1 source letter, 1 source figure, 1 source ground, 1 target letter, 1 target figure, and 1 target ground. The figure-ground training procedure was run as described in Chapter 4. Recall the two-step process:

Step 1. Place a source letter on the input layer (bank S of Figure 5-4, left-hand side). Select a part to be the figure by placing it on the input layer (bank SF on the input layer). Propagate the activation from the input layer, through the hidden layer ($H_t$), and onto the output layer. The desired output is the figure of the source on one bank (SF), and the ground of the source on the other (SG). Note the differences between the actual output and the desired output, and adjust the weights as specified by the backprop algorithm so that next time the network encounters this letter with this figure, the actual output will more closely resemble the desired output.

Step 2. Place a target letter into the same bank that held the source letter from the previous step (now labeled T in Figure 5-4, right-hand side). Copy the hidden layer activations ($H_t$, left-hand side) to the input layer where the figure was held last step ($H_t$, right-hand side). Propagate the activations, and adjust the weights based on the output layer error. Recall that the desired output of this step is the figure of the target on one bank (TF) and the ground of the target (TG) on the other.

I created the corpus for this experiment as follows. For each of the 229 a's, I did the following. First, I selected a letter, and marked the brim as the figure and the body as the ground. Holding this letter constant as the source, 3 other letters from the dataset were picked at random to act as a targets. The target letters (like the source letter) had their brims marked as the figure and their bodies as the ground. For each letter, this created 3 source-target pairs. This entire procedure was performed again, except that this time the *body* was marked as the figure and the *brim* as the ground. This doubled the entire corpus to a total of 1,374 source-target letter pairs.

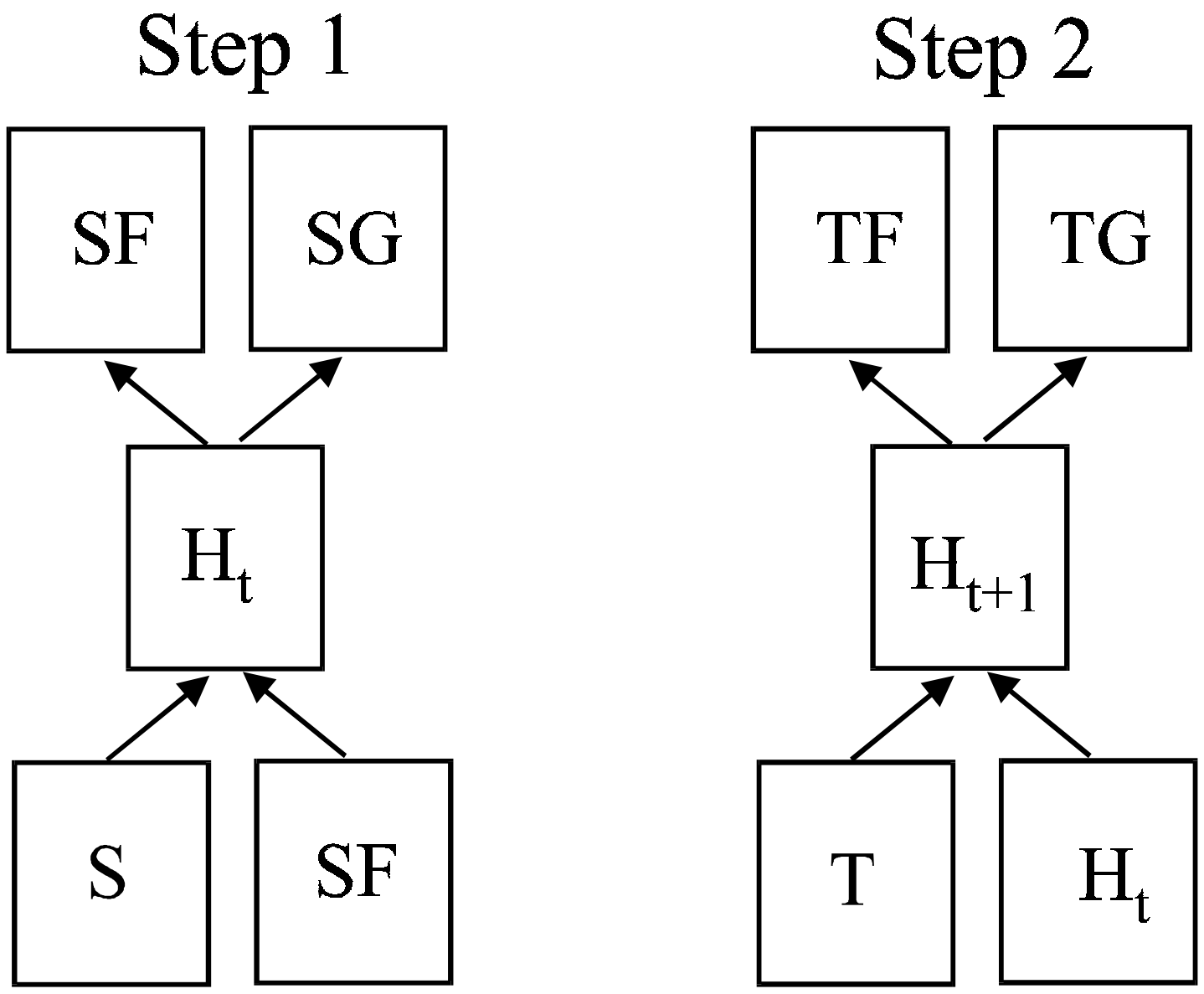

Figure 5-4. The steps of the recurrent figure-ground training procedure, revisited.

One such source-target pair is shown Figure 5-5. In this abbreviated representation of the problem, the source image and figure are shown across the top, and the target image and figure are shown on the bottom. One may think of this arrangement in terms of the standard "proportional" analogy problems often seen on I.Q. tests: "if 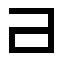 goes to ,

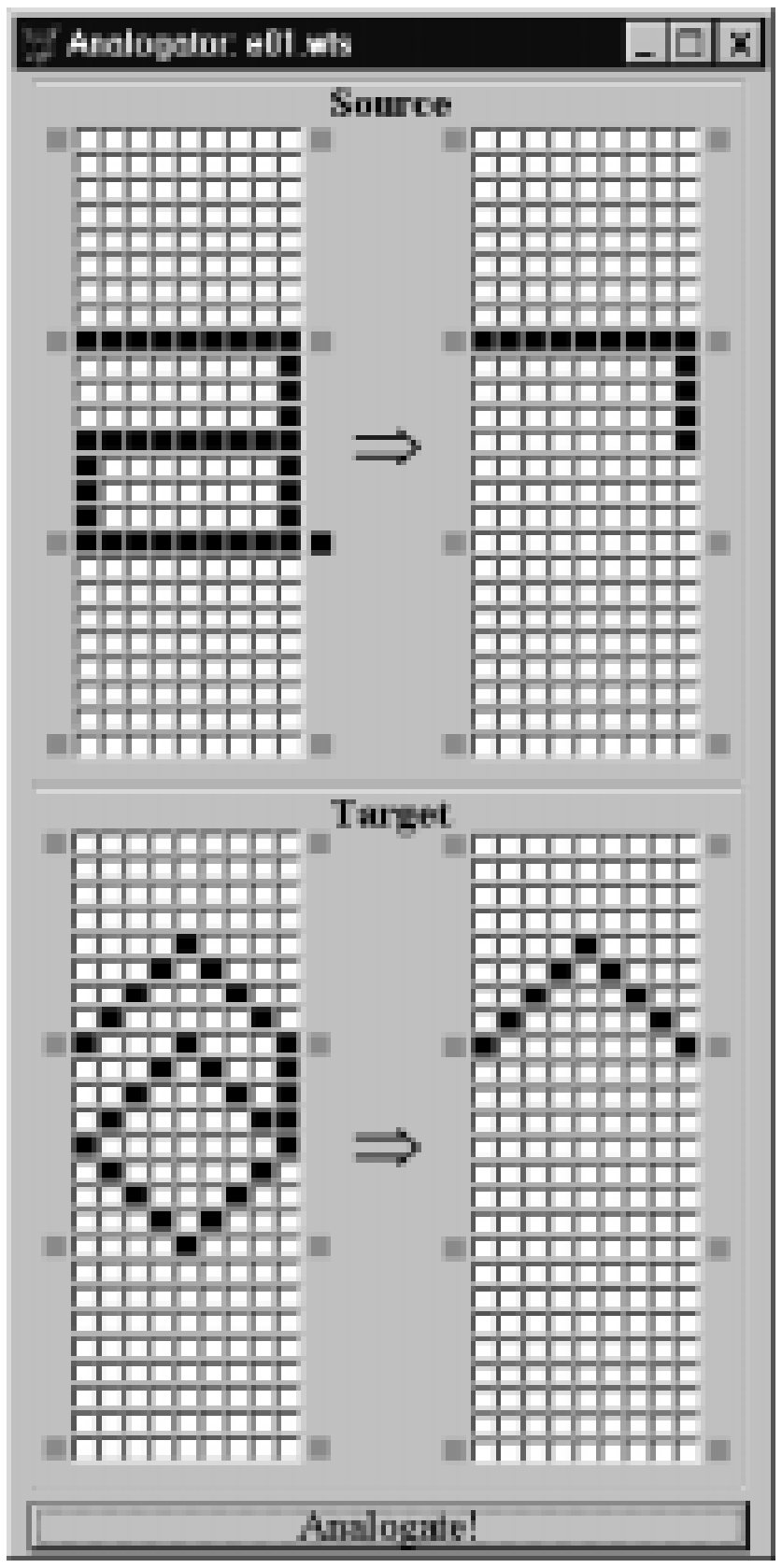

Figure 5-5. A format for examining letter-part analogies. In this version, the analogy can be thought of in this form: "if 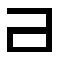 goes to , what does 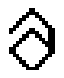 go to?" The network produces the bottom right-hand matrix; the others are inputs to the network.

what does 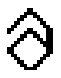 go to?" Notice that the top two matrices, and the bottom left-hand matrix, are all inputs into the network; the goal of the network is to complete the bottom right-hand picture. We will use the form of Figure 5-5 to display the output of the network.

Upon inspection of the 229 letter a's, it was realized that no 'a' used pixels in the top half of the ascender zone, or the bottom half of the descender zone. Leaving those unused pixels out gave 17 rows of 9 pixels for a total of 153 pixels per letter. Therefore, each bank of Figure 5-4 required 153 units. The layers were fully connected, so that 46,818 weights were needed between the input and hidden layer, and another 46,818 weights between the hidden and the output layer. Momentum was set to 0.9, the learning rate (*epsilon*) was set to 0.1, and the bias learning rate was also set to 0.1. The network was then trained as outlined in Chapter 3.

Table 5-1. Training times of Analogator on letter-part analogies. A source-target training pair counts as one trial. A sweep through all source-target pairs is an epoch.

| **Network Architecture** | **Average Number of Epochs to Reach 99% Correct** | **Standard Deviation** |
|---|---|---|
| Analogator | 6.71 (n = 17) | 0.77 |

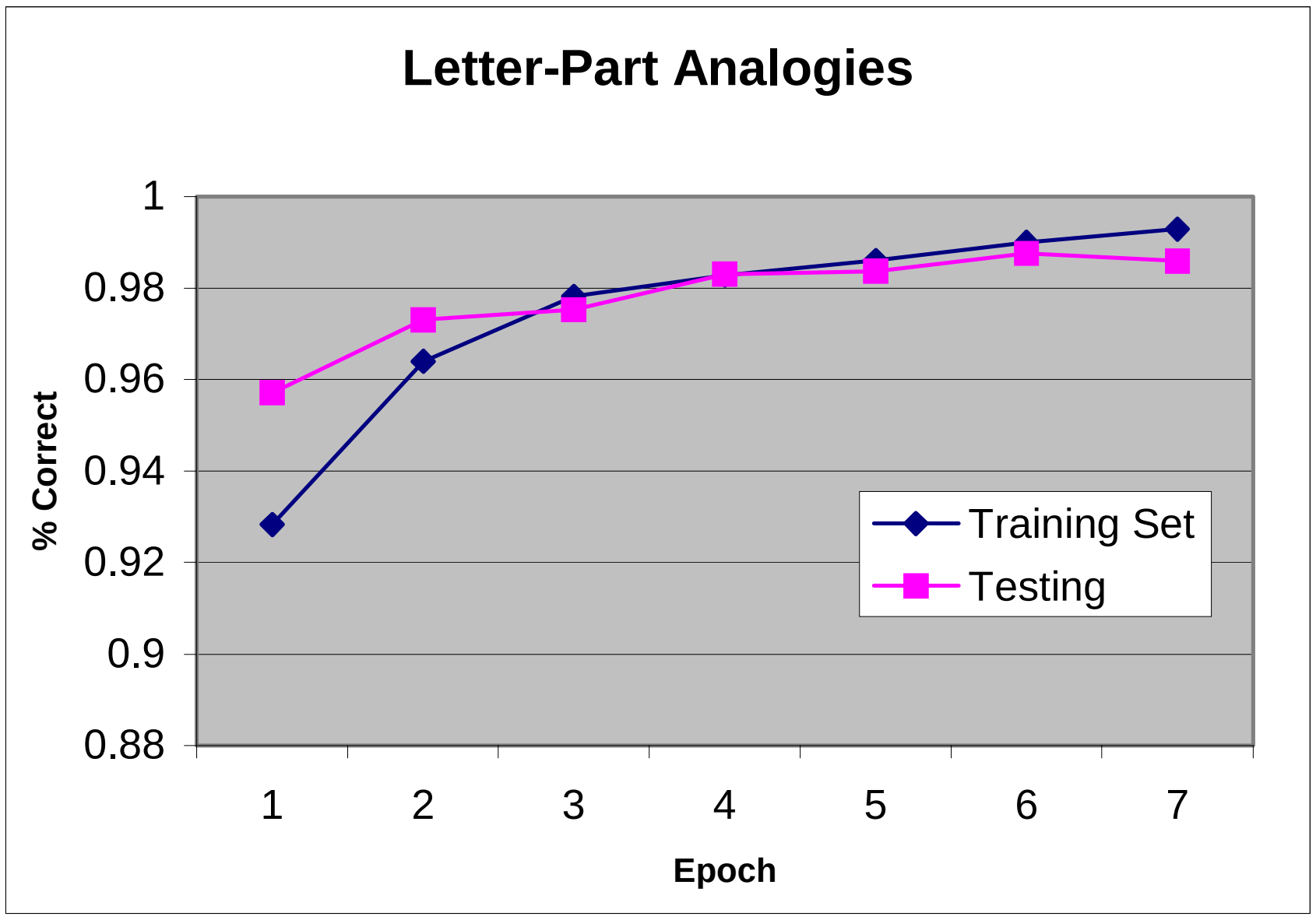

Figure 5-6. The training of an Analogator letter-part network.

The activations on the output layer were rounded to the nearest integer (i.e., 0 or 1) and marked correct or incorrect based on this value. The network was set to automatically stop when network performance reached 99% correct. Percentage correct per epoch was calculated by counting the number of pixels on the output layer that the network got correct, divided by the total number of outputs. There were a total of 612 (153 x 4) possible outputs. In all runs, training went quite quickly, taking usually only 6 or 7 epochs to reach 99% correct. Figure 5-6 shows the course of learning in a typical run. The statistics for 17 runs are shown in Table 5-1.

Figure 5-7 provides a glimpse of the activations in a network from Experiment #1a after training. The top row shows the hidden layer activations, and the output activations for Step #1 (the source letter). Likewise, the bottom shows those from Step #2 (the target letter). Columns from left to right are: 1) hidden layer activations, 2) figure desired outputs, 3) actual figure outputs, 4) ground desired outputs, and 5) actual ground outputs. The darker squares indicate more activation. These snapshots are like a PET scan or fMRI (positron emission tomography, functional magnetic resonance imaging) of

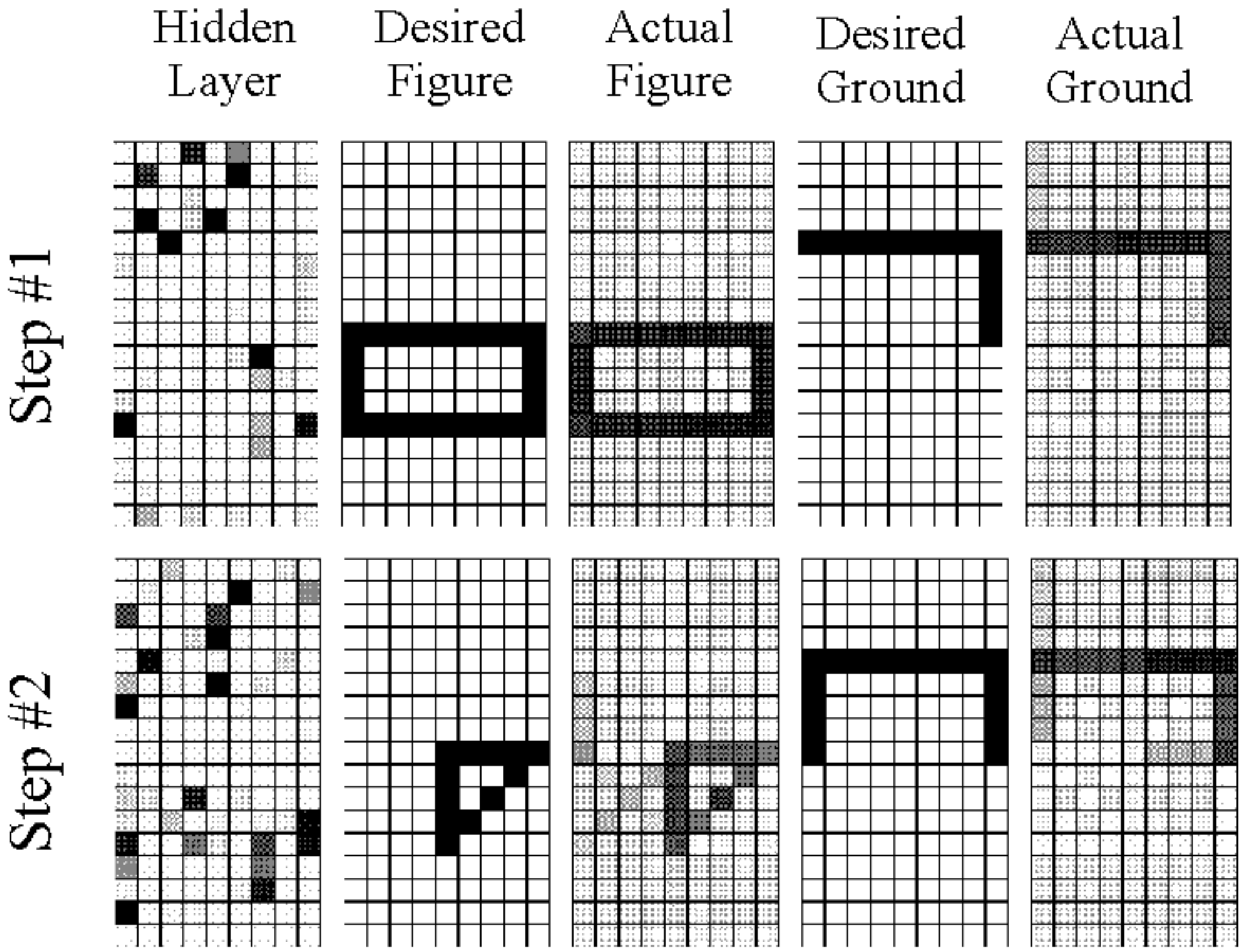

Figure 5-7. A glimpse of some activation values. Top row is Step #1; bottom is Step #2. Columns from left to right are: 1) hidden layer activations, 2) figure desired outputs, 3) actual figure outputs, 4) ground desired outputs, and 5) actual ground outputs. The closer a value is to 1, the darker it is.

Table 5-2. Comparison of training times between Analogator and standard feed-forward networks on the letter-part analogy problem. Analogator's epoch counts have been doubled to make the comparison fair.

| **Network Architecture** | **Average Number of Epochs to Reach 99% Correct** | **Standard Deviation** |
|---|---|---|
| Analogator | 13.41 (n = 17) | 1.54 |
| FF | 96.70 (n = 20) | 105.62 |

Analogator's "brain" – they can show you where the activation is most active, but do not give a clear picture of the details of the representations. Two items worth notice: 1) the hidden layer activations are relatively sparse, and 2) one can see other line segments slightly activated in the actual output activations (columns 3 and 5). Analysis of this experiment continues in the next section.

### 5.1.2 Experiment #1b: Comparison with a Feed-Forward Network

In order to better judge Analogator's performance in the letter-part domain, a standard feed-forward network was created for comparison. Instead of having a two-step recurrent network, the source letter image (S), the figure of the source (SF), and the target letter image (T) were all placed at once on the input layer as shown in Figure 5-8. The network was to produce the target figure (TF) and the target ground (TG) as output. Although the training procedure is simpler than that of Analogator, this network does not share any of the inputs between analogous parts like Analogator does.

The only difference from the Analogator letter-part architecture was that the feed-forward network had an extra bank of 153 units on the input layer, and thus had 23,409 more weights. The training of the feed-forward network used the same corpus as before, and all training parameters were kept the same. Training took much longer with the feed-forward network, with an average number of epochs to reach 99% correct at just under 100. Recall that Analogator actually has two propagate-and-backprop cycles per source-target pair. As this is the case, the number of epochs that Analogator took to reach 99% correct was doubled for comparison and is shown in Table 5-2. Even with this adjustment, analysis of variance showed that it took significantly longer to train the feed-forward network ($p < .005$).

As the feed-forward network had many more weights than Analogator's network, this finding may only be of slight interest. The more important comparison between the two architectures and training methods is the performance on novel letters. To compare the networks, I designed 10 new letter a's. On the ten test letters the brim was marked to be the figure, and on another 10 the body was marked as figure (see Figure 5-9). Therefore, there were 20 test problems. Analogator’s performance on the test set during training is also shown in Figure 5-6. Although each type of network was trained to the same level of performance, Analogator was significantly better at generalization as shown by Table 5-3 ($p < 0.005$).

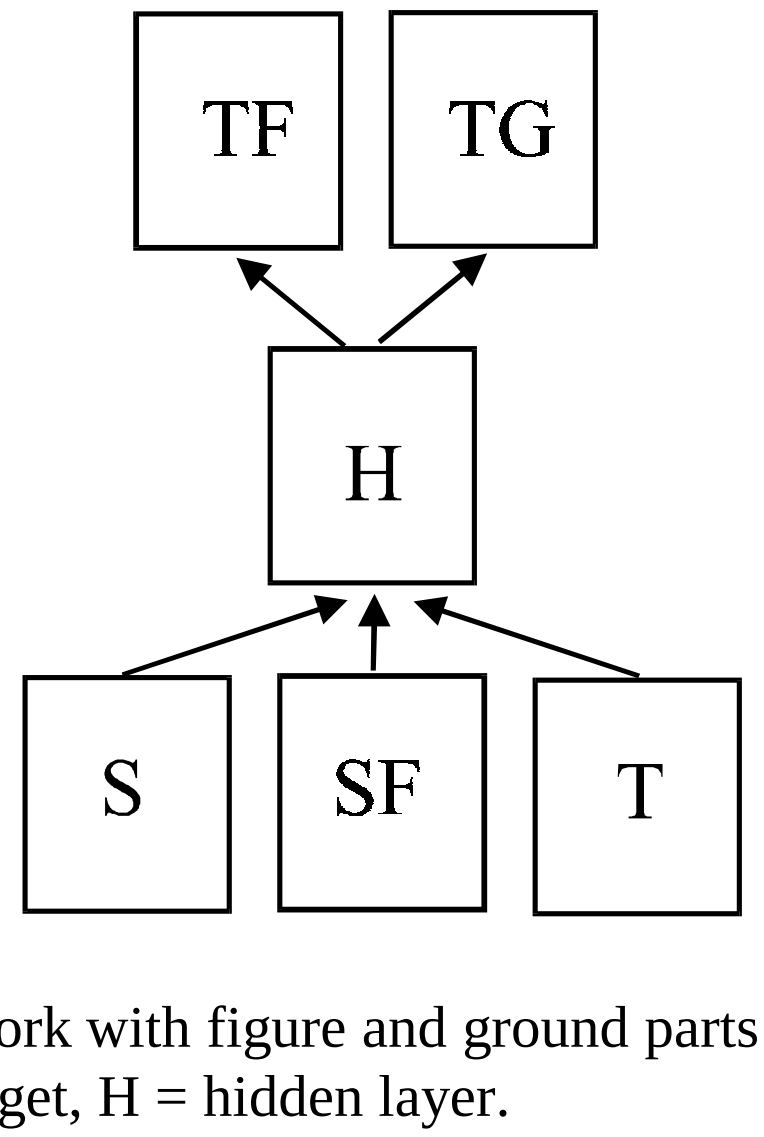

Figure 5-8. Feed-forward network with figure and ground parts as desired outputs. S = source, F = figure, G = ground, T = target, H = hidden layer.

| | | | | |
|---|---|---|---|---|
| .........<br>.........<br>.........<br>.........<br>IIIIIIIII<br>........I<br>........I<br>........I<br>....OOOOB<br>...O....O<br>..O.....O<br>.O......O<br>OOOOOOOOO<br>.........<br>.........<br>.........<br>......... | .........<br>.........<br>.........<br>.........<br>IIIIIIIII<br>I.......I<br>I.......I<br>I.......I<br>I...OOOOB<br>....O..O.<br>....O.O..<br>....OO...<br>....O....<br>.........<br>.........<br>.........<br>......... | .........<br>.........<br>.........<br>.........<br>IIIIIIIII<br>........I<br>........I<br>........I<br>OOOOO...I<br>O....O..I<br>O.....O.I<br>O......OI<br>OOOOOOOOB<br>.........<br>.........<br>.........<br>......... | .........<br>.........<br>.........<br>.........<br>....IIIII<br>...I....I<br>..I.....I<br>.I......I<br>I...O...I<br>...O.O..I<br>..O...O.I<br>.O.....OI<br>OOOOOOOOB<br>.........<br>.........<br>.........<br>......... | ....I....<br>...I.I...<br>..I...I..<br>.I.....I.<br>I...O...I<br>...O.O..I<br>..O...O.I<br>.O.....OI<br>O.......B<br>.O.....O.<br>..O...O..<br>...O.O...<br>....O....<br>.........<br>.........<br>.........<br>.........<br>......... |
| IIIIIIIII<br>........I<br>........I<br>........I<br>OOOOOOOOB<br>O.......O<br>O.......O<br>O.......O<br>O.......O<br>O.......O<br>O.......O<br>O.......O<br>OOOOOOOOO<br>.........<br>.........<br>.........<br>......... | .........<br>.........<br>.........<br>.........<br>IIIIIIIII<br>........I<br>........I<br>........I<br>OOOOOOOOB<br>O.......O<br>O.......O<br>O.......O<br>O.......O<br>O.......O<br>O.......O<br>O.......O<br>OOOOOOOOO | .........<br>.........<br>.........<br>.........<br>IIIIBOOOO<br>....O...O<br>....O...O<br>....O...O<br>....O...O<br>...O....O<br>..O.....O<br>.O......O<br>OOOOOOOOO<br>.........<br>.........<br>.........<br>......... | .........<br>.........<br>.........<br>.........<br>IIIIIIIII<br>........I<br>........I<br>........I<br>....O...I<br>...O.O..I<br>..O...O.I<br>.O.....OI<br>OOOOOOOOB<br>.........<br>.........<br>.........<br>......... | .........<br>.........<br>.........<br>.........<br>IIIIIIIII<br>........I<br>........I<br>........I<br>....O...I<br>...OO...I<br>..O.O...I<br>.O..O...I<br>OOOOBIIII<br>.........<br>.........<br>.........<br>......... |

Figure 5-9. The ten test letters. The letter ‘I’ connotes the brim portion, ‘O’ the body, and ‘B’ both. On ten trials, the brim was defined to be the figure, and the everything else the ground. On another ten trials, the body was defined to be the figure.

Figure 5-10 shows the output from three novel test letters from the Analogator network. In all tests of new letters, a prototypical ‘a’ was used as the source letter, as shown in the top rows of the images in Figure 5-10. The three target letters shown in the bottom, left-hand corner of the images in Figure 5-10 are all actually very similar. In fact,

Table 5-3. Comparison of generalization performance between Analogator and standard feed-forward networks on the letter-part analogies.

| **Network Architecture** | **Average Pixel Errors for 20 Images** | **Standard Deviation** |
|---|---|---|
| Analogator | 133.00 (n = 17) | 24.08 |
| FF | 170.85 (n = 20) | 23.59 |

they only differ by one line segment. However, this causes a large difference in where the brim is perceived to stop. In this case, the network has provided "correct" answers for all of these test problems as indicated by the bottom, right-hand matrixes of the images in Figure 5-10.

It is obvious that Analogator performs much better in terms of training speed and generalization ability. However, Table 5-3 shows that it still makes on average 133 mistakes for every 20 test problems. That is about six pixels Analogator gets incorrect for each analogy problem.[12] Since the average letter in the test set has 34 pixels on and 119 off; getting six pixels wrong is a substantial portion of the problem. Although Analogator did quite well on many of the test problems, such as those in Figure 5-10, on others it showed exactly how little it knew about letter a's and their parts. In an attempt to further reduce the amount of error another experiment was designed.

[12] The average of 6 pixels incorrect was the error from step 1 and step 2. Recall that there is a source and target for each step. Not surprisingly, almost all of the error occurred on step 2 (i.e., the figure and ground of the novel target letter).

### 5.1.3 Experiment #1c: Variable-position letter-part analogies.

Superimposing all of the 229 a's produced the image shown in Figure 5-11. The internal white squares indicate that there were 36 pixels that were never used by any 'a'. Although this does demonstrate the fairly good coverage of the letters over the matrix, it also shows that the letters were rigidly positioned on the grid. A strict placement on the grid helps learning, but hinders the ability to generalize. This is due to the limited ways that pixels are correlated with each other; limited correlations mean less to learn, but also a less rich learning environment. By moving the letters around on the grid, it was hoped that the network would develop a richer internal representation of the letter 'a'.

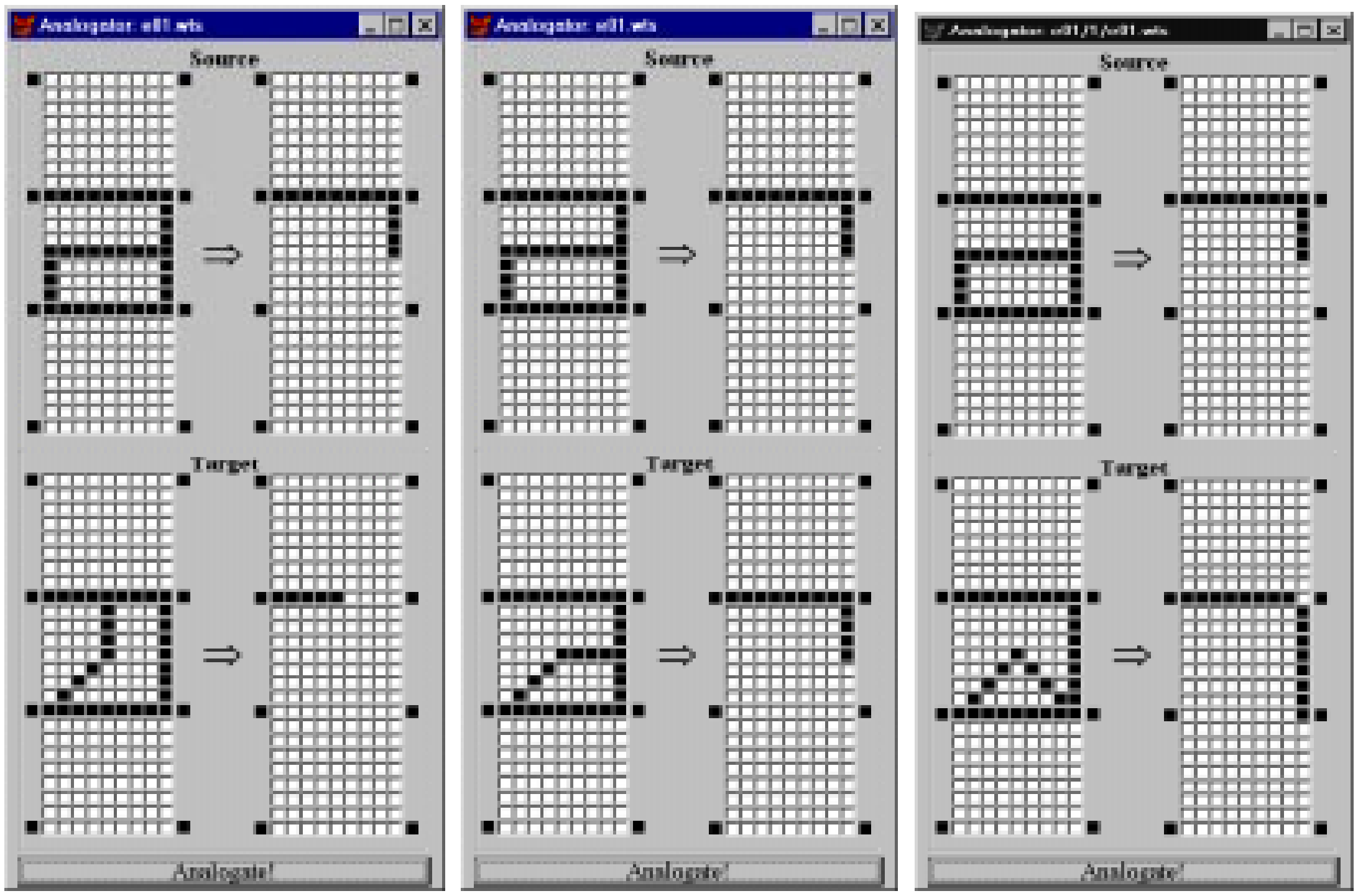

Figure 5-10. Three tests of Analogator's generalization ability from Experiment #1a.

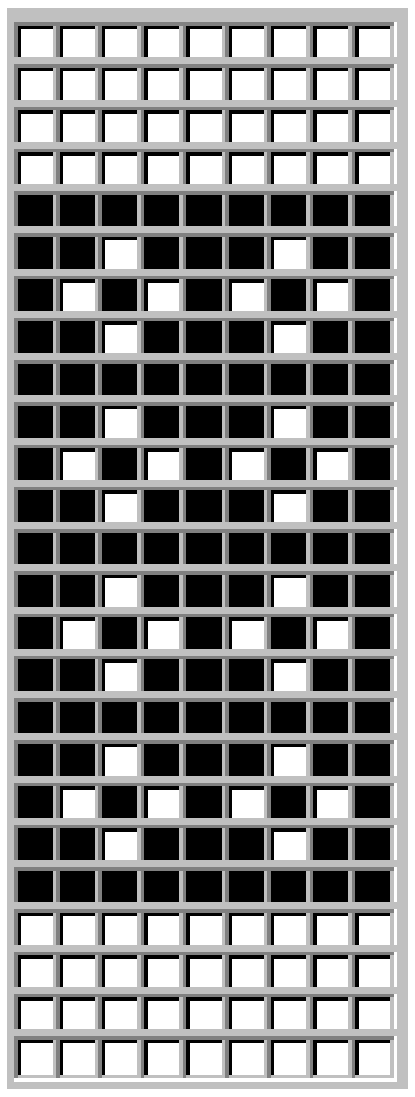

Figure 5-11. Pixels used in the gridfont tests. The top and bottom 4 rows were not used in the experiments.

A corpus similar to the previous was created. However, source a's were confined to be the prototypical 'a' in the standard position as shown at the top of Figure 5-10. Each target letter was put in 9 different positions in the matrix. The variable positions were created by sliding the target letter up on the grid between 0 and 4 rows, or down on the grid between 0 and 4 rows. No side-to-side variability was made. This created a corpus of 4,122 source-target pairs.

After training on this corpus, Analogator networks did not perform significantly better than the previous Analogator version. However, they did make very different *kinds* of errors. The networks trained on the grid-bound letters (i.e., those with no vertical variability) seemed to make errors that were a direct reflection of the frequency of the line segments. That is, pixels that were used frequently in the training letters would also be activated more frequently in the test letters – sometimes appearing out of the blue. For example, Figure 5-12 shows many pixels in the figure and ground of the target activated for no apparent reason (top left). The networks trained on non-grid-bound letters showed much less tendency to make such errors. Instead, the network made more "abstract" errors. Figure 5-13 shows such an error made by a network trained on the richer corpus midway through training. Notice that instead of marking the brim as the analogous part, the network has activated a part of the body that is similar in shape to the brim of this 'a'. This type of confusion would have been nearly impossible with the grid-bound a's as a brim rarely ever dropped that low. Since the variable-position corpus often had brims that low, the network appears to have seen the similarity between parts of bodies and brims, and was confused.

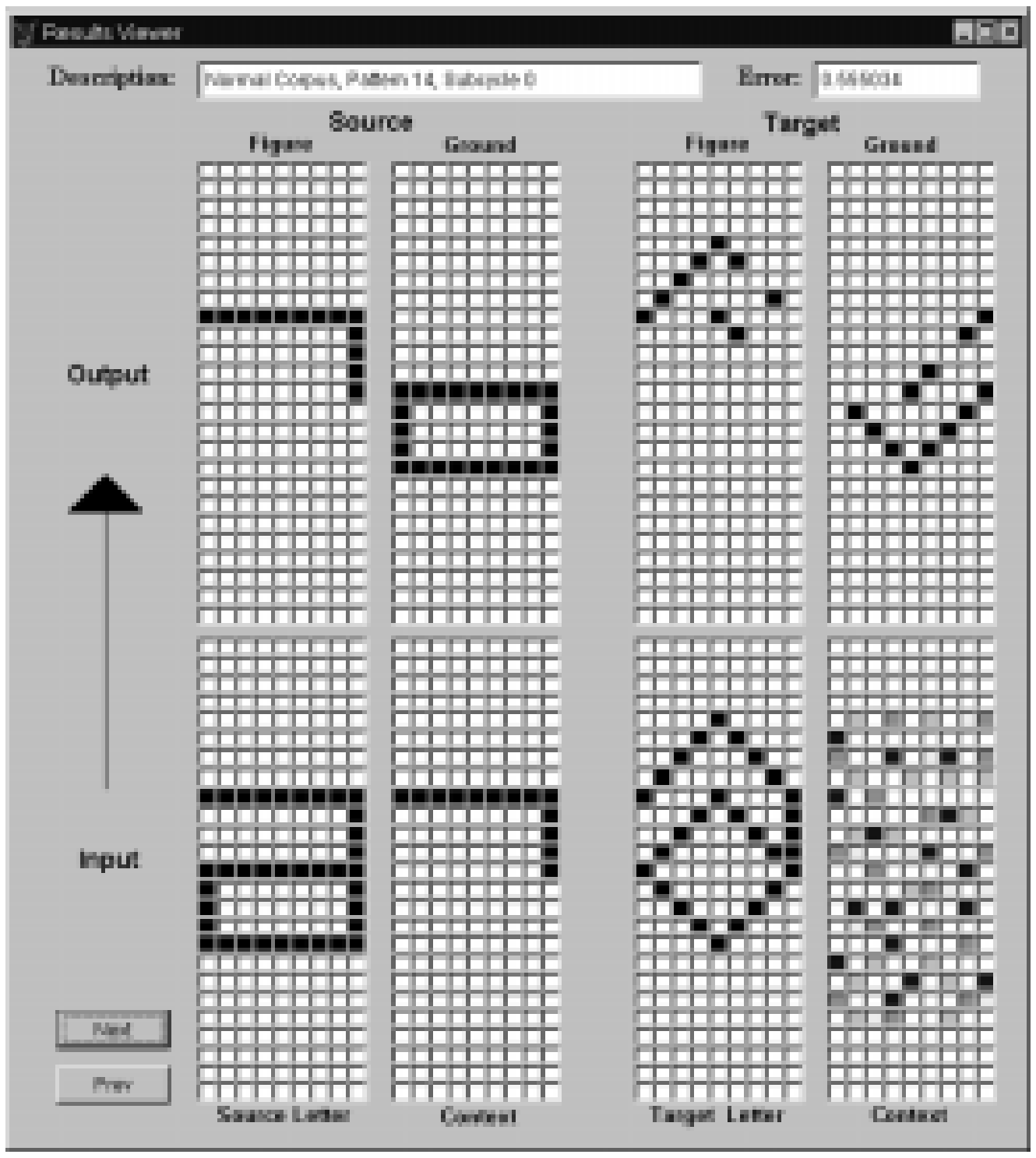

Figure 5-12. A poor performance in generalization ability. Shown here are all of the inputs and output for both steps of the training procedure. The outputs run across the top, while inputs run across the bottom. The left four matrixes represent Step 1, while the right four represent Step 2. The output of the source letter was perfect; however, the output of the target letter was somewhat scrambled.

In general, making these types of abstractions is good. For instance, seeing the similarity between all horizontal line segments can be a beneficial abstraction. Although the networks made more abstractions by seeing the similarity between pieces of the letters, and made less pixel-based statistical errors, it still was unable to harness the abstractions it had learned.

At this point, one might wish to ask, "How does the network actually make analogies? That is, how does it work?" One answer to this question can, of course, be given at the level of the backpropagation of error algorithm with which we trained the network. But this type of explanation does not shed any light on what a network is doing in order to solve any particular letter-part analogy problem. Unfortunately, connectionist networks usually can not provide high-level explanations of how they work, and such a description may not exist for a particular network.[13] Of course, the ability to function without such high-level "rules" is one of the reasons connectionist networks are less brittle than their symbolic counterparts.

---

[13] There are researchers working on the problem of "rule extraction" from trained neural networks (see, for example, McMillian, Mozer, and Smolensky, 1991; Craven and Shavlick, 1994; Andrews, Diederich, and Tickle, 1995).

One method of analysis that can help us to understand the inner workings of a network is the examination of the hidden layer activations. Recall that each input pattern is recoded on the hidden layer on its way to the output layer. The values of activation on the hidden layer are said to define a point in “hidden layer activation space.” A single point does not reveal anything as to how the network operates, but looking at clusters of many points in this space can. Principal Component Analysis (PCA), as described in Chapter 3, provides a method for reducing the vast number of dimensions in such a space down to a manageable few. Experiment #1a had 153 activations on its hidden layer on each step of each analogy problem, and thus 153 dimensions in hidden layer activation space.

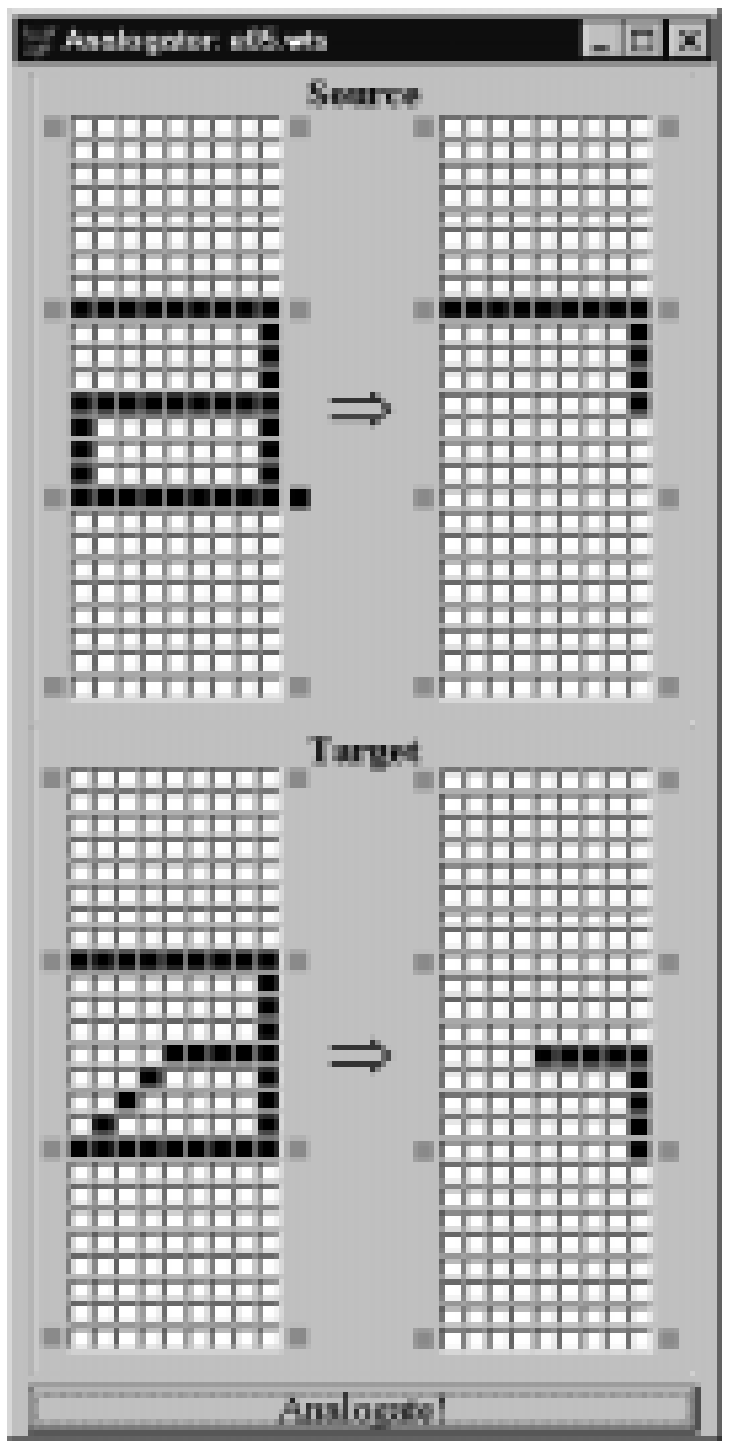

Figure 5-13. A type of error that would have been impossible in earlier experiments. This error was made by a network form Experiment #1c. Notice that there is confusion between the brim and a similar piece of the body. The error does indicate abstractions were made regardless of position on the grid.

Figure 5-14 shows the first two principal components of all of the Step 1 hidden layer activations from Experiment #1a. Recall that the Step 1 hidden layer contains a recoding of the source letter, and the part identified as figure. Although I never explicitly encoded “brim” or “body” in any representation, those are really the only categorizations that the network needed to glean from the source letters in order to complete the analogy in Step 2. Figure 5-14 shows that those source letters that had their bodies marked are clustered together, as are those that had their brims marked. In fact, only the first principal component need be used to cleanly divide the two sets.

That is not to say that the only information contained in the hidden layer activations was a simple categorization of “brim” or “body”. Recall that Step 1 still required the network to produce on the output layer the exact pixels of the figure and ground for a particular letter. The information necessary to do that task must be contained in the other 151 dimensions not shown.

Analysis of the hidden layer activations at Step 2 is not so straightforward. Although there are clusters of activations, they do not reflect easily describable categories. Of course the clusters are based partly on the target letter, and partly on the hidden layer activations from Step 1, however, not much more can be said. This issue will be discussed further in Chapter 7.

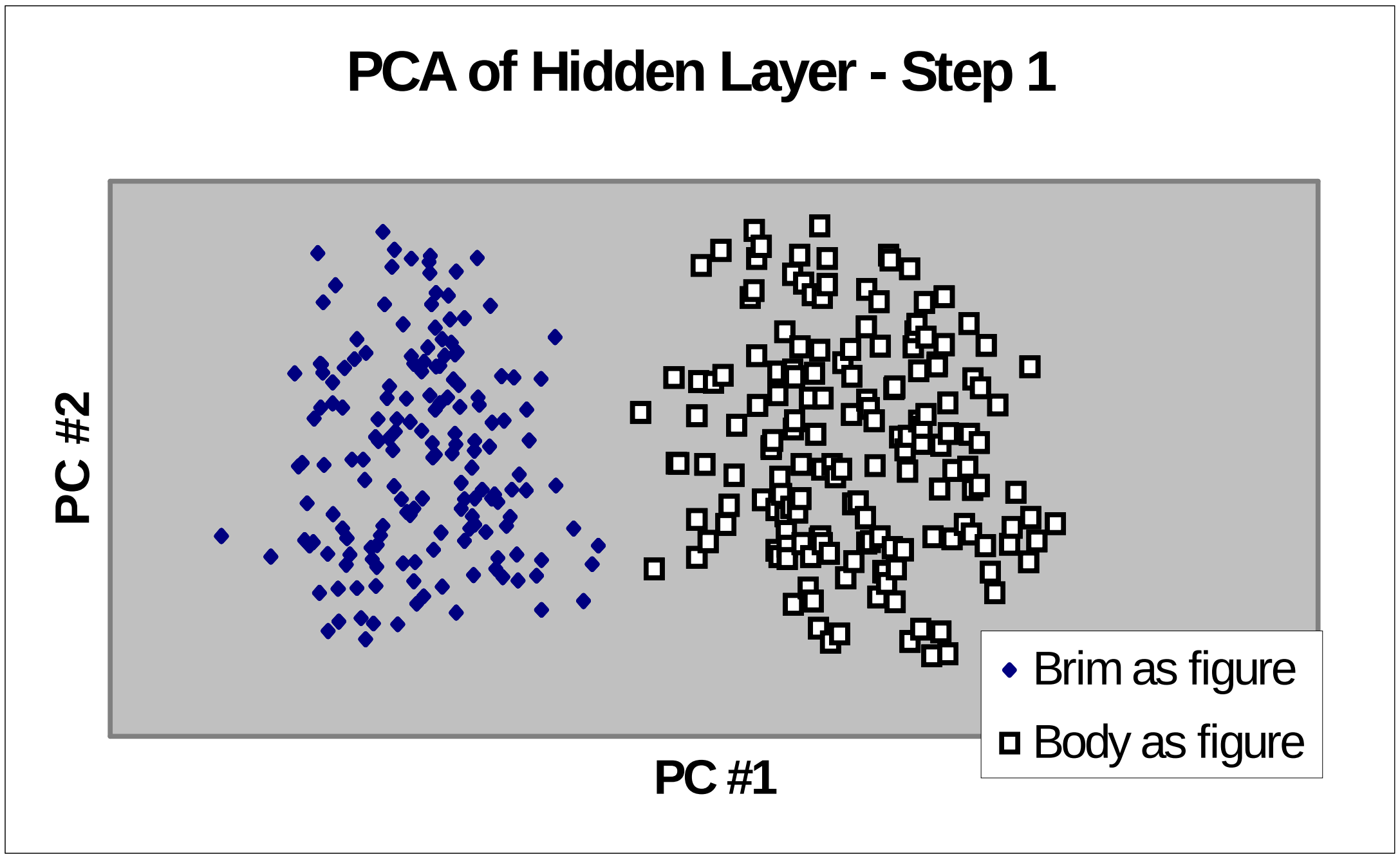

Figure 5-14. Principal Component Analysis of the hidden layer activations of Step 1, Experiment #1a. The activations have been marked according to the type of part selected as figure. Notice that principal component #1 alone allows easy separation of the two types.

In this section we have examined the performance of Analogator on a pair of relations between parts of a single, abstract letterform. Although there are many ways to instantiate the roles of “brim” and “body” of the abstract letter, these were the only components to the problem. We now turn to a domain where relationships are perceived between more than two components, and the relations must be determined on the fly.

## 5.2 Geometric-Spatial Analogies

This section describes analogy-making experiments performed in the geometric-spatial domain. These analogy problems are composed of scenes of objects of differing shape and color. Each object’s shape was defined as either square, circle, or triangle, and each object’s color was defined as either red, green, or blue. All of the following experiments were limited to three objects in each scene, and the objects were constrained to locations near the four corners unless otherwise noted. These limitations were used only to constrain the size of the corpus and the size of the network.

The source and target scenes were created using the iconic representations (IR) described in Chapter 4. Scenes were composed of a 7 x 7 area. Each object occupied a 2 x 2 area in the scene, and was a combination of 2 of the 6 attributes (one of square, circle, or triangle, and one of red, green, or blue). Therefore, a single IR of a scene required 294 (7 x 7 x 6) units. The selected object (i.e., the figure) was marked by activating its location on the 7 x 7 matrix, as with the letter-part analogies.

The network had the same structure as the previous experiments; however, the number of units per bank was altered. The input layer consisted of 343 units (294 + 49); the hidden layer was required to be the same size as the context bank and was therefore 49 units; the output layer consisted of 98 units (49 + 49). As before, standard backpropagation was used to adjust the weights.

Again, the network was trained to make analogies using the recurrent figure-ground associating procedure. First, an IR of a source scene was created and placed on the input layer (see Figure 5-15, step 1a). A representation of the object being pointed to was created by activating nodes (2 x 2) in the context bank (7 x 7) corresponding to the object's location in the scene.[14] The network was then trained to produce the representations of the figure and ground objects of the source scene on the output layer (Figure 5-15, step 1b). Here, the figure of a scene is defined to be the object being pointed to and the ground is everything else in the scene. Next an IR of the target scene was placed on the same units that held the source IR from the previous time step (Figure 5-15, step 2a). Recall that instead of placing activations representing an object being pointed to,

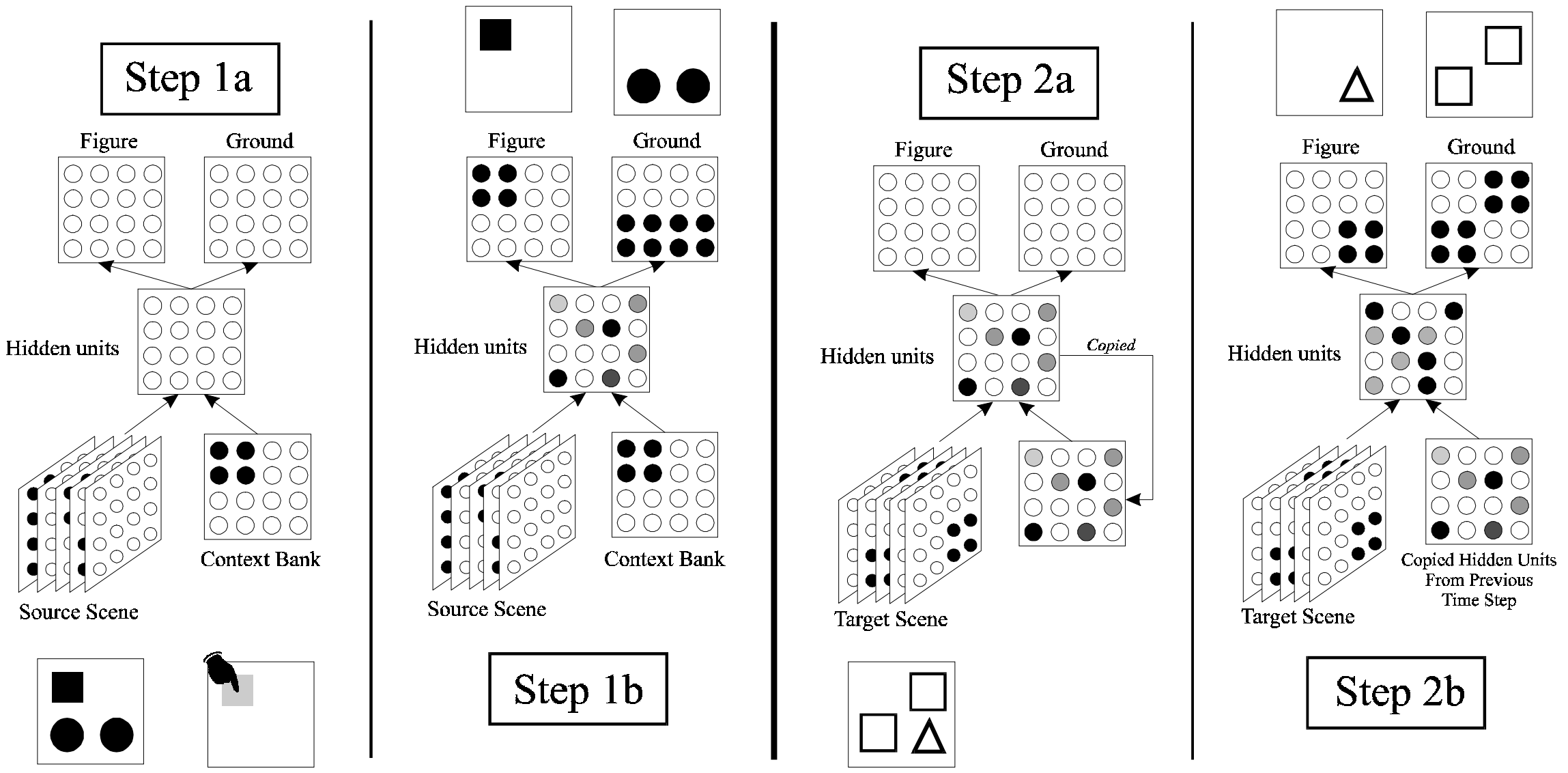

Figure 5-15. The recurrent figure-ground associating training procedure for the geometric-spatial analogy domain. The 7 x 7 grid is shown here as 4 x 4 for clarity.

---

[14] My use of the context bank as a method of pointing to an object is similar to that of Mozer & Sitton's "attentional map" (1997). In fact, Analogator's architecture could be

the hidden unit activations were copied from the previous time step onto the context bank. The network was then trained to complete the analogy, that is, to produce on the output layer the analogous figure and ground (Figure 5-15, step 2b).

We will now examine the kinds of analogies made in the geometric-spatial domain.

### 5.2.1 Experiment #2a: Relative position-based analogies

This experiment was designed to see if a network could learn to make analogies based solely on the relative positions of objects in a scene. This was tested by training a network to select the analogous object in a relative spatial pattern. Each source scene was created by generating three identical objects and placing them in 3 of the 4 positions. The target scene was created by taking the source scene and rotating it 0, 1, 2, or 3 times. A

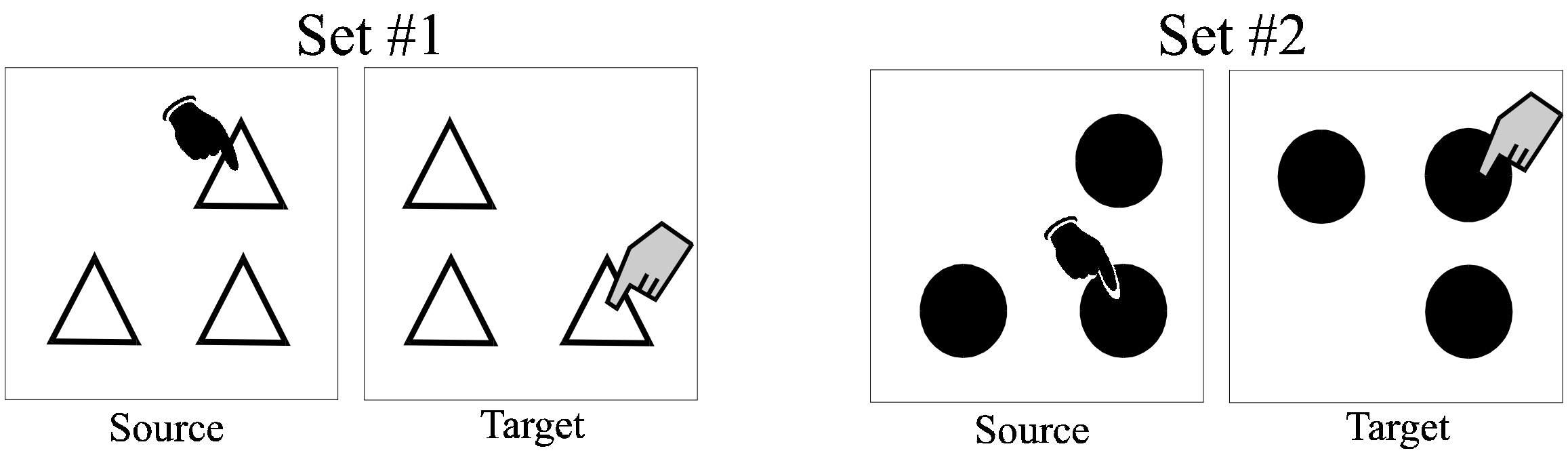

Figure 5-16. Two samples from experiment #2a. Set #1 has been rotated 1 position clockwise. Set #2 has been rotated 3 positions clockwise. The gray hands indicate the correct answer to the analogy problem.

rotation was defined to be a shift to the next position clockwise. For example, the target

---

seen as sitting on top of their model.

scene of Set #1 of Figure 5-16 was generated by rotating the source scene 1 turn; the target scene of Set #2 shows a 3 turn rotation. Notice that in order to solve relative position-based analogies, one must pay attention to the relative position of the objects in the scene.

For each source scene, one object was chosen at random to be the figure. This also dictated which object in the target scene should be regarded as the correct answer (see the gray hands in Figure 5-16). There were 108 (4 x 9 x 3) unique source scenes, and 432 (108 x 4) unique corresponding target scenes per source, for a total of 46,656 unique source-target pairs. Of those, 1,000 (2.1% of total) were saved for testing.

Training proceeded very quickly; the network reached 100% correct in 11,800 trials, requiring less than a complete sweep through the entire dataset. The network was then tested on the remaining 1,000 source-target pairs, and it got each one correct.

This experiment showed that the network could learn to make analogies based on the relative positions of the objects in a scene. The next experiment was designed to see if the architecture and learning procedure could make more abstract analogies.

### 5.2.2 Experiment #2b: Abstracted Category analogies

The next experiment was designed to see if the network could learn to identify objects based on abstracted categories such as “the object that is different from the other two on the dimension of color.” This was tested by creating a source scene that contained three identical objects. One object was chosen at random and made to differ from the other two by randomly changing either its shape or its color. The target scene was created by also creating three identical objects (not necessarily identical to those in the source). One object from the target scene was selected at random and its shape was changed. A second object was then chosen in the target scene and made to differ by color (see Figure 5-17.) The object that was different in the source (by either shape or color) was "pointed to" using activation on the context bank. The network's goal was to select the object in the target scene that differed "in the same way." For instance, a **green circle** differed from two **blue circles** in the same way that a **red triangle** differed from a **green triangle** and a **green square**. That is, they both differ on the dimension of color. Note that "same" and "different" were not explicitly represented but had to be learned by the network. This task required that the network pay attention to the attributes; position played no role in determining an answer. However, recall that location is the "binding" dimension (see Chapter 4).

Given these constraints there were a possibility of 216 variations of the source scene for each of the two types (e.g., "differ by shape" and "differ by color"). There were a total of 1,080 possible target scenes and, therefore, a total of 466,560 unique combinations of pairs of source and target scenes (2 x 216 x 1,080). Notice that this combination of possible variations is exactly an order of magnitude larger than that of the previous experiment.

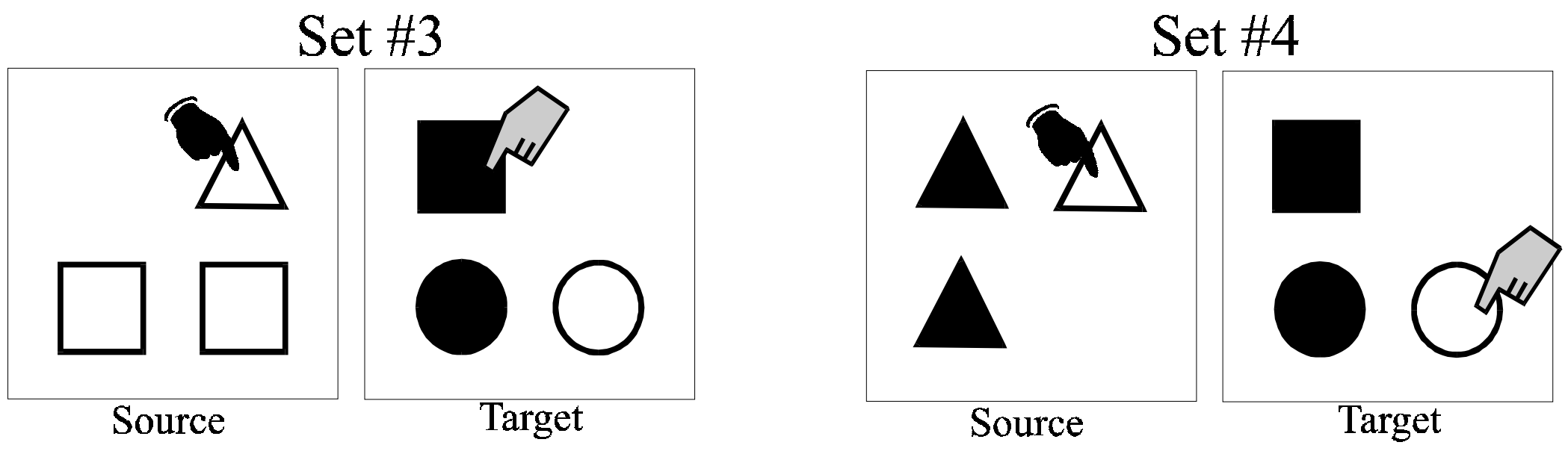

Figure 5-17. Two sample problems from Experiment #2b. The source scene from Set #3 shows an object selected which differs from the other two source objects on the dimension of shape. Therefore, the correct answer for Set #3 is the black square of the target (pointed to by the gray hand), as it also differs on the dimension of shape from the other two objects. Set #4 shows an analogy based on a difference on the dimension of color.

A database of 10,000 unique source-target pairs was generated randomly. The network was trained on 9,850 of these, leaving 150 (1.5% of total) to test for generalization after training. After being trained on 100,000 trials on instances of the 9,850 source-target training pairs, the network reached 100% performance. That is, the network had learned that if the figure in a source scene differed from the ground (in the source scene) by shape then the object in the target that also differed by shape was the figure of the target (as with Set #3 of Figure 5-17.) Likewise, the network had learned that if the figure in a source scene differed from the ground by color then the object in the target that also differed by color was the figure of that target.

To test the generalization ability, the 150 pairs of source-target scenes that were not used in training were tested and the network got each problem correct. Therefore, the network had learned to make analogies based on abstracted categories which were themselves based on the attributes of the objects.

### 5.2.3 Experiment #2c: Abstract category + Position-based analogies

The next experiment was designed to test analogies that required information from both relational positions and attributes. Target and source scenes were created as before with the following changes. Target scenes were versions of the source scene with a literal twist: they were either flipped (horizontally or vertically), rotated (0, 1, 2, or 3 times), or flipped and rotated. Set #5 of Figure 5-18 shows a sample of a target scene that was flipped about a vertical axis. Set #6 shows a sample flipped about a horizontal axis.

This task was more difficult than the previous two experiments. In an attempt to get the networks to learn to perform the task, the training parameters were adjusted many ways, and the corpus was altered in the midst of training. However, the network was unable to learn to do all of the types of twists and flips no matter how the network was trained.

To test to see if this was just too hard a task for the network to learn, a series of “concept units” were added to the input. These units provided “hints” as to the type of twisting for each analogy (Abu-Mostafa, 1990). Four units indicated the amount of rotation (0, 1, 2, or 3 turns) and 4 units indicated the flipping (horizontal, vertical, both, or none.) Using these hints, 20,000 pairs of source-target sets were generated. The network was trained on 19,500, leaving 500 for testing. With the hints, the network was able to learn all of the tasks getting 100% of the training corpus correct after only 66,000 trails. Of the remaining 500 testing pairs left, it got 484 correct (96.8%).

The hints provided help in identifying the “type” of a particular analogy problem (i.e., rotated, mirrored, etc.). These hints could have been the output of another classification network, as the type would have been straightforward to learn. Yet, without those hints on the input layer, the network was unable to learn to make the analogies.

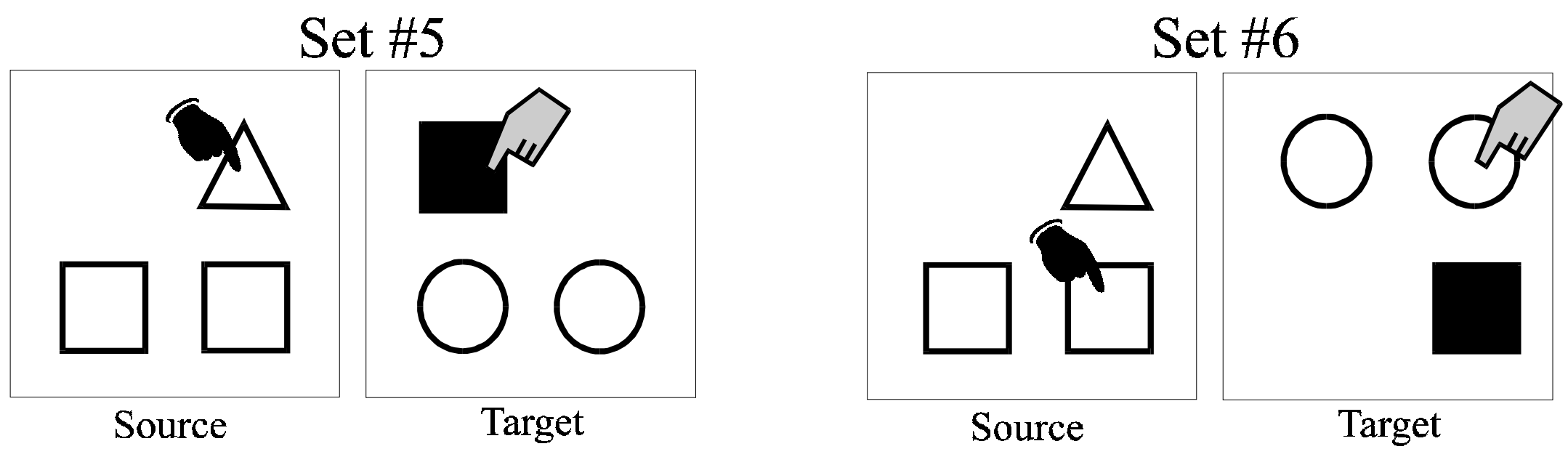

Figure 5-18. Sample from Experiment #2c. Set #5 shows a target that is a vertical-mirror image of the source. However, this fact is not needed as it is easy to see the analogous shape is the one that differs from the other two. Set #6 is not so straightforward. The selected shape is one of two squares. To pick the analogous circle, one must realize that this problem involves a horizontal flip.

### 5.2.4 Experiment #2d: Random-position, Abstract Category Analogies

Although the design of the iconic representation and network architecture allows object placement anywhere in the scene, so far that placement has been limited to 4 locations. This experiment was designed to test the generality of the placement of objects.

As before, this experiment was a test to identify similar objects based on an abstract category. A corpus was created like that from Experiment #2b above. However, this time the objects were placed in random positions covering the full area of the 7 x 7 scene. Again, they were not allowed to overlap.

The network did not learn this problem very easily, or very well. After hundreds of epochs, it had finally learned the training corpus, but never got above 95% on tests of generalization.

The tests of Experiment #1 and #2 showed that an analogical mechanism could be learned by being trained on example analogies in spatial domains. Could the figure-ground associating method work on non-spatial analogies? The next set of experiments was designed to explore this question.

## 5.3 Syntactic Analogies

All of the following experiments are based on a domain created by Hinton (1988). Hinton's family tree model is a connectionist network that learns about relations between people in two isomorphic families, one Italian and the other English. Hinton's network was trained to answer questions, such as "*Who is Victoria's mother?*", and "*Who is Alphonso's grandfather?*" After being trained on most of the relations from the two families, the question was: what will the network do with relationships on which it has not been trained? Hinton's model did learn, arguably by analogy, about the missing relations. This section describes a set experiments done in comparison to Hinton's model. Before exploring the Analogator version, let us examine Hinton's method and model in detail.

### 5.3.1 Hinton's Feed-forward Model

Hinton constructed two isomorphic family trees as shown in Figure 5-19. Next, Hinton created representations for the relationships and the people to be used as input into a connectionist network. Each of the 24 people from the two families was assigned a representation in an orthogonal encoding scheme. As this was a localist representation, there were no features representing age, gender, nationality, etc. In addition, there were 12 relationships defined between the people: *father*, *mother*, *husband, wife, son, daughter, uncle, aunt, brother, sister, nephew*, and *niece*. Each of those relations was also assigned a representation in a localist encoding scheme. Among the two families, there are 104 unique relationship "facts", such as *Sophia's father is Marco*.

A feed-forward network was created as shown in Figure 5-20. The network was trained via back-propagation to answer questions regarding relationships in the two family trees as follows. First, a fact was chosen, such as *The mother of Sophia is Lucia*. The representations of *Sophia* and *mother* were placed on the input layer, and the network was trained to produce the representation of *Lucia* on the output layer as shown Figure 5-20. The network was trained on 100 of these facts. The 4 that remained were set aside for testing of generalization after training.

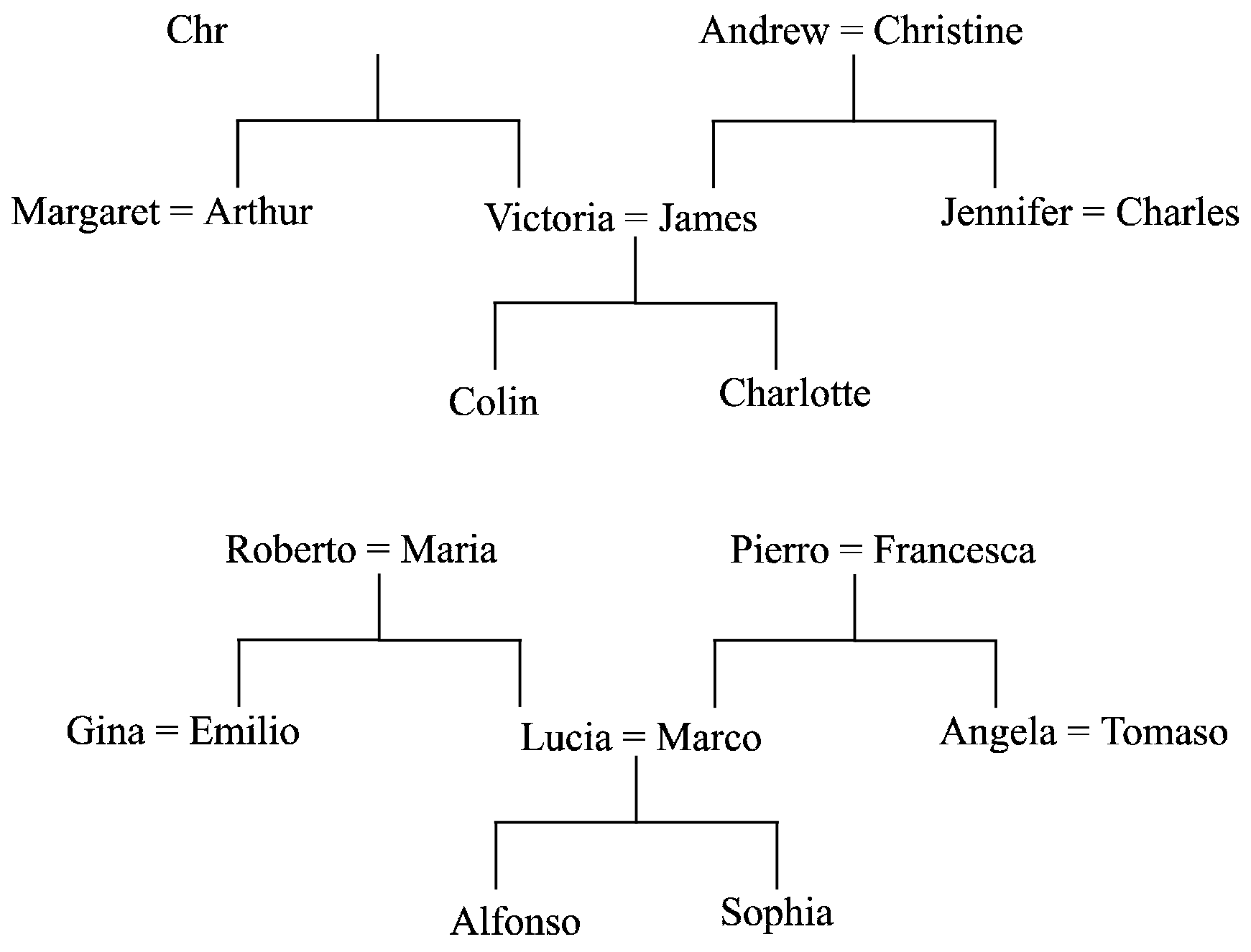

Figure 5-19. Two isomorphic lineages used in Hinton's family tree task (1990).

After training, the weights of the network were examined, and the network was found to have "discovered" many features that were not explicit in the representations of the facts. For instance, the network had discovered the distinction between the two family trees (English vs. Italian), generation (first, second, or third), and to what branch of the family each person belonged.

The four untrained relations were then used to test for generalization. For instance, assume that one of the untrained relations was the fact that *Sophia's father is Marco*. In some sense, the network knew nothing about *Sophia*'s relationship to *Marco*, as that fact was never given to the network. However, when tested on the unknown relations after training, Hinton's network did give the correct answers (3 out of 4 on one trial, and 4 out of 4 on another) (Hinton, 1990).

Although the family tree model took around 15,000 epochs to learn the task, once it did, it generalized over the test patterns.[15] To see if the Analogator model could perform this type of non-perception based task, I altered the task slightly and performed the following experiments.

### 5.3.2 Experiment #3a: Analogator's Family Tree Analogies

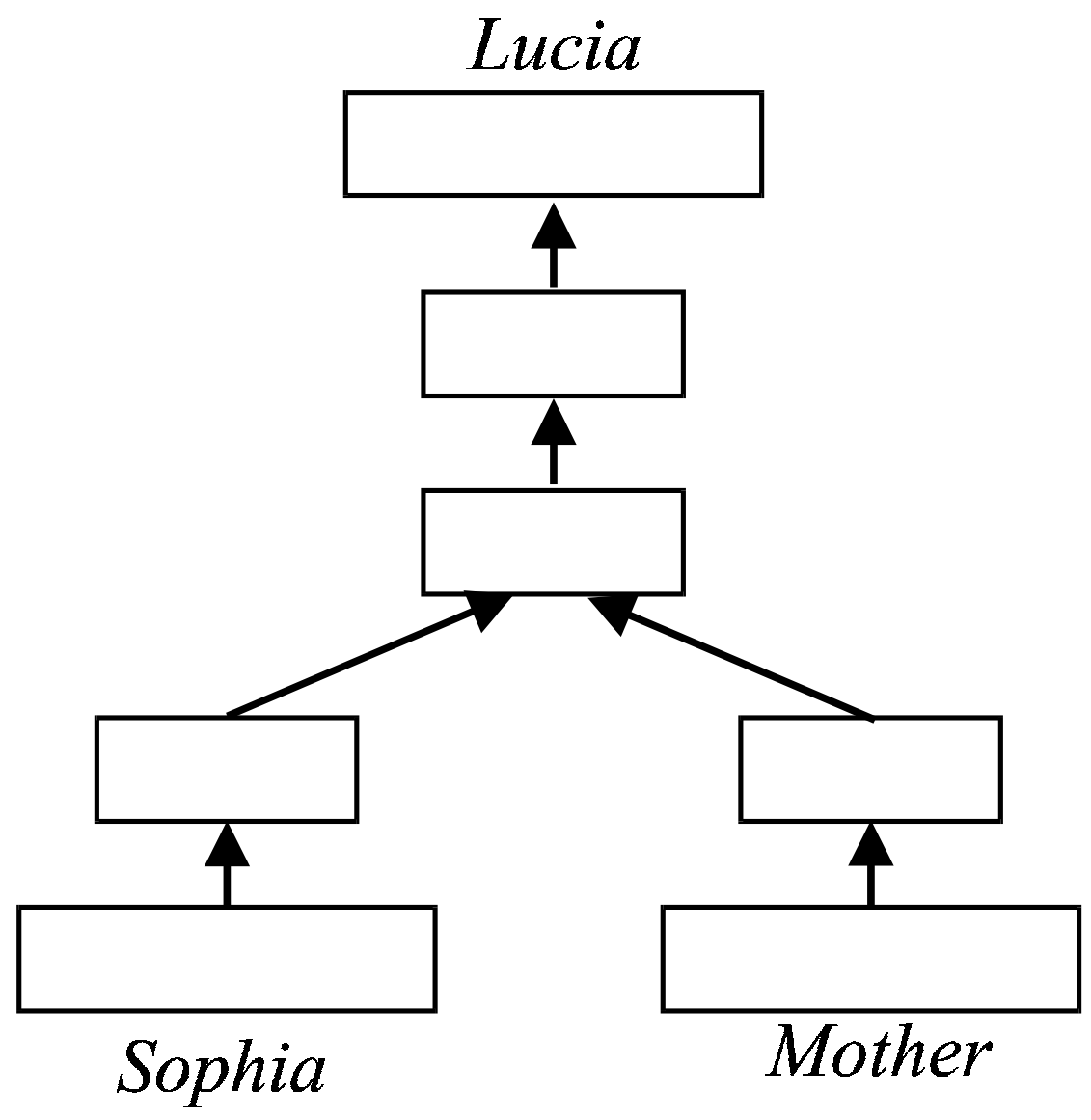

Figure 5-20. Hinton's simple network for learning family relationships. Shown here is the fact that *Sophia's mother is Lucia*.

[15] Melz (1992) states that Hinton's (1990) original reported training times were probably in error. Originally, Hinton had reported training times around 1500 epochs. Melz confirmed that this was probably off by a factor of ten through the replication of Hinton's network. I also replicated Hinton's network and found training times to be closer to 15,000 than Hinton's original report.

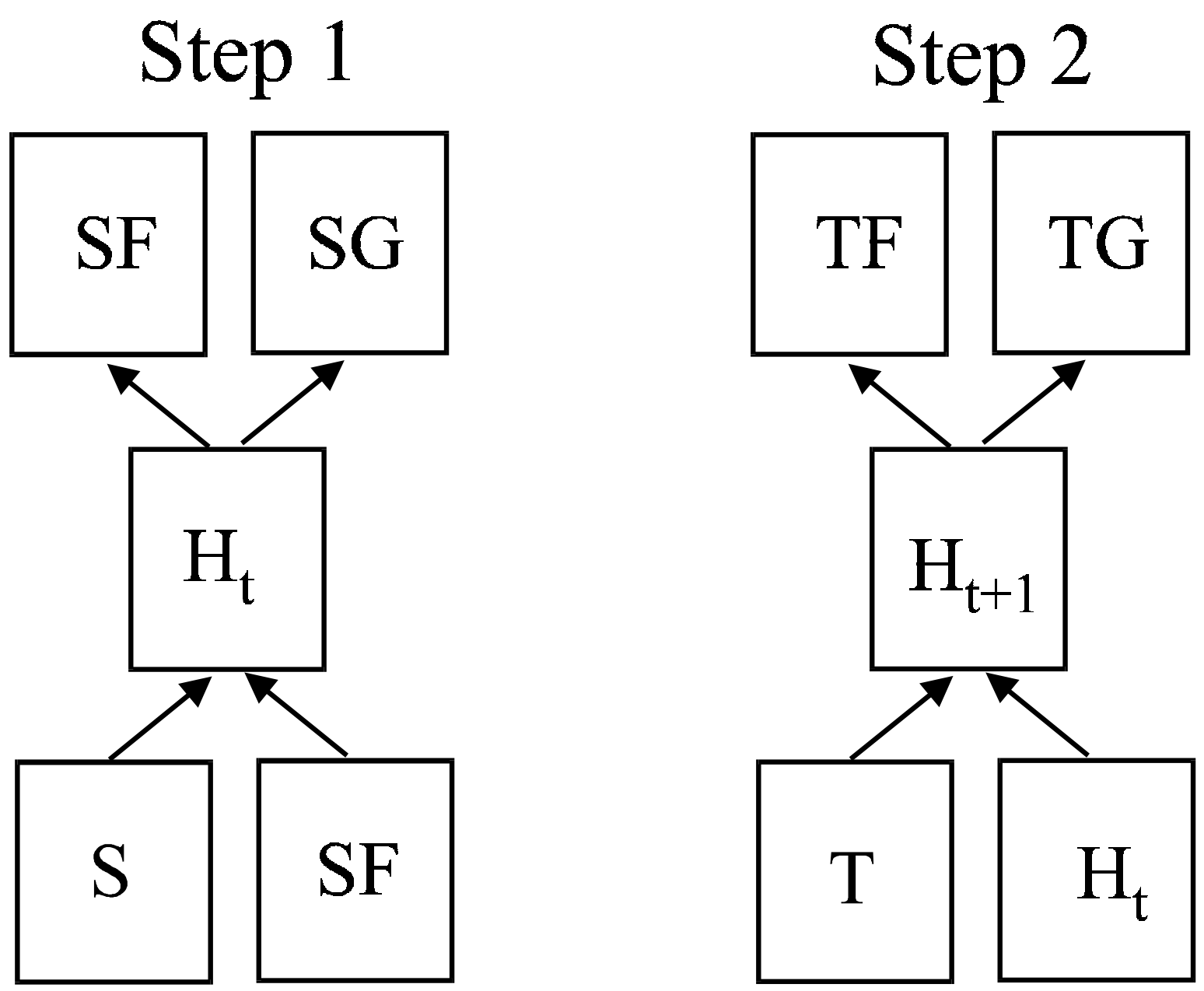

Figure 5-21. The steps of the recurrent figure-ground training procedure, revisited.

In the next set of experiments, the same relationships were used as described above (i.e., the 104 relationships between the two families). I also used a localist encoding scheme to represent the people. In this version, I used the recurrent figure-ground associating training procedure. In order to do so, it was necessary to change the task slightly.

As an example of the altered task, consider the relationship facts *Victoria is the daughter of Penelope* and *Angela is the daughter of Pierro*. In order to get these facts into a figure-ground form, I combined and transformed them into the form *Victoria is to (Victoria and Penelope), as Angela is to (Angela and Pierro)*. Notice that *(Victoria and Penelope)* is the source scene, and *Victoria* is the figure. Similarly, *(Angela and Pierro)* is the target scene, with *Angela* as the figure. Also notice that in the new form, the name of the relationship, (i.e., *daughter*), was never mentioned. This keeps the relationship implicitly represented.

The same network architecture was used as with the previous Analogator domains (Figure 5-21). Again, the only change was the size of the network banks. As there were 24 people to represent, there were 24 units per bank. When training the network, representations of both people from a “scene” were activated on the source (S) and target (T) banks. As these are localist representations, this meant that two of the 24 units were activated.

In our example above, Step 1 would have the representation of *(Victoria and Penelope)* activated on bank S along with the representation of *Victoria* on bank SF. The network would be trained to produce *Victoria* on bank SF and *Penelope* on bank SG of the output layer. Step 2 would have the representations of *(Angela and Pierro)* placed on the T bank. The network would then be trained to have the representations of *Angela*, and *Pierro* appear on the output banks TF and TG, respectively.

Pairing the original facts in this manner creates 1,008 unique pairs of relations between the two families. For example, all of the “mother of” relations from one family are paired with all of the other “mother of” relations from another. The network was trained on all but 8 (0.8% of total) of the pairings. Training went fairly quickly, reaching 99% correct in less than 400 epochs. When tested on the 8 relation pairs saved to test for generalization, Analogator got each one correct.

In a manner similar to Hinton’s network, Analogator had learned internal representations that reflected meaningful concepts. Figure 5-22 shows principal component #1 and #2 of the hidden layer activations from Step 1. The representations of the people from the English family tree have clustered together, as have those from the Italian family.

Although the relations were not explicitly represented as with Hinton's network, the network was still able to learn to make the analogies. In addition, Analogator's hidden representations reflect useful concepts, as did Hinton's.

In one way, this experiment differs significantly from analogies made by humans. Recall that I trained the network on the majority of the corresponding pairs of people in the two families. The network needed only fill-in a few missing correspondences. It was significant that the network was able to generalize and complete the missing facts; however, this task is very different from the analogies humans spontaneously make in the course of everyday life. This point raises a number of questions: Could a network learn to make correspondences between two separate domains without being trained on a majority the mappings? Or better yet, could a network learn to make correspondences between two separate domains without being trained on *any* of the mappings? Could a network learn to match similar entities based on the structure of the relationships alone? The following experiment was designed to explore these questions.

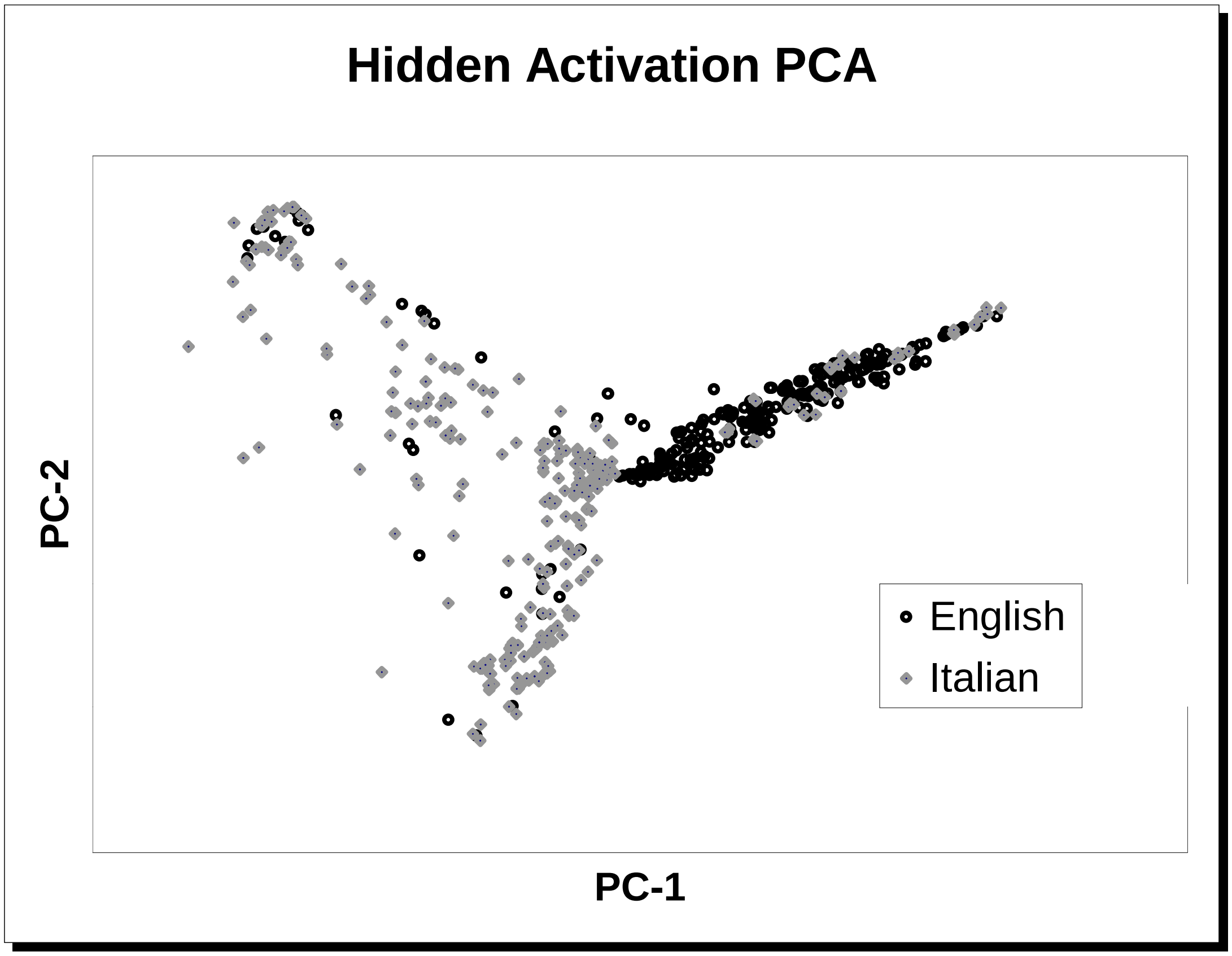

Figure 5-22. Hidden layer activation PCA for Analogator's family tree problem. Shown here are the hidden layer activations for all relations in Step 1 of Experiment #3a.

### 5.3.3 Experiment #3b: Cross-domain Family Tree Analogies

The initial experiment that I attempted in order to address these questions was to simply eliminate all cross-domain pairings from the previous corpus. That is, I simply did not train any English sources to be paired with Italian targets, and vice versa. In one way, this is more similar to Hinton's experiment. However, recall that Hinton explicitly used relation names, such as *mother of*, whereas I made no mention of relation names. My version was a true cross-domain analogy – the relations could really have been completely different sets of relations. Because he used the same relation names for both families, Hinton was learning to make analogies in a single domain (intra-domain analogies). However, when trained without the cross-domain pairings, my version was unable to learn to perform the task.

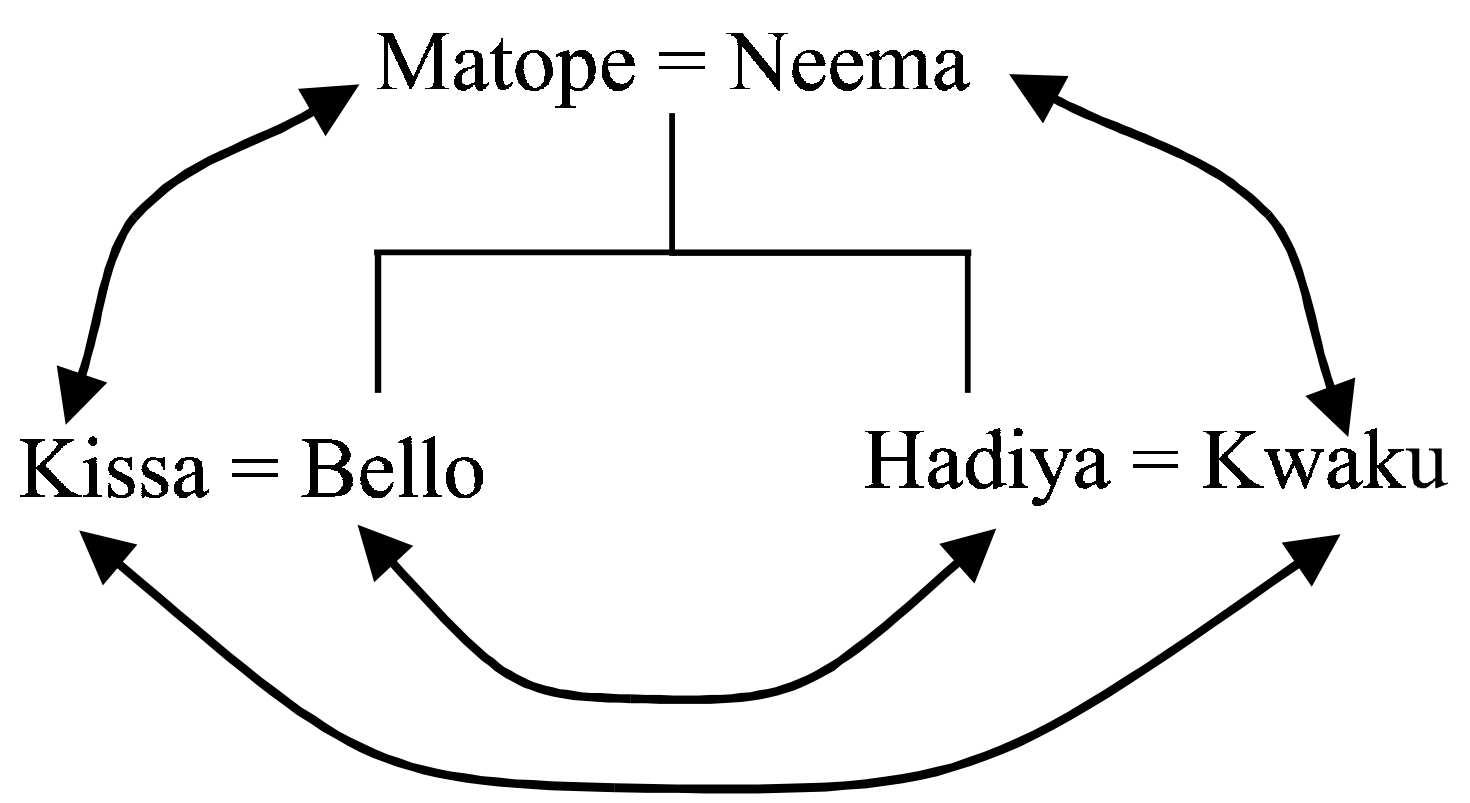

Figure 5-23. The 11 relationships between the 6 African people. Details provided in Appendix C.

Of course, by eliminating the cross-domain pairings, I had effectively removed the examples of analogy-making. There were intra-domain examples, but not a single example of a cross-domain analogy. What was needed were other examples of cross-domain analogies. To this end, I introduced other families and their trees, and learned to make analogies between *them*. The training details were as follows. In order to have enough cross-domain analogies to learn from, I introduced 4 additional families for a total of 6. To keep the number of total people relatively small, I limited each family to 6 members. I defined 11 relationships between them. The relations were: *brother, sister, mother, father, son, daughter, wife, husband, in-law parent,* and *in-law child* (see Figure 5-23). Appendix C lists all of the relationship facts for one family.

All of the families had isomorphic relationships. I paired all of the relationships from all families to those of all other families, except for those relations involving families 1 and 6 (see Figure 5-24). These two families were never paired together. That is, if family #1 was the English family, and family #6 was the new African family, then no pairing ever involved people from those two families together. There were a total of 1,768 pairings that did not involve the English and the Africans together, and 104 pairings that involved only the English and African families (5.9% of total). The 104 were not trained on, but were saved to test for generalization.

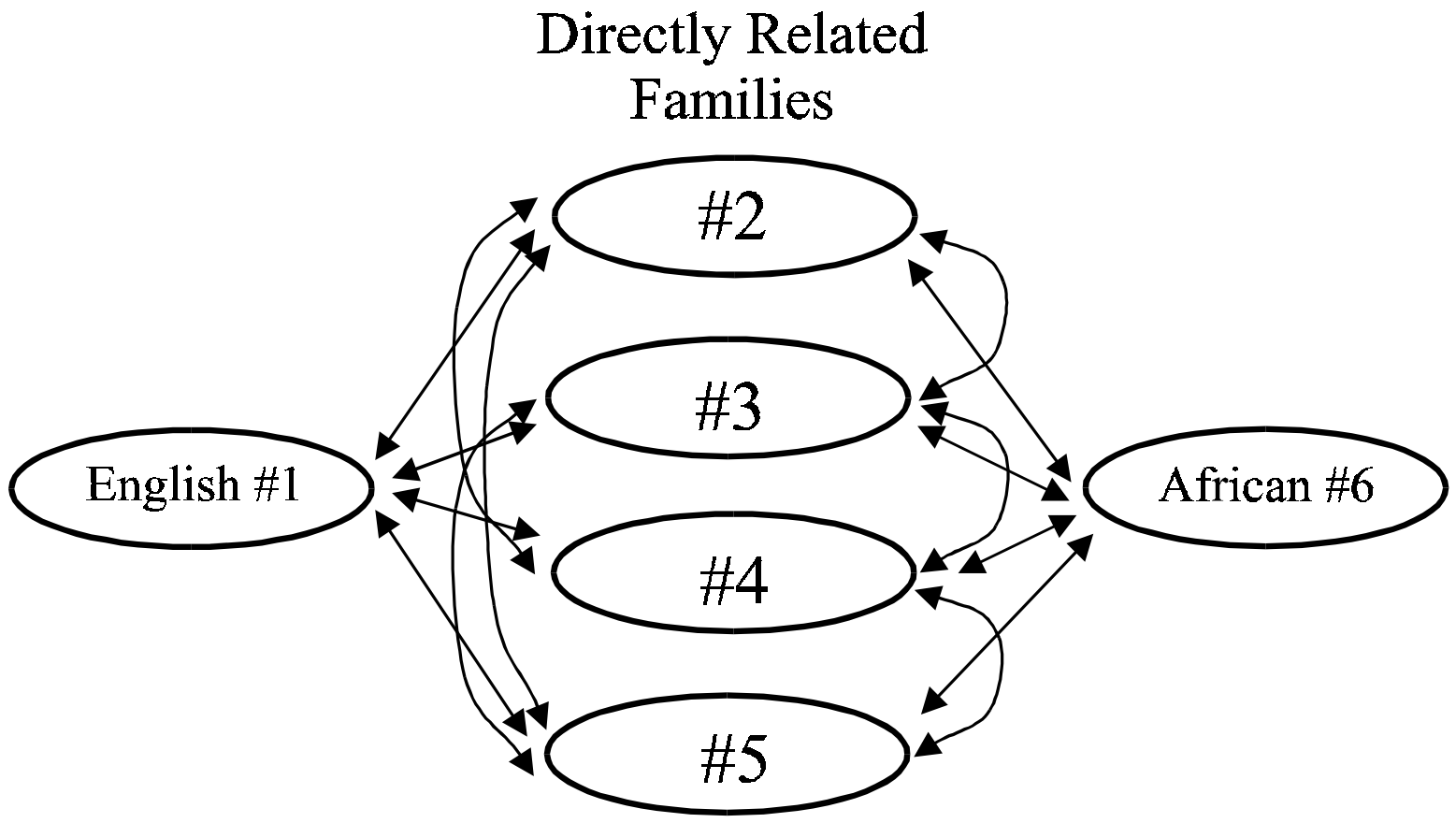

Figure 5-24. The 6 families used in Experiment #3b. Each arrow represents the 104 source-target pairings between two families. The English family was trained to make analogies between the 4 directly related families, as was the African family. Notice that there were no direct associations made between the African and English families. The question was, would the network know how to make analogies between the African and English families?

I trained a network in the Analogator style on the complete set of 1,768 source-target pairs. Training went relatively slowly, reaching 99% correct in 530 epochs. The question, of course, was what would happen with the 104 relationships between the English and the Africans? When tested, the network got 88.5% correct – significantly better than the 2.8% that one would expect from random guessing, and about the same level of performance as Hinton achieved with his intra-domain family tree analogies. Therefore, the Analogator network was able to make cross-domain analogies in a non-spatial domain when given cross-domain examples.

How was the network able to generalize over the relations between two sets of people when it had not been trained on any of pairings? One might speculate that the network was using “relationship transitivity.” For example, the African family could be mapped onto its analogs in, say, family #3. And family #3 could be mapped onto its analogs in the English family. By transitivity, the network could map the Africans onto the English. This may be a useful description of how the network was behaving, but one would be hard pressed to find an implementation of that description in the activations of the network. In addition, I had tried learning the task with a single directly-related family, but that did not generalize.

Knowing that it took a few directly-related families before the network would generalize gives a clue to how the network might have learned to perform the task. I believe that there needed to be a sufficient number of relationships to force the network to develop internal representations that shared units in an appropriate way. In this manner, there were internal representations forming during training which were taking advantage of the isomorphisms between the families. When the network was finally tested on novel pairings, the network was not searching through structures looking for analogous components. Rather, the analogous components had, in some sense, already been found and exploited during the training of the network.

## 5.4 Summary of Experiments

This chapter has explored three domains using the Analogator architecture and training method. The letter-part and geometric-spatial domains are both based on visually oriented, spatially structured stimuli, while the family tree domain is based on symbolic, structureless facts. However, the letter-part analogies are actually quite different from those in the geometric-spatial domain; whereas the letter-part analogies deal with making correspondences between two well-known parts, geometric-spatial analogies deal with many correspondences between pieces whose roles must be determined on-the-fly. In effect, all three domains are fundamentally different from one another.

Since the Analogator architecture successfully learned to make analogies in all three domains, one might be tempted to look for isomorphisms between them. However, I believe that Analogator's performance is less about the similarity of the domains than it is about the similarity of *similarity*. That is, I believe that there is something fundamental between two similar situations or scenes, no matter what the domain. Our concept of "sameness" captures that notion and applies whether one is considering art, speech, music, sentences, unicorns, or baseballs. If there is an isomorphism between the three domains examined in this chapter, then I believe it can be found in the manner that figure-ground pairs are seen as being the same. Of course, the Analogator architecture and the recurrent figure-ground training procedure were designed to make the most of that similarity.

On the other hand, Analogator was unable to learn all of the different variations of the analogy problems. Analogator had difficulty learning to generalize over certain invariances, specifically the sliding up and down of the letter-part analogies, and dealing with the many ways of twisting geometric-spatial analogies. These limitations will be discussed in Chapter 7.

At this point, one might want to ask: "How do we know that Analogator's performance is not due to the straightforward kind of similarity-based generalization that we have associated with backprop systems for years?"[16] Before we can answer this question, we must first examine the notion of "similarity-based generalization." Connectionist networks have shown for a decade and a half that they are capable of learning to make generalizations based on features and attributes (see, for example, Sejnowski and Rosenburg, 1987). Let us call those types of generalizations "similarity-based" as they can be seen as a direct measure of similarity of the input features. Analogy-making is typically seen as not being based on low-level features, but based on higher-level relations between multiple objects. The distinction between "features" and "relations" seems clear-cut, especially when one has representations that are created based on exactly these distinctions (as do most models, including Analogator). However, this distinction is quite blurry and artificial (Hofstadter *et al.*, 1995).

---

[16] Thanks to Rob Goldstone for this question.

The distinction between the feature of a single object and the relation between multiple objects is particularly (and intentionally) blurry in Analogator. Consider the representation used in encoding scenes of geometric shapes described in Chapter 4. Although the notions of "attribute" and "object" are well defined, the representations of "circle of shapes" (a relation between multiple objects) and "circle" (an attribute of a single object) can be very similar, or even identical.

Nevertheless, one may still be left with the feeling that Analogator is performing the same old interpolation-type of generalization, rather than the extrapolation-type of generalization we would expect from analogy-making. For example, an analogy made between *birds* and *airplanes* seems convincingly "relational" rather than "featural." Such a comparison demonstrates abstract knowledge of a concept, such as CONTROL-SOURCE, independent of domain. In one sense, this is exactly what the family tree cross-domain analogies show.  Recall the 6 families of Figure 5-24. None of the 6 families actually name the relations between the members of a family. Therefore, one could imagine that Family #1 actually represents a bird (or the parts and relations of a letter 'a'). Likewise, Family #6 could actually represent a plane (or objects in a geometric scene). Recall that Analogator was never trained on making any connection between Families #1 and #6, and yet, it was able to make appropriate correspondences between them. By doing so, Analogator can be seen as making very abstract analogies.

Can the amount of "abstractness" of Analogator's analogies be qualified? To do so would require a detailed examination of exactly the types of problems used in training an Analogator network and the representations used. Furthermore, an analogy problem may be abstract by relying on identifying and mapping corresponding relations, but a network may solve such a problem in a non-relational manner. In fact, I believe that collapsing such abstract, relational analogy problems onto concrete, featural similarity judgments is exactly how Analogator bridges the gap. Is this an inherent weakness, or a fundamental strength? This question, I believe, can only be answered by further empirical tests to see if such a system can scale up to larger problems.

In summary, the experiments described in this chapter demonstrated that the figure-ground associating process was capable of learning to make analogies in widely differing domains. In addition, the methodology was shown to train much more quickly than comparative feed-forward networks performing the same task. Most importantly, the networks trained in this style exhibited good generalization ability inside and across domains. In the next chapter we compare Analogator with other models of analogy-making.

# 6 Comparisons with other Models of Analogy-Making

*There is no word which is used more loosely or in a greater variety of senses, than 'Analogy'.*

-J.S. Mill

As J.S. Mill has noted, the word 'analogy' has been used in many ways. Even the 'experts' use the term in various senses; I suspect that no word has been used more often to describe cognitive models than 'Analogical'. Much has been written describing computation models in the areas of "Analogical Reasoning", "Analogical Mapping", "Analogical Learning", and "Analogical Problem Solving". Each of these areas defines and examines analogy-making from a slightly different perspective. Although not unrelated, analogy-making models from one camp often have little in common with models from another.

Many of the differences have arisen between theories in the various camps because of very different initial assumptions made about analogy-making, as noted in Chapter 1. Consider this statement made by Keane, Ledgway, and Duff (1994): "In order to solve a problem by analogy one must first represent the problem in some form." All researchers would probably agree that this is the first step in creating a computational model of analogy-making. However, as we have seen, the exact details of the initial representation have a large impact on exactly what problem is to be solved by the analogical model. The assumptions about the initial representations divide computational models into two major camps: the traditionalists, and, for lack of a better term, the non-traditionalists. This chapter will examine these two groups and their specific computational models.

## 6.1 Traditional Analogical Computer Models

In Chapter 1, I introduced the assumptions made by the traditionalists. Recall the assumptions:

1. *Analogy-making begins with two structures.*
2. *Analogy-making is a search through the structures in an attempt to find analogous parts.*
3. *Syntax alone determines the similarity between any two objects, attributes, or relations.*

4. *For any two relations to be seen as analogous, they must exactly match in terms of their number of arguments, and types of arguments*.

5. *Relations, attributes, and objects are forever distinctly different things*.

6. *Context plays no part in making the analogy*.

7. *The result of making an analogy is the creation of a mapping between corresponding pieces of the two structures*.

In a nutshell, the traditional view defines "analogy" as a special-purpose module that takes two structures as arguments and returns a set of links connecting analogous items. Analogy-making is construed as a search for the set of coherent correspondences between two structured representations. The set of coherent correspondences connects items from two domains. Often, one domain is considered better defined and is called the source domain or the base domain. The other is called the target domain, and the mapping of the analogy to the target is called *analogical transfer*.

In modeling analogy-making, the traditionalists have not been limited to GOFAI ("Good Old Fashioned Artificial Intelligence", Haugeland, 1985) symbolic techniques. Many researchers have used more modern mechanisms, such as connectionist networks. The following section discusses those models, connectionist and symbolic, that make some, or all, of the traditional assumptions.

### 6.1.1 Structure Matching Engine

Gentner's theories of structure-mapping (Gentner, 1980; Gentner, 1982; Gentner, 1983; Gentner and Gentner, 1983; Gentner, 1986; Gentner, 1988) are well-known in cognitive science. Her theories have been implemented in a model called the Structure

Mapping Engine, or SME (Falkenhainer, Forbus, and Gentner, 1986). SME may be the best known of all the analogy-making models and is the prototypical traditional model.

The intuition behind Gentner's structure-mapping theory "is that an analogy is a mapping of knowledge from one domain (the base) into another (the target) which conveys that a system of relations that hold among the base objects also holds among the target objects." (Gentner, Markman, Ratterman, and Kotovsky, 1990). Of central importance to Gentner's structure-mapping theory is the notion of *systematicity*: "… People prefer to map connected *systems of relations* governed by higher-order relations with inferential import, rather than isolated predicates." (Gentner, 1989). A *higher-order relation* is a relationship that has as an argument another relation. Therefore, those relations that connect other relations are considered more important in an analogical mapping.

Gentner's main focus is the explanation of an analogy after a target has been retrieved from memory. As an example, consider the situation diagramed in Figure 6-1. On the left sits a large, full beaker of water connected to a nearly empty, small vial via an open tube. On the right is a warm cup of coffee containing a silver bar with an ice cube on the end. The Water-flow situation is to be considered well understood. That is, it is known that water will begin to flow into the small vial because of the unequal pressure. The question is then: What will happen in the Heat-flow situation?

SME processes the analogy as follows. First, representations of the *relevant information* are constructed from the source and target domains. In this case, the Water-flow information forms the source domain, and the Heat-flow information forms the target domain. Breaking the two domains into the object, attribute, and relation types, one might end up with a representation similar to that shown in Figure 6-2. Notice that the relation CAUSE in the source domain is a higher-level relation having the relation GREATER as one of its arguments.

It is worth noting at this point that this initial step of SME, which Hofstadter calls "gist extraction" (1995), is performed by the researcher. Humans typically find it very easy to consider Figure 6-2 and Figure 6-3 as being, in some sense, equivalent. However, it is important to realize how much mental processing occurs in order to make the leap from the pictures to the gists. There are myriad details that can be seen or inferred from the pictures that did not make it into the small, structured representations (i.e. coffee is black, ice is made out of water, water is clear, the cup is sitting on a table, people sometimes drink coffee in the morning, etc.) Also, it is only in this specific context that we might consider the gists in Figure 6-2 to be the “correct” ones.

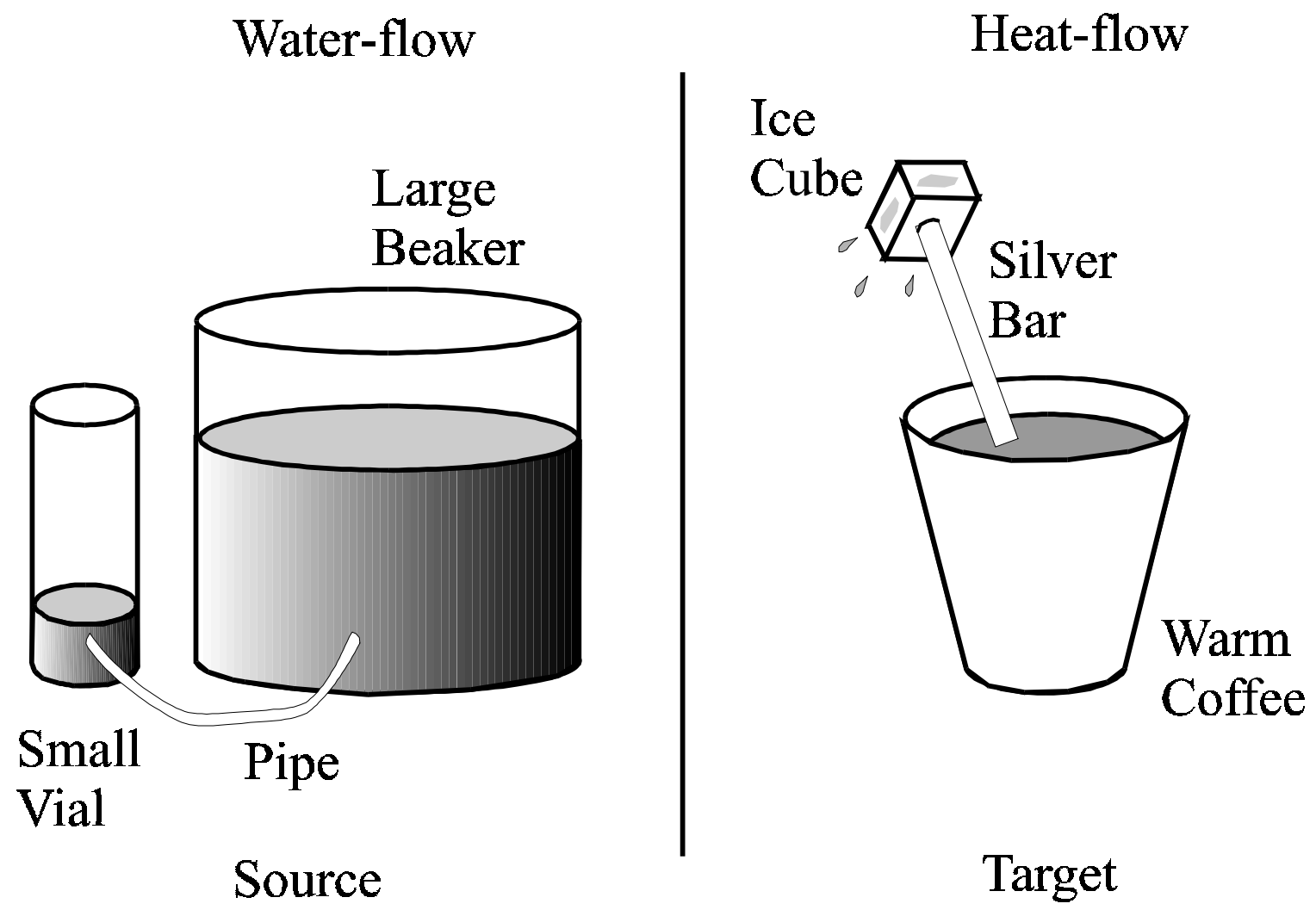

Figure 6-1. A problem for SME.

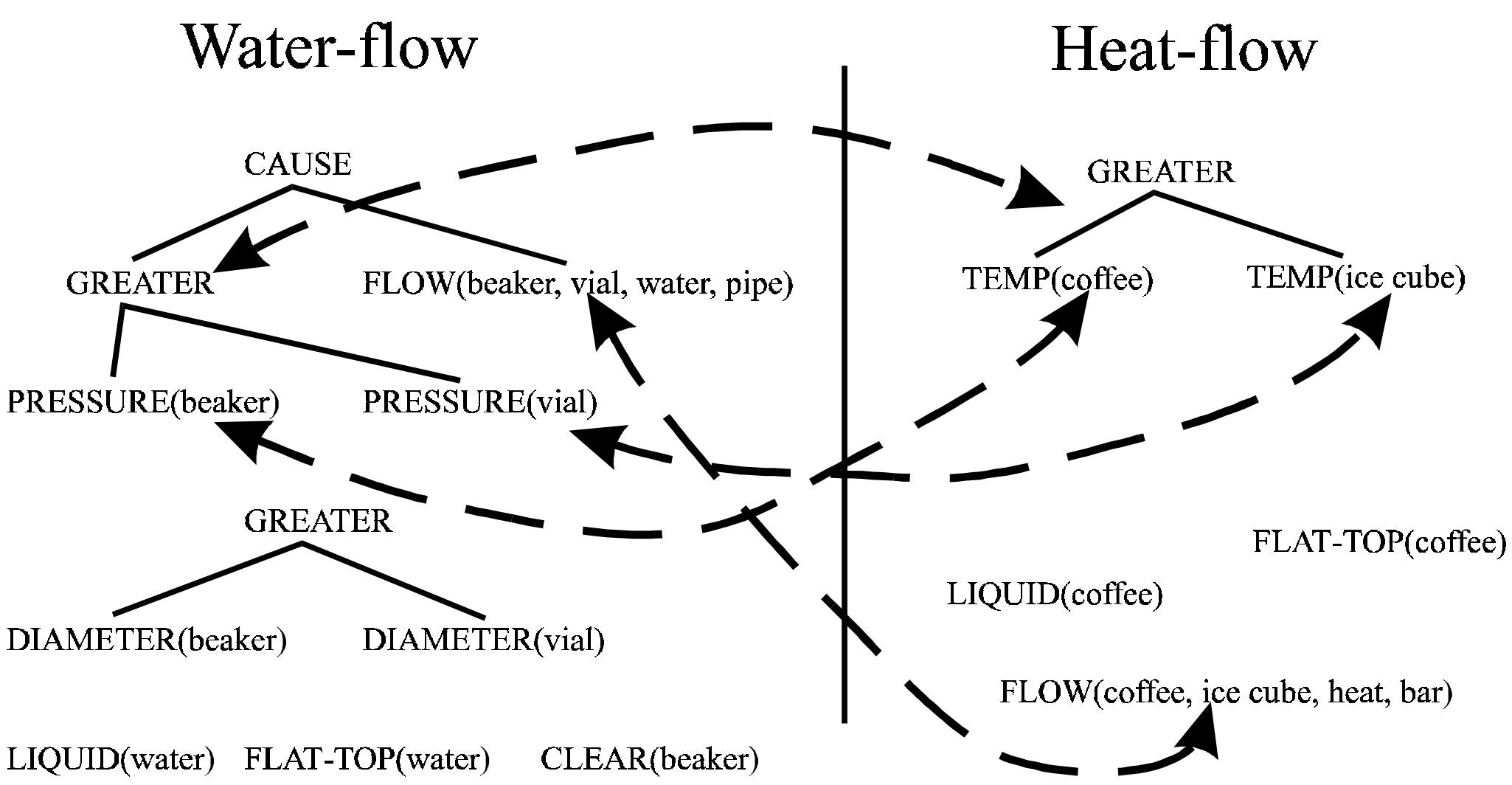

Figure 6-2. The problem from Figure 6-1 boiled down to its gists. The dashed lines indicate a valid mapping from source to target.

In the next step, SME is handed the representations of the two domains. SME then begins searching for possible local matching relations. This is dictated by a set of simple rules:

1. If two relations have the same name, create a match hypothesis
2. For every match hypothesis between relations, check their arguments. If both are objects, or both are relations then create a match hypothesis between them.

(after Gentner, 1989)

Matches are made between all possible entities. A matching is only possible between relations with identical numbers of arguments. For instance, the relation FLOW would not be even considered for comparison to the relation TEMP. Also, matches are not attempted between two entities unless they are of the same type (i.e., object or relation). Each match hypothesis is given a score based on how good the match is. Scores are based on whether or not two entities have the same name, etc. Next, processing turns toward finding global matches. A global match is a system of matches that is consistent (systematic). Match hypotheses turn into full-fledged matches if their scoring is high enough. Finally, from the set of matches, a single coherent set of a mappings is chosen based on its total matching score, and the analogy is "understood" (the gray arrows in Figure 6-1). In the example shown in Figure 6-1, **beaker** would be found to map to **coffee**, **vial** to **ice cube**, **water** to **heat**, and **pipe** to **bar**. Transferring the well-known information from the Water-flow to the Heat-flow situation produces the previous unknown relation:

CAUSE( GREATER( TEMP(**coffee**), TEMP(**ice cube**) ), FLOW(**coffee, ice cube, heat, bar**) )

That is, "the coffee's warmer temperature causes heat to flow up the bar to the ice cube."

SME's basic algorithm has provided the foundation for many extensions (see for example, Gentner and Forbus, 1991; Forbus and Oblinger, 1990), and as the basis for many models in psychology and artificial intelligence (see for example, Fallkenhainer, 1987; Bhansali and Harandi, 1997). In addition, Gentner's structure mapping theories provide the foundation for many other models, including some of the most recent connectionist models (see Gentner and Markman, 1993, for a discussion of analogy-making directed at connectionists; see also, Plate, 1991; Kanerva, 1996; Handler and Cooper, 1993).

In summary, SME is a serial *generate-test-and-select* algorithm based on the syntax of pre-formed representations. It narrowly defines analogy-making as a specialized module that has no connection to other processes of similarity, such as categorization. Context plays no role in the model. It is deterministic, and does not incorporate learning.

As its focus is on the psychological explanation of mapping, it probably will not scale-up well as it requires all possible matches to be evaluated.

### 6.1.2 Analogical Constraint Matching Engine

The Analogical Constraint Matching Engine (ACME) created by Holyoak and Thagard (1989) was the first connectionist analogy-making model. However, it has far more in common with SME than Analogator. ACME, like SME, also creates mappings between symbolic structures, and, in fact, has been tested on many of the same problems that SME has tackled. ACME solves analogy problems by allowing activation to settle in a localist network with weighted links acting as soft constraints.

The ACME algorithm works as follows. First, like SME, the structured representations of the source and target domains are designed by the researchers and fed into the analogical engine. ACME then creates a node for each possible corresponding pair, although some mappings are not allowed based on syntactic constraints (see Figure 6-3). For example, an object cannot map onto a relation, and so that pairing would not receive a node. The nodes explicitly represent all possible analogous pairings. Weighted links are then created between pairs of nodes. Links allow activation to flow through the network. The links can have positive weightings if the pairings support each other, or may be negatively weighted if the pairings are mutually exclusive. The network also has positive links connecting all pairs of relations based on their *a priori*-judged similarity. The researchers assign these links, called *semantic links*, in advance. For example, one may decide that the relations FLOW and FLUSH are very similar, and so should tend to be mutually excitatory. Finally, there is another set of links representing other important context and goal information. These links are called *pragmatic links*, and are also assigned by the designer prior to processing.

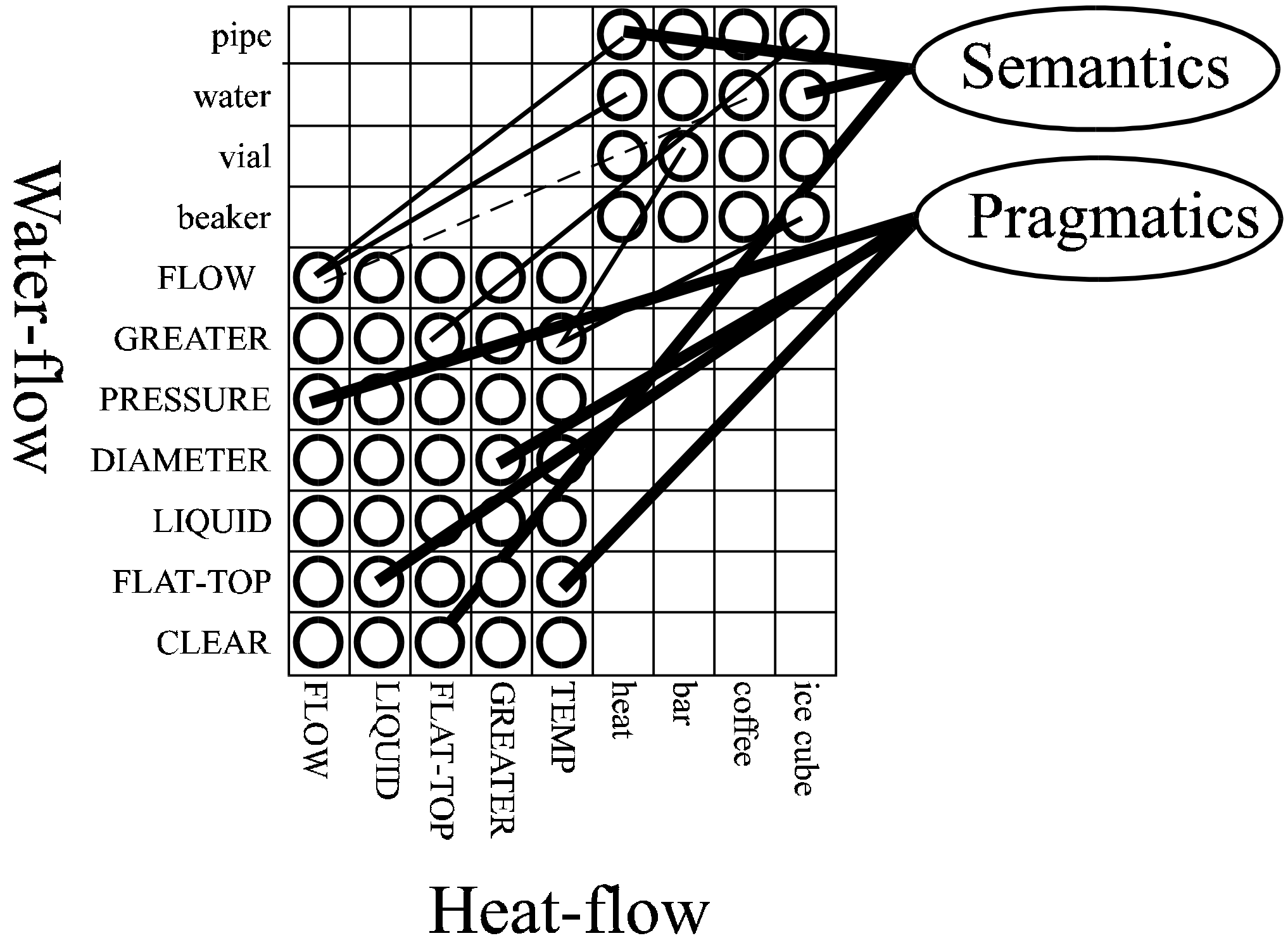

Figure 6-3. An example ACME network. This network depicts ACME's representation of Figure 6-1. ACME actually takes as input representations similar to those shown in Figure 6-2. It then constructs a network similar to that shown here. Not all links are shown.

Once the network has been established activation is allowed to ebb and flow over the links as defined by a relaxation algorithm. If the network settles into a fixed state, a solution has been found for the analogy problem. Researchers have claimed that ACME "has achieved considerable success in simulating human intuitions about the natural mappings for a broad range of cases, including examples of scientific explanatory analogies, story analogies, problem analogies, metaphors, analogical arguments, and analogies between political situations" (Hummel, Burns, and Holyoak, 1994). Again, it is important to realize the distinction between a situation and its gist. It could be the case that the representations used for making an analogy between Vietnam and the United States in the 1960's are exactly the same symbolic structures as used in an analogy between the solar system and the Rutherford atom. The names of the symbols might be different, but, of course, symbols' names do not effect their use in the model. We will further discuss this issue later in this chapter.

One of the major criticisms against ACME has been its psychologically implausible requirement to allocate a node for every possible analogous correspondence (Hofstadter and Mitchell, 1994).[17] Hummel, Burns, and Holyoak (1994) have addressed this issue by applying dynamic binding via synchrony of neural firing (von der Malsburg, 1981) to analogy-making. Hummel *et al*. (1994) have shown that an ACME-like connectionist network can make analogies via dynamic binding without explicitly representing all possible matches. This is an important improvement over ACME. ACME has been extended in many other ways (Holyoak, 1994; Nelson, 1994). However, all of the extensions rely on the pre-structured initial representations.

[17] It should be noted that many of the traditional models consider all possible pairings. SME, for example, explicitly calculates scores for all legal, possible matches.

In summary, ACME is a parallel constraint satisfaction network. It solves analogy problems via a relaxation algorithm that satisfies many soft constraints in parallel. Context may play a part in the analogy; however, goals and other contextual pragmatic information must be set by the researcher. ACME does not incorporate learning. In its most basic design it does not scale well in terms of capacity as it requires a node for all possible pairings.

### 6.1.3 Incremental Analogy Machine

Keane's Incremental Analogy Machine (IAM; Keane, 1994) adds a serial processing dimension to analogy-making not found in SME or ACME.[18] IAM has the ability to eliminate many possible matches by incrementally establishing further constraints. By starting with a few "seed" matches, IAM limits the number of pairings that it might otherwise consider. If those seeds turn out to not pan out, then the system is capable of backtracking to find other seeds, and incrementally building coherent structures from those.

Although Keane (1994) has identified evidence of such serial processing in humans, his model still requires researchers to enter the initial representations.

### 6.1.4 Similarity as Interactive Activation and Mapping

Goldstone (1991) has applied some of the core ideas from SME and ACME to the problem of similarity judgment. Similarity judgment has traditionally been viewed as a simple calculation based on a comparison between two static lists of features, or between two static lists of dimension values. Goldstone's model, Similarity as Interactive Activation and Mapping (SIAM), accurately predicts subtle similarity judgment choices made by humans (Goldstone and Medin, 1994). In addition, SIAM makes accurate predictions about serial processing effects, like IAM does.

Although the inputs to SIAM are as static, rigid, and explicit as those used by ACME, SME and IAM, SIAM does allow for attribute values to dynamically become aligned during processing. Although this ability has limited effect, it allows SIAM more flexibility than the other traditional models discussed so far. Recall that SIAM was designed to only model the judgement of similarity, while the other models were specifically designed to process analogies. However, Goldstone (1991) introduced the ability for SIAM to not only make attribute-to-attribute and object-to-object connections, but also role-to-role connections as well. Although this could lead to a model capable of making more abstract similarity measures, or even analogies, it is still based on static, explicit representations.

---

[18] Forbus, Ferguson, and Gentner (1994) have created an incremental version of SME (called Incremental SME, or I-SME) in response to Keane's suggestions (1994). ACME could also have such an incremental mechanism added. The criticisms made regarding ACME and SME's representations remain, however. Keane (1994) provides a detailed comparison between IAM, SME and ACME.

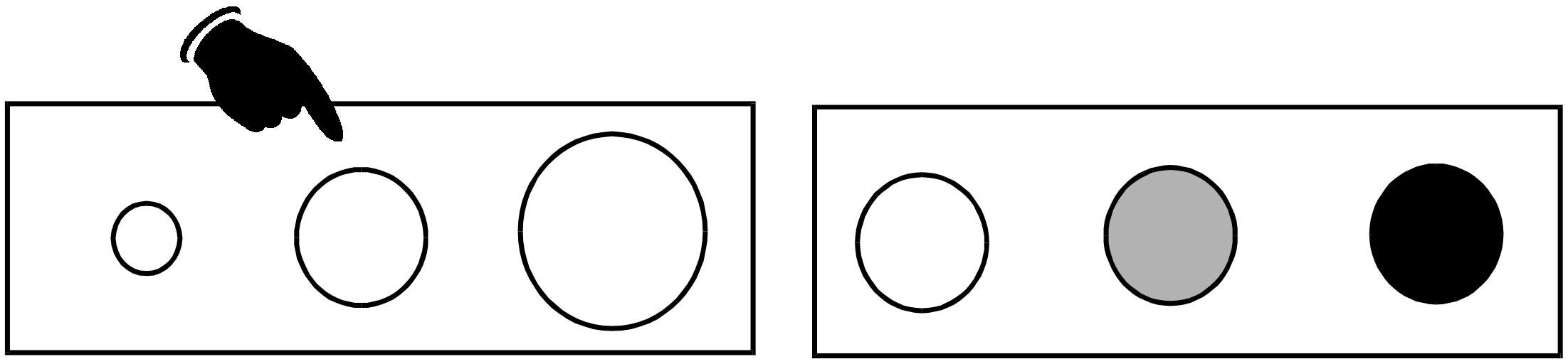

Figure 6-4. A sample problem from Handler and Cooper's SMERF domain (1993).

### 6.1.5 SMERF

Handler and Cooper (1993) have implemented a connectionist model that is capable of solving problems similar to those that lie in Analogator's geometric-spatial domain (see Figure 6-1). Their system, called SMERF, is meant to be an instantiation of Gentner's structure-mapping theory, and, not surprisingly, is very similar to SME in many respects. Like SME, SMERF also attempts to make a match between every pair of predicates. Like its connectionist cousin ACME, it does not incorporate learning. SMERF uses an algorithm that is very similar in style to the constraint satisfaction algorithm used by Holyoak and Thagard (1989). However, Handler and Cooper have adapted Markov Random Fields (Kindermann and Snell, 1980) to satisfy their constraints.

One point of similarity with Analogator is the form of the representations of relations. Rather than having explicitly structured representations like SME and ACME, SMERF's input is simply a set of features, one set for each object. They claim that their representation is better than that of ACME or SME because it does not explicitly represent relations between objects (Handler and Cooper, 1993). However, their solution has been to simply convert a relation between two objects, such as LARGER-THAN, into two values along a dimension. Instead of having explicitly structured representations, such as LARGER-THAN(A, B), they encode the size difference as values along a dimension, such as SIZE(A, .8) and SIZE(B, .4). Although, this does make the representation implicitly structured and is similar in spirit to the representation used by Analogator, their model bypasses a major operation: perception. Their (unstated) assumption is that they do not need to have perception in the model as long as they can represent relevant attributes as scalar values along a dimension. This works fine for linear attributes, such as size and shading (two dimensions they happen to model). But this methodology does not necessarily work for dimensions such as color or position in two dimensions. Their representation is better than ACME's only in that they refuse to use explicitly structured representations, even if it means avoiding the problems that would seem to require them. Because they use values along a dimension, their model is similar to Goldstone's SIAM. However, it is doubtful that SMERF could account for the subtle judgment of similarity effects that SIAM does.

Like ACME, SMERF explicitly represents object-to-object mappings. The mappings are created in a constraint-satisfaction-like manner based on prior probabilities and weightings provided by the researcher. This information dictates to the network how to weigh the attribute dimensions in order to make the mapping.

Although they claim that their model could "naturally incorporate real perceptual input," I do not see how. First, their network is created based on the number of objects in a scene. Adding another object to a scene would require a completely different network. In addition, placing a fourth object into a scene already containing three would require a network nearly twice as large. Secondly, their network is incapable of creating, or perceiving, groupings within a scene. For example, a pair of circles could not be perceived as such. As mentioned, their method of dealing with structure is to simply avoid it. Finally, SMERF is incapable of representing context, let alone being influenced by it. For these reasons, I find it doubtful that SMERF could accommodate real perceptual input.

### 6.1.6 Other Connectionist Traditional Implementations

Recently, many connectionist models have been designed to study analogy-making. These traditional connectionist models fall roughly into two categories: those that are representation-centered, and those that are process-centered. The representation-centered projects have focused on distributed representations, and how they can naturally be used in analogy-making and similarity judgment. The process-centered projects have focused on incorporating connectionist mechanisms into large, analogy-making systems.

Kanerva's spattercode (1996), and Plate's Holographic Reduced Representations (HRR, 1991) are two connectionist representation-centered projects. In fact, the spattercode is equivalent to a HRR with certain limitations (Kanerva, 1996). Both are techniques for representing a collection of variables and their values, much like Smolenky's tensor product. However, both the spattercode and the HRR are able to bind sets of variables to their values without increasing the dimensionality of the representation. That is, all resulting representations remain the same size as the originating patterns.

In addition to this ability, Plate has applied the HRR to the task of estimating analogical similarity (Plate, 1993).[19] Plate's goal was to devise a method that could quickly and efficiently estimate the similarity between two structured representations. This is a required step in Gentner and Forbus' MAC/FAC extension (1991) to SME.

First, we will describe the problem as presented by Gentner and Markman (1992). Imagine that we have many situations like the Water-flow and the Heat-flow representations from Figure 6-2. Imagine that we now have a new situation. How can we quickly find the old situations that are most similar to the new situation? Plate found that if he encoded two situations via an enhanced HRR, the dot-product between the two can be used as a direct metric for their structural similarity. This appears as a much better solution than that originally proposed by Gentner and Forbus (1991). However, there are two problems with this methodology. The first is that this technique requires calculating the dot-product between the target and all sources in memory. This may be faster than other methods, however, it is still too computational intensive as a viable model of memory retrieval. The second problem is that the representations are still the same old prefabricated symbols and structures. This methodology, like most of the traditional models, leaves little room for perceptual and contextual effects.

---

[19] It is assumed that the spattercode is equally capable of performing the tasks to which Plate applied the HRR.

Kanerva (personal communication) has suggested that the spattercode and HRRs may be able to be built from low-level sensory stimuli rather than from traditional symbolic representations. If this could occur, and can reflect structure and contextual pressures, this could be a very useful mechanism. However, without those abilities I don't believe HRRs will advance the state-of-the-art of analogy-making.

Halford, Wilson, Guo, Gaylor, Wiles, and Stewart's STAR model (1994) is another connectionist representation-centered project, and is similar to the spattercode and HRRs. STAR uses Smolensky's tensor product to create representations of the source and target scenarios. The STAR model uses a tensor of rank-3 to represent a predicate of two arguments, one for the predicate, and one each for the two arguments. In that manner, the relation *ABOVE(square, circle)* would be a three-way binding of *ABOVE*, *square*, and *circle*. STAR, like the HRRs and spattercode, uses explicitly structured representations, and therefore, places it squarely in the traditional camp.

Although STAR, HRR, and the spattercoding use explicit structures, all of them allow for holistic connectionist processes to operate on their distributed patterns. Such a holistic process might be able to learn to manipulate the patterns without explicitly decomposing the patterns into their constituent components (see Chalmers, 1990, and Blank, Meeden, Marshall, 1992). For example, a network could learn to extract the representation of the figure given some context, much like Analogator does.

One might wonder whether representations of this type could be used as the input representation in an Analogator network. I think that this prospect is unlikely to work with HRRs and the spattercode. HRRs are complex encoding schemes with varying patterns for each element being represented. That is, the representation of **square** in the relation OVER(**square, triangle**) would likely be very different from the representation of **square** in the relation OVER(**square, circle**). In Analogator the representation of any object or person was always the same (on the input layer) no matter what the relation was.

Because STAR contains explicit patterns for relations, I think that using its representation as inputs to Analogator would prevent generalization. As an example, consider the family tree cross-domain analogies from Chapter 5. If the relationships had had explicit patterns, the network would have "paid attention" to them, as they relay useful information. However, in order to make cross-domain analogies, the network needs to ignore the specific relations, and look instead to the implicit structure of the relations. I believe that using implicit relations forces Analogator to learn the structure; ironically, providing more information (i.e., the relation names) would make Analogator unable to generalize across domains.

The second group of traditional connectionist models are those focused on process. Barnden's ABR-Conposit (1994), Eskridge's ASTRA (1994), and Kokinov's AMBR (1994) are all complex, hybrid analogy-making systems revolving around connectionist mechanisms. All of them incorporate spreading activation and marker passing as methods of coordinating processing between symbolic and connectionist subsystems. All three models are impressively complex systems that have shown how symbolic and subsymbolic processes may operate in concert. Unfortunately, all three systems also subscribe to the traditional assumptions regarding pre-structured representations. Although all of these three models incorporate learning, none of them address the problem of how the analogy-making mechanism could develop.

## 6.2 Non-traditional Analogical Computer Models

Recall that the non-traditional models do not subscribe to the traditional assumptions. Like the traditional models, the non-traditional models come in GOFAI and connectionist versions.

### 6.2.1 Evans' ANALOGY program

The first AI program to address head-on the issues of analogy-making was Evans' ANALOGY program (1968). ANALOGY was designed to answer proportional geometric analogy problems (see Figure 6-5) that were, at one time, common on American I.Q. tests. The idea of the ANALOGY task is to answer the question "if the picture A changes into picture B, then what does picture C change into?" The question is answered by simply selecting one of five pictures (the bottom row of pictures in Figure 6-5). Based on ANALOGY's abilities, there are at least a couple of answers for Figure 6-5. One possible choice is #4, as it completes the analogous abstraction "removal of the inner object." However, ANALOGY could also have answered choice #2, as it was capable of complex transformations, such as "take the big object and delete it. Then, make the little object bigger."

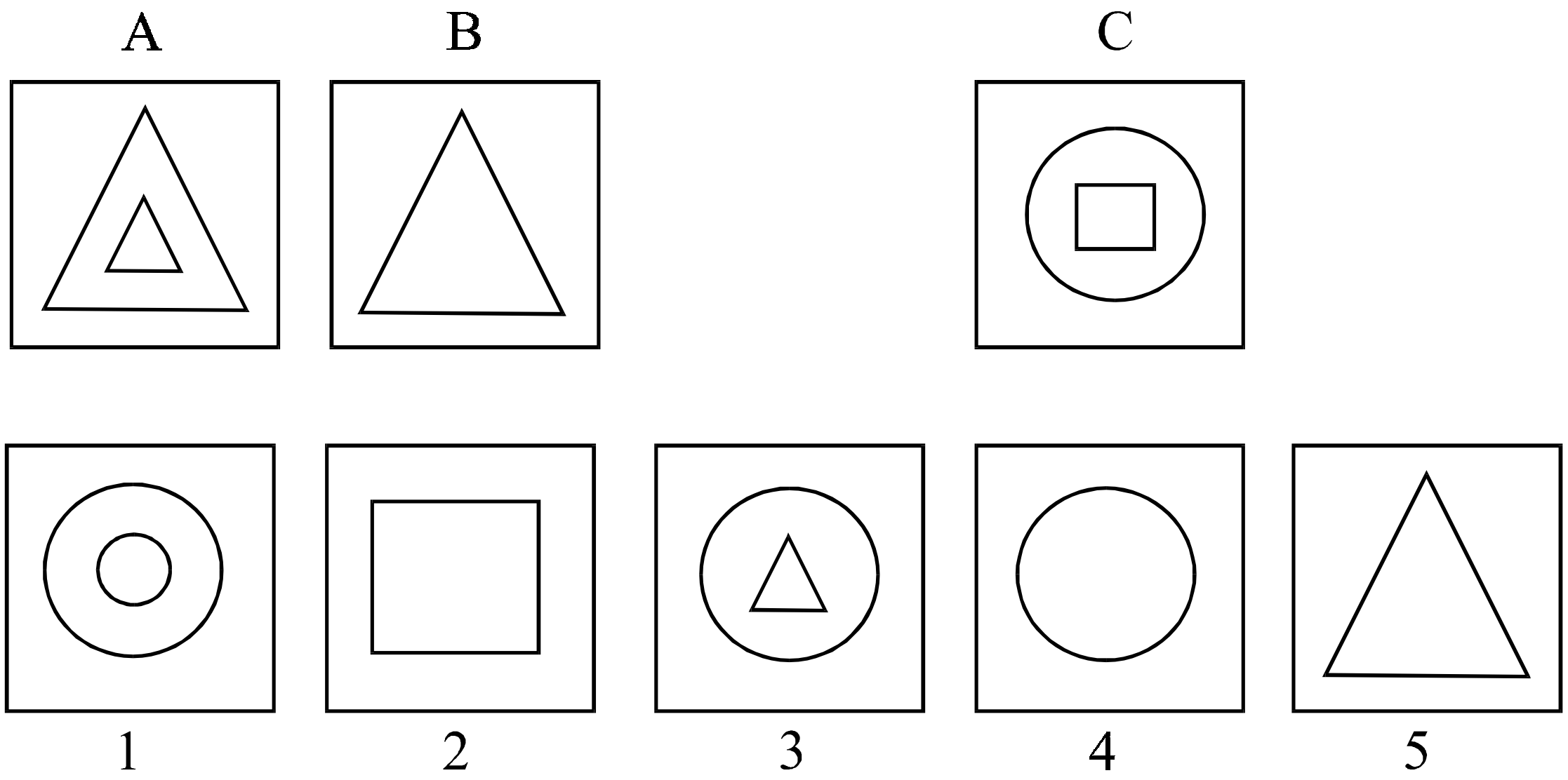

Figure 6-5. A sample problem from Evans' ANALOGY domain.

Unlike any program so far discussed in this chapter, objects and relations in ANALOGY were provided implicitly. For instance, the two triangles in frame A of Figure 6-5 might have been initially represented as:

```
(OBJECT-1
     (LINE .5 .2 .8 .8)
     (LINE .8 .8 .2 .8)
     (LINE .2 .8 .5 .2))
(OBJECT-2
     (LINE .5 .4 .5 .5)
     (LINE .5 .5 .4 .5)
     (LINE .4 .5 .5 .4)).
```

Relations such as *inside*, had to be recognized by the program. The categorization of objects and relations could be slightly effected by context, but this was not a major focus of the model. The size of an object was its only attribute, and was called *scale*. ANALOGY sequentially turned each frame's implicit representation into an explicit representation.

After the creation of explicitly structured representations for all of the frames, a list of possible transformations that could turn picture A into picture B was then created. Any series of transformations that could successfully transform A into B was considered as a possible 'rule' with which described the analogy. These rules were then compared with the set of rules that transformed picture C into each of the pictures 1 through 5. Rules that matched, either partially or exactly, were then ranked according to a similarity measure. Partially matching rules were penalized. If a rule described a transformation that left the scene unchanged, then the rule described two scenes that were very similar, and given a high ranking. The various kinds of transformations (e.g., scaling, rotation, and reflection) and their combinations were given lower rankings, as their related scenes were less similar. For instance, a rule that described a simple rotation was ranked higher on the similarity measure than a rule describing a rotation, a scaling, and a reflection.

Once the rankings were made for all of the remaining rules, the one with the highest score won. At the time, ANALOGY was highly regarded and has been touted as one of the classic AI programs. Recently, however, it has had its critics:

> Yet ANALOGY has its serious limitations. Using the domain of geometric figures is the most serious limitation of the program. Because of the many hidden assumptions carried along with such a domain, AI work in analogical reasoning has been misdirected in a number of critical ways.
>
> (Kedar-Cabelli, 1988)

Of course, as ANALOGY doesn't learn, I, too, think that ANALOGY has its limitations. However, I disagree with Kedar-Cabelli as to what those problems are. She states that the domain of geometric figures has "misdirected" the entire study of analogy-making because of "many hidden assumptions." By this, I suspect that she means that ANALOGY was specifically designed for geometrical analogies, and, as such, has relied on techniques that do not transfer to other domains, such as analogies involving solar systems, Vietnam, heat flow, etc. Tabletop (French, 1992), as we shall see, has also been criticized in this manner. The idea that analogy-making should be free of any domain specifics and based on syntax alone has become quite popular in the last decade.

Although ANALOGY was very primitive in many ways, it did touch on issues (such as the perception of objects) that were completely ignored in the field for a couple of decades.

### 6.2.2 Tabletop

As mentioned, Hofstadter and his colleague’s ideas have provided much inspiration for the Analogator project. Their influences on this project have been in three major areas: 1) the idea of analogy as high-level perception, 2) the idea of modeling high-level cognitive behavior via low-level, emergent processes, and 3) the Tabletop domain itself. These three points will be examined in this section.

Tabletop was designed to model *high-level perception* (French, 1992; Hofstadter *et al.*, 1995). High-level perception is the level of perceiving similarity at which concepts begin to play a role. Indeed, the concept of “concept” lies at the center of all of the projects that have come from Hofstadter’s group, the Fluid Analogies Research Group (FARG). As stated previously, I believe the notion that analogy-making can be viewed as the categorization and recognition of conceptual patterns is one of the most important to cognitive science. This simple notion places the emphasis of analogy-making on the perception of situations rather than on a search through rigid representations. Of course, this is exactly the distinction used in this chapter in which to discuss all models of analogy-making.

The Tabletop domain was the inspiration for the geometric-spatial analogy domain used by Analogator. Tabletop operates in a microworld composed of a café tabletop, complete with saucers, cups, spoons, forks, etc. The Tabletop domain is more flexible than that used by Analogator in that the program has the ability to pick any object on either side of the table as the analogous object (see Figure 6-6). For instance, the program can point to the very same cup as the one being pointed to. It is also a richer environment, as each object may have many relationships between other objects.

In Tabletop, conceptual distances between concepts can dynamically change to reflect the how the scene is currently perceived. Tabletop analogy-making is a complex process which can be described at multiple levels. At the lowest level, a series of little programs, called *codelets*, take stochastic turns doing little bits of processing. For instance, one codelet might pick two objects and, if they are currently labeled the same, might suggest that a group be built from them. At a higher-level one might describe the behavior of the whole system as "heading toward" an answer as things look like they are falling into place. In this way, global control emerges from the tiny, random operations of the codelets.

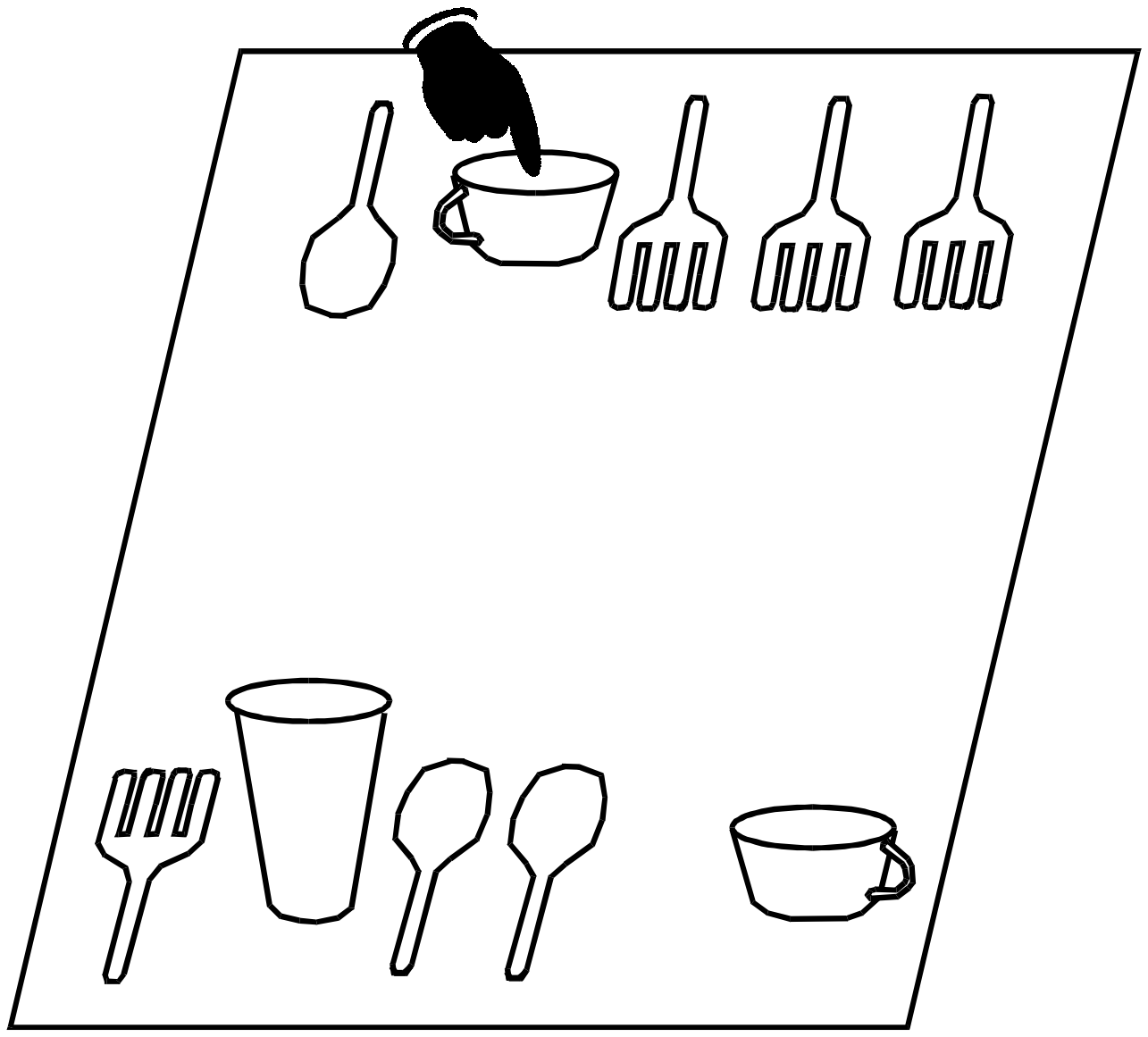

Figure 6-6. A sample problem from the Tabletop domain (French, 1992). Which object is analogous to the cup being pointed to? The cup would be a literal choice, but the glass would be a more abstract one.

Probably the biggest difference between Analogator and Tabletop is the role of concepts in the two systems. Tabletop was designed to operate at the level of concepts, and serves as a model to explore them. On the other hand, Analogator does not explicitly manipulate concepts, but was designed to focus on the learning of analogical behavior. In this context, Tabletop raises many questions: Where do codelets come from? Can they be learned? Where does the semantic knowledge come from? Can it be learned? What type of hardware could perform both low- and high-level perception? Tabletop does not incorporate learning, although it does in the course of running produce and destroy many structures built in working memory.[20]

One common complaint heard about Tabletop, and its sibling Copycat (Mitchell, 1993), is that it is restricted to a single domain whereas other models, such as SME and ACME, are not (Holyoak and Thagard, 1995 pp. 259; Burns, 1995; Morrison and Dietrich, 1995). This is a prime example of the confusion that can be caused by conflating a situation with its gist. As we have seen, a gist is often represented by a set of symbols and relationships; no information is gained by labeling the symbol that represents our solar system as *SOLAR-SYSTEM*. That is, models that use these representations would operate exactly the same if that same symbol were re-label as *DELTA-GAMMA-64* (Hofstadter *et al.*, 1995). Of course, without descriptive labels, such as *SOLAR-SYSTEM*, one would be hard pressed to understand what a model was actually doing. However, by giving the representations meaningful names, one is willing to give much more credit to the system than it actually deserves. What is it, then, that differentiates one domain from another? Unfortunately, the labels assigned to the symbols are often the only criteria used. Mitchell and Hofstadter (1995) appeal to another technique in an attempt to discover the "meaning" behind a symbol labeled *SIGMA*:

> An astute human watching the performance of Copycat and seeing the term "SIGMA" evoked over and over again by the presence of successor relationships and successorship groups in many diverse problems would be likely to make the connection after a while, and then might say, "Oh, I get it – it appears that the Lisp atom 'SIGMA' stands for the idea of successorship."

[20] Metacat is an extension to Copycat (Mitchell, 1993) and incorporates the learning of episodic memories (Marshall, forthcoming).

Of course, this is exactly the method required with connectionist networks, as there aren't any nicely labeled symbols. Hofstadter's models have also been criticized because they operate in a "toy domain" compared to richer domains like solar systems and water-flow-physics (Mitchell and Hofstadter, 1995). Of course, this is the same confusion detailed above manifested in a slightly different light. These are not mere technical arguments, but highlight many of the places that give rise to confusion when very different assumptions are made regarding analogy and representations.

Are the FARG programs and architecture fundamentally different from the traditional models? Burns and Holyoak (1994) have constructed initial representations for ACME to allow it to work on problems in Copycat's domain. After setting up the problem, certain pragmatic and semantic values are clamped and certain structures are added or deleted. The model is capable of getting many of the same answers that the Copycat architecture does. To get a specific answer requires the researcher to assign pragmatic values and tweak the representations, a process that is built into the Copycat and Tabletop architecture. Burns and Holyoak (1994) claim that assigning the pragmatics and tweaking the representations "does not in itself provide ACME with a solution to the problem." They argue that this merely biases the constraint satisfaction process so that the appropriate answer emerges. However, this step is required by the researchers in order for ACME to perceive the analogy problem in a particular way. Of course, as perception is a major focus of the Copycat and Tabletop projects, these steps are critical and included in the models.

In summary, Tabletop is the most psychologically plausible model to date, and the most sophisticated. However, it does not incorporate learning.

### 6.2.3 Connectionist Non-Traditional Implementations

Although Analogator basically defines this category, there have been some non-traditional connectionist models that have addressed analogy-making, if only peripherally. One such example that we have already examined was Hinton's family tree model (1988) discussed in Chapter 5. Another such example, Harris' polysemy model (1994), demonstrates basic analogy-making in a simple feed-forward network. Harris trained sentences from a simple *noun-verb-noun* grammar to be associated with a particular meaning. The goal was to learn the different meanings of the word "over". Over can mean "beyond", "above", or "across", among other things.

Specifically, Harris took a sentence, such as "cow belong (over) hill", and trained it to produce as output the representation for "beyond", the meaning of over in this situation.[21] In the network, each word was represented by a random, orthogonal pattern. The input layer had three banks on the input layer, one for the verb, and two for the nouns. The network was trained on many sentences like those shown in Table 6-1.

Table 6-1. Simple grammar training examples (after Harris, 1994).

| # | Sentence | Status | Meaning |
|---|---|---|---|
| 1) | Cow belongs (over) hill. | Trained | Beyond |
| 2) | Car drives (over) bridge. | Trained | Above |
| 3) | Person lives (over) spot. | Trained | Above |
| 4) | Person lives (over) bridge. | Not Trained | Beyond |

To examine the network's generalization ability, consider sentence #4 in Table 6-1. Notice that it has much in common with sentences #2, and #3. Yet, the meaning of "over" as used in the sentence is actually more similar to that of sentence #1. Harris trained the network on sentences like 1-3, but did not train on sentence #4. Although sentence #4 is more similar to sentences that produced the representation for "above", she found that the network could ignore that superficial fact, and produce the proper output, namely "beyond".

Although this is far removed from the seemingly impressive analogies from other models discussed in this chapter, nothing about this task was built into the network or the representations; the network learned everything. This was a very simple task, however, models such as Harris' provided some of the initial inspiration for the Analogator project.

One non-traditional, connectionist model that did squarely confront more complex problems in analogy-making was Greber *et al.*'s GridFont network (1991). Although GridFont's designers used an incredibly small corpus, had a very impoverished representation, required the network to perform an impossibly difficult task, and used a simple feed-forward network, the model did not perform as badly as one might expect (see Hofstadter *et al.*, 1995, for a less generous critique). Although I believe that almost every aspect of GridFont was poorly designed, I also believe that it had the correct motivation and spirit: it attempted to learn how to make analogies in a given context.

---

[21] The sentence did not actually contain the word "over" but it is included here to make it easier for the reader to understand.

This chapter has examined a few of the traditional and non-traditional models developed over the last 30 years. There are many more analogy-making models than are mentioned here. However, I believe that the selection provided is representative of the rest of those in the field. Having seen how Analogator compares and contrasts to other computational models of analogy-making, we now examine Analogator's limitations, and summarize.

# 7 Conclusion

*You perceive I generalize with intrepidity from single instances.*
*It is the tourist's custom.*

-Mark Twain

To my mind, there are two main reasons for creating a computational model of a cognitive behavior: 1) to help cognitive scientists understand the inner workings of the phenomena by providing necessary concepts and high-level explanations, and 2) to actually create a program capable of performing the phenomena. We might hope that these two goals could be satisfied with a single model, but there is no reason that the goals need be related; a model that is easy for us to understand may not perform well or completely, and one that does perform well may not be easy to understand. I believe that the two camps of analogy-making models described in Chapter 6 are divided by these two conflicting goals. In this way, those that subscribe to the traditional view model analogy-making in terms of easy-to-understand concepts such as “mappings”, “objects”, “relations”, and “structured representations”. On the other hand, those who wish to create more realistic, truly functional, stand-alone models end up with explanations that are nearly as complex as the actual cognitive system being modeled. These models tend to explain analogy-making not in terms of easy-to-understand representations, but rather in terms of complicated, emergent processes.

Tabletop and the rest of the FARG models strike an interesting balance between these two viewpoints, straddling the boundary between an explanatory and functional model. In designing Analogator, I completely ignored the goal of explaining analogy at the level of easy-to-understand representations. For this reason, one might consider Analogator to be one of a growing number of “radical connectionist” models (Cummins and Schwarz, 1987; Dorfner, 1990; Peschl, 1991).

That is not to say that Analogator does not help to explain analogy-making, but it does so at a level other than of the traditional analogical concepts (i.e., mappings, objects, relations, etc.) By breaking with tradition, it de-emphasizes concepts and representations – those ideas that Hofstadter claims that we need to be focusing on in the field of computational analogy-making (Hofstadter *et al.*, 1995). I do not disagree. However, I believe that we can not *directly* program concepts and representations; the representations that we create by hand are far too rigid. It would be like trying to construct clouds with hammers and nails. We need to develop the tools that allow us to *indirectly* build representations. I believe that we can accomplish this through learning systems; this is the level at which Analogator helps explain analogy-making.

One might think that I believe that the best method for modeling human cognition is one big associative network, starting *tabula raza*, which learns everything. Charniak and McDermott, in their Introduction to Artificial Intelligence, put it this way:

> One idea that has fascinated the Western mind is that there is a general-purpose learning mechanism that accounts for almost all of the state of an adult human being. According to this idea, people are born knowing very little, and absorb almost everything by way of this general learner. (Even a concept like "physical object," it has been proposed, is acquired by noticing that certain visual and tactile sensations come in stable bundles.) This idea is still powerfully attractive. It underlies much of behavioristic psychology. AI students often rediscover it, and propose to dispense with the study of reasoning and problem solving, and instead build a baby and let it just learn these things. We believe that this idea is dead, killed off by research in AI (and linguistics, and other branches of "cognitive science"). What this research has revealed is that for an organism to learn anything, it must already know a lot.
>
> (Charniak and McDermott, 1985 pp. 609-610)

This belief may have been more dead in 1985 when they wrote the above passage (which, in their defense, was right before the resurgence in connectionism); however, one may feel that that there is a bit of this "let it learn everything" flavor to Analogator. Certainly I do not think that one large associating network is the whole story. But, I must admit, I do agree with some of the philosophy of the behavioral psychologists. After the fall of behaviorism, cognitive scientists could not only assume the existence of internal representations, but could also insert them into their models wherever they felt necessary with little or no justification. I believe that researchers should look at a high-level cognitive ability, such as analogy-making, from a purely behavioristic perspective before making any assumptions regarding proposed internal representations.

Creating systems capable of learning abstract cognitive abilities, such as analogy-making, is currently a very difficult goal to reach for many reasons. For instance, learning such a task imposes additional constraints that force researchers to frame a problem in a *learnable* way. In addition, cognitive modelers are typically only peripherally interested in explaining how a model could come to be; they are more interested in the abilities of a model as an end product. However, learning is an interesting topic in and of itself, and will hopefully give us the appropriate tools with which to build flexible representations.

## 7.1 Analogator Limitations

Currently, Analogator suffers from a few major limitations. Probably the most serious is its lack of any mechanism of *attention*. Because of this, Analogator has problems with certain types of generalization, especially those involving invariance based on position. This problem was evident in both of the experiments involving spatial relations. Having some form of attention would, I believe, allow Analogator to scale-up to larger problems, such as being able to view scenes larger than 7 x 7 pixels.

Of serious concern is that fact that Analogator only operates in “pop-out mode”, never pondering a problem for more than a single propagation. In other words, analogies immediately come to Analogator’s “mind”, or not at all; it never *reasons* about a problem.

One possible extension to the basic Analogator framework could include a method of controlling its flow of recurrent activations by an attentional mechanism. I believe that Analogator could be adapted so that it could self-control recurrent activation, so that it could “explore” problems more thoroughly, and could solve more logic-based analogical reasoning problems (see Rumelhart, 1989, for other future visions of such a system). Eventually, problems that initially required sequential reasoning could be “compiled” into operating in the pop-out mode. This type of recurrent self-control could also be used to handle more complex, hierarchical problems that are currently out of reach of Analogator’s simple associative network. Recursively attending to objects (or groups of objects) might possibly give Analogator the power and flexibility to address other cognitive abilities, as suggested by Hinton's discussion of representing part-whole hierarchies in connectionist networks (1988).

I believe that a more complex architecture is needed to hardwire in certain other invariances, such as object rotation. Currently, backprop is used to learn everything. Although this makes the model quite general, it has been shown to limit Analogator's abilities, which required "hints" to overcome. I believe that the basic architecture could be adapted in a manner similar to Fukushima *et al.*'s Neocognitron (1983).

Some might suggest that the large number of trials needed to allow Analogator to make analogies is a serious drawback of the model. However, this does not concern me. After all, humans are years old before they are capable of making less superficial analogies. On the other hand, Analogator does currently need a teacher to point out the right answer at every step. It would be desirable to have an unsupervised training phase, to partially eliminate this requirement. This might be accomplished by hardwiring into the network certain "interesting" features in the world that seem to naturally to be the focus of attention (i.e., the figure).

Analogator's performance has not yet been correlated in any way with human analogy-making performance. A preliminary set of experiments was made with the help of Rob Goldstone and his lab. A set of analogy problems similar to those from Analogator's geometric-spatial domain was presented to college-age students, and choices and response-times were recorded. The subjects did show a preference for making analogies, even in situations that did not require, or even suggest, it. These initial experiments showed that humans naturally made analogies, but the exact analogies made with geometric shapes were affected by low-level criteria such as objects encountered first in sequentially scanning a scene, and objects that were brightly colored. Such criteria were, of course, not built into Analogator. Currently, the only way to correlate Analogator with this type of performance would be to train it to respond in such a manner; however, that would be a superficial method of gaining correlation, even if it could learn all of the subtleties. For proper correlation the model should incorporate sequential scanning, as well as many other low-level abilities.

Recall that Analogator is completely deterministic; it does not involve any randomness, or exploration of a problem. Because of that, it produces only a single answer for each problem, unlike Tabletop. Tabletop produces many answers for a single problem, and the distribution can be compared to human performance (Hofstadter *et al.*, 1995). This is impossible for Analogator in its current form. In addition, the one answer that Analogator does provide is based on the details of how it was trained. In short, Analogator is not quite ready for a fine-grained comparison with humans.

## 7.2 Contributions of this research

The contributions of this dissertation were made in two categories: the creation of a representation, and the creation of a training process. In the first category, iconic representations were developed which allow relations between objects to be easily compared and contrasted to single objects. For instance, a circle of triangles can be represented in a way that allows easy comparison to the representation of a single circle object. In this manner, the iconic representation creates a single, smooth landscape with which to represent objects, relations between objects, and object attributes.

A new training process was designed called the recurrent figure-ground associating procedure. This process associates the figure-ground components of source-target pairs in a connectionist network. As an associating process, it has much in common with other low-level perception models, and thus provides a unification with recognition and categorization. The training method also provides a unique means for “symbol grounding" by forcing hidden patterns to share the same activation space with retinotopic representations. Of central importance, the associating method was shown to be able to learn to make analogies in widely differing domains. In addition, the methodology was shown to train much more quickly than comparative feed-forward networks. Most importantly, the networks trained in this style exhibited good generalization ability by being able to make intra- and cross-domain analogies.

# Appendix A:

# Resources

All experiments in this dissertation were performed using the Con-x backprop simulator, which I wrote. There are two versions: a slow, fully programmable, Scheme version, and a fast, less flexible C++ version. Both are available via the Internet at:

`http://www.cs.indiana.edu/hyplan/blank/thesis/`

There you will also find code to create the networks and the datasets to duplicate the exact experiments from this dissertation. A sample Con-x script is shown below.

```
/* *************************************
   gridfont.cx: Con-x BackProp Simulator
   *************************************
*/

set   session e01     /* set DAT, ERR, WTS, OUT filenames     */
layer input   153
layer context 153
layer hidden  153
layer output  306
```

```
connect input hidden      /* connect input layer to hidden      */
connect context hidden    /* connect context layer to hidden     */
connect hidden output     /* connect hidden layer to output      */

set winner_take_all 0     /* turn winner take all off         */
set stoperr     0.99   /* stop at correct         */
set momentum    0.9    /* set weight momentum       */
set epsilon     0.1    /* set learning rate        */
set bepsilon    0.1    /* set learning rate of biases   */
set maxepoch   1000
set reportrate   1
set maxrand     0.003
set round_off    1     /* changes round of display only   */

set subcycle 2

copyprior context 0 A file 153 I 153
copyprior output 0 T file 153 I 306

copyafter input  0 A file 459 I 153
copyafter context 0 A hidden 0 A 153
copyafter output 0 T file 612 I 306
```

# Appendix B:

# Letter-part Analogies Dataset

This appendix is the complete dataset of annotated a's as used in Experiment #1. These are directly based on those used in McGraw (1995) have been used with permission. Legend: I = brim, O = body, B = both.

```
.........    .........    .........    .........    .........    .........    .........
.........    .........    .........    .........    .........    .........    .........
.........    .........    .........    .........    .........    .........    .........
.........    .........    .........    .........    .........    .........    .........
.........    OOOOOOOOO    ....IIIIB    ....IIIIB    IIIIIIIII    IIIII....    IIIII....
.........    ........O    ...I...OO    .......OO    ........I    .....I...    .....I...
.........    ........O    ..I...O.O    ......O.O    ........I    ......I..    ......I..
.........    ........O    .I...O..O    .....O..O    ........I    .......I.    .......I.
....IIIIB    I.......B    I...O...O    OOOOO...O    OOOOO...I    OOOOO...I    OOOOO...I
.......OO    I.......I    ...O....O    O......O.    O...O...I    O...O...I    O....O..I
......O.O    I.......I    ..O.....O    O.....O..    O...O...I    O...O...I    O.....O.I
.....O..O    I.......I    .O......O    O....O...    O...O...I    O...O...I    O......OI
....OOOOO    IIIIIIIII    OOOOOOOOO    OOOOO....    O...BIIII    O...BIIII    OOOOOOOOB
.........    .........    .........    .........    .........    .........    .........
.........    .........    .........    .........    .........    .........    .........
.........    .........    .........    .........    .........    .........    .........
.........    .........    .........    .........    .........    .........    .........
```

```
.........   .........   .........   .........   .........   .........   .........
.........   .........   .........   .........   .........   .........   .........
.........   .........   .........   .........   .........   .........   .........
.........   .........   .........   .........   .........   .........   .........
IIIII....   ........I   ....I....   I........   ....IIIIB   ....IIIII   ....I....
.....I...   .......II   ...II....   .I.......   .......OO   ...I....I   .....I...
......I..   ......I.I   ..I.I....   ..I......   ......O.O   ..I.....I   ......I..
.......I.   .....I..I   .I..I....   ...I.....   .....O..O   .I......I   .......I.
OOOOO...I   O...I...I   I...B....   OOOOB....   OOOOO...O   I...OOOOB   OOOOO...B
.O.....I.   OO......I   ...O.O...   .O...O...   ...O....O   ...O...OO   O.......O
..O...I..   O.O.....I   ..O...O..   ..O...O..   ..O.....O   ..O...O.O   O.......O
...O.I...   O..O....I   .O.....O.   ...O...O.   .O......O   .O...O..O   O.......O
....B....   OOOOBIIII   OOOOOOOOO   ....OOOOO   OOOOOOOOO   OOOOO...O   OOOOOOOOO
.........   .........   .........   .........   .........   .........   .........
.........   .........   .........   .........   .........   .........   .........
.........   .........   .........   .........   .........   .........   .........
.........   .........   .........   .........   .........   .........   .........
```

```
.........   .........   .........   .........   .........   .........   .........
.........   .........   .........   .........   .........   .........   .........
.........   .........   .........   .........   .........   .........   .........
.........   .........   .........   .........   .........   .........   .........
IIIII....   IIIIIIIIB   ....IIIII   IIIIIIIII   ........I   IIIIB....   ....IIIIB
.....I...   .......OO   ........I   .I......I   .......II   ....BO...   .......OO
......I..   ......O.O   ........I   ..I.....I   ......I.I   ....B.O..   ......O.O
.......I.   .....O..O   ........I   ...I....I   .....I..I   ....B..O.   .....O..O
OOOOOOOOB   ....O...O   ....O...I   OOOOOOOOB   ....I...B   OOOOB...O   ....O...O
O......O.   ...O....O   ...O.O..I   .O......O   ...I...OO   .O.....O.   .....O..O
O.....O..   ..O.....O   ..O...O.I   ..O.....O   ..I...O.O   ..O...O..   ......O.O
O....O...   .O......O   .O.....OI   ...O....O   .I...O..O   ...O.O...   .......OO
OOOOO....   OOOOOOOOO   OOOOOOOOB   ....OOOOO   I...OOOOO   ....O....   ........O
.........   .........   .........   .........   .........   .........   .........
.........   .........   .........   .........   .........   .........   .........
.........   .........   .........   .........   .........   .........   .........
.........   .........   .........   .........   .........   .........   .........
```

```
.........   .........   ........I   .........   .........   .........   .........
.........   .........   .......II   .........   .........   .........   .........
.........   .........   ......I.I   .........   .........   .........   .........
.........   .........   .....I..I   .........   .........   .........   .........
IIIIIIIII   IIIIIIIII   ....I...I   ....IIIII   O...IIIIB   IIIII....   ....I....
........I   .I......I   ...I....I   ...I...II   OO.....OO   .....I...   .....I...
........I   ..I.....I   ..I.....I   ..I...I.I   O.O...O.O   ......I..   ......I..
........I   ...I....I   .I......I   .I...I..I   O..O.O..O   .......I.   .......I.
OOOOO...I   OOOOO...B   I...OOOOB   I...B...I   O...O...O   OOOOBIIII   OOOOO...I
....O...I   O.......O   ....O...O   ...OO...I   .O.....O.   O..O.....   .O.....I.
....O...I   O.......O   ....O...O   ..O.O...I   ..O...O..   O.O......   ..O...I..
....O...I   O.......O   ....O...O   .O..O...I   ...O.O...   OO.......   ...O.I...
OOOOBIIII   OOOOOOOOO   ....OOOOO   OOOOO...I   ....O....   O........   ....B....
.........   .........   .........   .........   .........   .........   .........
.........   .........   .........   .........   .........   .........   .........
.........   .........   .........   .........   .........   .........   .........
.........   .........   .........   .........   .........   .........   .........
```

```
.........   .........   .........   .........   .........   ....I....   .........
.........   .........   .........   .........   .........   ...II....   .........
.........   .........   .........   .........   .........   ..I.I....   .........
.........   .........   .........   .........   .........   .I..I....   .........
IIIIIIIII   ....IIIII   IIIIIIIII   IIIIIIIII   IIIII....   I...B....   IIIIIIIII
.I......I   ...I....I   ........I   ........I   .....I...   ...OO....   .......I.
..I.....I   ..I.....I   ........I   ........I   ......I..   ..O.O....   ......I..
...I....I   .I......I   ........I   ........I   .......I.   .O..O....   .....I...
OOOOOOOOB   I.......B   OOOOO...I   OOOOO...I   OOOOO...I   O...O....   OOOOB...O
O.......O   .......OO   O.......I   .O..O...I   O..O...I.   O...O....   O..O.O..O
O.......O   ......O.O   O.......I   ..O.O...I   O.O...I..   O...O....   O.O...O.O
O.......O   .....O..O   O.......I   ...OO...I   OO...I...   O...O....   OO.....OO
OOOOOOOOO   ....OOOOO   OOOOO...I   ....BIIII   OOOOBOOOO   OOOOO....   O.......O
.........   .........   .........   .........   .........   .........   .........
.........   .........   .........   .........   .........   .........   .........
.........   .........   .........   .........   .........   .........   .........
.........   .........   .........   .........   .........   .........   .........
```

```
.........      .........      .........      .........      .........      .........      .........
.........      .........      .........      .........      .........      .........      .........
.........      .........      .........      .........      .........      .........      .........
.........      .........      .........      .........      .........      .........      .........
....I....      IIIIIIIII      ....I....      IIIII....      IIIIIIIII      IIIII....      ....I....
.....I...      .......I.      .....I...      .....I...      I.......I      .....I...      ...I.I...
......I..      ......I..      ......I..      ......I..      I.......I      ......I..      ..I...I..
.......I.      .....I...      .......I.      .......I.      I.......I      .......I.      .I.....I.
OOOOO...B      OOOOB....      OOOOO...I      OOOOO...I      I...O...I      OOOOO...I      I...OOOOB
.O......O      O..O.....      O....O..I      .O...O..I      ...O.O..I      O......I.      ...O....O
..O.....O      O.O......      O.....O.I      ..O...O.I      ..O...O.I      O.....I..      ..O.....O
...O....O      OO.......      O......OI      ...O...OI      .O.....OI      O....I...      .O......O
....OOOOO      O........      OOOOOOOOB      ....OOOOB      OOOOOOOOB      O...I....      OOOOOOOOO
.........      .........      .........      .........      .........      .........      .........
.........      .........      .........      .........      .........      .........      .........
.........      .........      .........      .........      .........      .........      .........
.........      .........      .........      .........      .........      .........      .........
```

```
.........      .........      .........      .........      .........      .........      .........
.........      .........      .........      .........      .........      .........      .........
.........      .........      .........      .........      .........      .........      .........
.........      .........      .........      .........      .........      .........      .........
....IIIII      ....I....      I........      IIIIBOOOO      IIIIIIIII      ....I....      IIIIB....
........I      .....I...      II.......      ....O...O      I.......I      .....I...      ....OO...
........I      ......I..      I.I......      ....O...O      I.......I      ......I..      ....O.O..
........I      .......I.      I..I.....      ....O...O      I.......I      .......I.      ....O..O.
OOOOBIIII      ....O...I      I...B....      OOOOO...O      I...O...B      OOOOOOOOB      ....O...O
.O..O....      ...O.O..I      ...OO....      O.......O      ...O....O      .O..O....      ...O...OO
..O.O....      ..O...O.I      ..O.O....      O.......O      ..O.....O      ..O.O....      ..O...O.O
...OO....      .O.....OI      .O..O....      O.......O      .O......O      ...OO....      .O...O..O
....O....      OOOOOOOOB      OOOOO....      OOOOOOOOO      OOOOOOOOO      ....O....      OOOOO...O
.........      .........      .........      .........      .........      .........      .........
.........      .........      .........      .........      .........      .........      .........
.........      .........      .........      .........      .........      .........      .........
.........      .........      .........      .........      .........      .........      .........
```

```
.........      ....I....      .........      .........      IIIIIIIII      .........      ....I....
.........      ...I.....      .........      .........      ........I      .........      ...I.I...
.........      ..I......      .........      .........      ........I      .........      ..I...I..
.........      .I.......      .........      .........      ........I      .........      .I.....I.
IIIIIIIII      IIIII....      ....IIIIB      IIIIIIIII      OOOOOOOOB      IIIIIIIII      IIIII...I
.......I.      .....I...      ...I...OO      ........I      O.......O      ........I      .....I...
......I..      ......I..      ..I...O.O      ........I      O.......O      ........I      ......I..
.....I...      .......I.      .I...O..O      ........I      O.......O      ........I      .......I.
OOOOB...O      OOOOOOOOB      I...O...O      OOOOO...I      OOOOOOOOO      OOOOO...I      OOOOOOOOB
O....O..O      .O.....OO      ...O...OO      O..O....I      O.......O      .O...O..I      .O.....OO
O.....O.O      ..O...O.O      ..O...O.O      O.O.....I      O.......O      ..O...O.I      ..O...O.O
O......OO      ...O.O..O      .O...O..O      OO......I      O.......O      ...O...OI      ...O.O..O
OOOOOOOOO      ....O...O      OOOOO...O      BIIIIIIII      OOOOOOOOO      ....OOOOB      ....O...O
.........      .........      .........      .........      .........      .........      .......O.
.........      .........      .........      .........      .........      .........      ......O..
.........      .........      .........      .........      .........      .........      .....O...
.........      .........      .........      .........      .........      .........      ....O....
```

```
.........      .........      .........      .........      .........      .........      .........
.........      .........      .........      .........      .........      .........      .........
.........      .........      .........      .........      .........      .........      .........
.........      .........      .........      .........      .........      .........      .........
IIIIB....      IIIIIIIII      IIIIIIIIB      IIIII....      ....IIIIB      I........      IIIII....
....OO...      ........I      .......OO      .....I...      .......OO      .I.......      .....I...
....O.O..      ........I      ......O.O      ......I..      ......O.O      ..I......      ......I..
....O..O.      ........I      .....O..O      .......I.      .....O..O      ...I.....      .......I.
....O...O      OOOOO...I      OOOOO...O      ....O...B      ....O...O      ....B....      OOOOOOOOB
...O...O.      O......I.      .O......O      ...O...O.      ...O....O      ...O.O...      .O.....OO
..O...O..      O.....I..      ..O.....O      ..O...O..      ..O.....O      ..O...O..      ..O...O.O
.O...O...      O....I...      ...O....O      .O...O...      .O......O      .O.....O.      ...O.O..O
OOOOO....      OOOOB....      ....OOOOO      OOOOOOOOO      OOOOOOOOO      OOOOO...O      ....O...O
.........      .........      .........      .........      .........      .........      .........
.........      .........      .........      .........      .........      .........      .........
.........      .........      .........      .........      .........      .........      .........
.........      .........      .........      .........      .........      .........      .........
```

```
.........    .........    .........    .........    .........    .........    .........
.........    .........    .........    .........    .........    .........    .........
.........    .........    .........    .........    .........    .........    .........
.........    .........    .........    .........    .........    .........    .........
IIIIIIIII    ....I....    ....I....    I........    ....BOOOO    I........    IIIIIIIII
........I    .....I...    ...I.I...    II.......    ...IO...O    II.......    ........I
........I    ......I..    ..I...I..    I.I......    ..I.O...O    I.I......    ........I
........I    .......I.    .I.....I.    I..I.....    .I..O...O    I..I.....    ........I
....O...B    OOOOO...I    I...O...I    I...B....    I...O...O    I...BOOOO    OOOOOOOOB
...O....O    O..O...I.    ...O.O.I.    ...O.O...    ...O...OO    ...O...O.    .O......O
..O.....O    O.O...I..    ..O...B..    ..O...O..    ..O...O.O    ..O...O..    ..O.....O
.O......O    OO...I...    .O...O.O.    .O.....O.    .O...O..O    .O...O...    ...O....O
OOOOOOOOO    BIIII....    O...O...O    OOOOOOOOO    OOOOO...O    OOOOO....    ....OOOOO
.........    .........    .........    .........    .........    .........    .........
.........    .........    .........    .........    .........    .........    .........
.........    .........    .........    .........    .........    .........    .........
.........    .........    .........    .........    .........    .........    .........

.........    .........    .........    .........    .........    .........    .........
.........    .........    .........    .........    .........    .........    .........
.........    .........    .........    .........    .........    .........    .........
.........    .........    .........    .........    .........    .........    .........
I........    IIIII...O    IIIIB....    IIIII...I    IIIII....    ....IIIIB    ....I....
.I.......    ........O    .....O...    .....I.I.    .....I...    ...I...OO    ...I.I...
..I......    ........O    ......O..    ......I..    ......I..    ..I...O.O    ..I...I..
...I.....    ........O    .......O.    .....I.I.    .......I.    .I...O..O    .I.....I.
OOOOB....    O...OOOOO    OOOOO...O    OOOOB...I    ....OOOOB    I...O...O    I...O...B
O...OO...    O.......O    .O..O..O.    O....O.O.    ...O....O    ...O...O.    ....OO.OO
O...O.O..    O.......O    ..O.O.O..    O.....O..    ..O.....O    ..O...O..    ....O.O.O
O...O..O.    O.......O    ...OOO...    O....O.O.    .O......O    .O...O...    ....OO.OO
OOOOO...O    OOOOO...O    ....O....    OOOOO...O    OOOOO...O    OOOOO....    ....O...O
.........    .........    .........    .........    .........    .........    .........
.........    .........    .........    .........    .........    .........    .........
.........    .........    .........    .........    .........    .........    .........
.........    .........    .........    .........    .........    .........    .........

.........    .........    .........    .........    .........    .........    .........
.........    .........    .........    .........    .........    .........    .........
.........    .........    .........    .........    .........    .........    .........
.........    .........    .........    .........    .........    .........    .........
IIIII...O    IIIIB...O    ....IIIII    O...IIIII    ....I....    IIIII....    I........
........O    ...OO..O.    ........I    OO.....I.    ...II....    .....I...    II.......
........O    ..O.O.O..    ........I    O.O...I..    ..I.I....    ......I..    I.I......
........O    .O..OO...    ........I    O..O.I...    .I..I....    .......I.    I..I.....
OOOOOOOOO    O...O....    OOOOOOOOB    O...B....    I...B...O    OOOOO...B    I...B....
O.......O    O..OO....    .O......O    O..O.....    ...OO..O.    O..O...IO    ...OOO...
O.......O    O.O.O....    ..O.....O    O.O......    ..O.O.O..    O.O...I.O    ..O.O.O..
O.......O    OO..O....    ...O....O    OO.......    .O..OO...    OO...I..O    .O..O..O.
OOOOO...O    O...O....    ....OOOOO    O........    OOOOO....    OIIII...O    OOOOOOOOO
.........    .........    .........    .........    .........    .........    .........
.........    .........    .........    .........    .........    .........    .........
.........    .........    .........    .........    .........    .........    .........
.........    .........    .........    .........    .........    .........    .........

.........    .........    .........    .........    .........    .........    .........
.........    .........    .........    .........    .........    .........    .........
.........    .........    .........    .........    .........    .........    .........
.........    .........    .........    .........    .........    .........    .........
I........    IIIIIIIII    ....IIIII    IIIIB....    ....IIIII    IIIII....    IIIIIIIII
.I.......    ........I    ........I    ....OO...    ........I    ....I....    .......II
..I......    ........I    ........I    ....O.O..    ........I    ....I....    ......I.I
...I.....    ........I    ........I    ....O..O.    ........I    ....I....    .....I..I
O...I....    OOOOOOOOB    OOOOO...B    OOOOO...O    ........B    OOOOB....    O...I...I
OO...I...    O.......O    ........O    .O......O    .......OO    O....O...    OO...I..I
O.O...I..    O.......O    ........O    ..O.....O    ......O.O    O.....O..    O.O...I.I
O..O...I.    O.......O    ........O    ...O....O    .....O..O    O......O.    O..O...II
OOOOBIIII    OOOOO...O    OOOOOOOOO    ....OOOOO    ....OOOOO    OOOOOOOOO    OOOOBIIII
.........    .....O..O    .........    .........    .........    .........    .........
.........    ......O.O    .........    .........    .........    .........    .........
.........    .......OO    .........    .........    .........    .........    .........
.........    ........O    .........    .........    .........    .........    .........
```

```
.........   .........   .........   .........   .........   .........   .........
.........   .........   .........   .........   .........   .........   .........
.........   .........   .........   .........   .........   .........   .........
.........   .........   .........   .........   .........   .........   .........
....IIIII   IIIIIIIII   ....I...I   IIIIIIIII   IIIIIIIII   IIIIIIIII   ....IIIII
........I   ........I   ....I..II   ........I   .......I.   ........I   ........I
........I   ........I   ....I.I.I   ........I   ......I..   ........I   ........I
........I   ........I   ....II..I   ........I   .....I...   ........I   ........I
OOOOO...I   OOOOOOOOB   O...I...I   OOOOOOOOB   O...I....   OOOOOOOOB   OOOOO...I
O......I.   O......OO   OO......I   .O.....O.   OO...I...   O...O...O   .O...O.I.
O.....I..   O.....O.O   O.O.....I   ..O...O..   O.O...I..   O...O...O   ..O...B..
O....I...   O....O..O   O..O....I   ...O.O...   O..O...I.   O...O...O   ...O.O.O.
OOOOB....   OOOOO...O   OOOOBIIII   ....O....   OOOOBIIII   OOOOOOOOO   ....O...O
.........   .........   .........   .........   .........   .........   .........
.........   .........   .........   .........   .........   .........   .........
.........   .........   .........   .........   .........   .........   .........
.........   .........   .........   .........   .........   .........   .........
```

```
.........   ....IIIII   .........   .........   .........   .........   ....I...I
.........   ........I   .........   .........   .........   .........   ...I.I..I
.........   ........I   .........   .........   .........   .........   ..I...I.I
.........   ........I   .........   .........   .........   .........   .I.....II
IIIIIIIII   OOOOOOOOB   ....IIIII   I...IIIII   IIIIIIIII   I........   IIIII...I
.......I.   O.......O   ........I   I..I....I   .......I.   .I.......   .....I...
......I..   O.......O   ........I   I.I.....I   ......I..   ..I......   ......I..
.....I...   O.......O   ........I   II......I   .....I...   ...I.....   .......I.
OOOOB...O   O.......O   OOOOO...I   I...OOOOB   OOOOB...O   OOOOB....   OOOOOOOOB
.O...O.OO   O.......O   O....O...   ...O....O   O......O.   O..OO....   .O.....OO
..O...O.O   O.......O   O.....O..   ..O.....O   O.....O..   O.O.O....   ..O...O.O
...O.O.OO   O.......O   O......O.   .O......O   O....O...   OO..O....   ...O.O..O
....O...O   OOOOOOOOO   OOOOO...O   OOOOO...O   OOOOO....   O...OOOOO   ....O...O
.........   .........   .........   .........   .........   .........   .......O.
.........   .........   .........   .........   .........   .........   ......O..
.........   .........   .........   .........   .........   .........   .....O...
.........   .........   .........   .........   .........   .........   ....OOOOO
```

```
I........   .........   .........   .........   .........   ....I....   .........
I........   .........   .........   .........   .........   ...I.....   .........
I........   .........   .........   .........   .........   ..I......   .........
I........   .........   .........   .........   .........   .I.......   .........
IIIIIIIII   IIIIIIIII   O...IIIII   ....I....   I........   IIIIIIIIB   ....IIIII
I.......I   ........I   OO......I   ...I.I...   .I.......   .......OO   ........I
I.......I   ........I   O.O.....I   ..I...I..   ..I......   ......O.O   ........I
I.......I   ........I   O..O....I   .I.....I.   ...I.....   .....O..O   ........I
I...OOOOB   OOOOOOOOB   O...OOOOB   I...OOOOB   OOOOB...O   OOOOO...O   ....OOOOB
....O...O   O.......O   O.......O   ...O...O.   .O..OO..O   O......OO   ...O....O
....O...O   O.......O   O.......O   ..O...O..   ..O.O.O.O   O.....O.O   ..O.....O
....O...O   O.......O   O.......O   .O...O...   ...OO..OO   O....O..O   .O......O
....OOOOO   OOOOOOOOO   OOOOOOOOO   O...O....   ....O...O   OOOOO...O   OOOOOOOOO
.........   .........   .........   O..O.....   .........   .........   .........
.........   .........   .........   O.O......   .........   .........   .........
.........   .........   .........   OO.......   .........   .........   .........
.........   .........   .........   O........   .........   .........   .........
```

```
.........   .........   ....I....   .........   ....IIIII   .........   .........
.........   .........   ...I.I...   .........   ...I....I   .........   .........
.........   .........   ..I...I..   .........   ..I.....I   .........   .........
.........   .........   .I.....I.   .........   .I......I   .........   .........
IIIIIIIII   ....B....   I...O...B   ....IIIII   I.......B   IIIII...I   ....IIIII
........I   ...IOO...   ...OO..OO   ........I   .......OO   ....I..II   ...I.....
........I   ..I.O.O..   ..O.O.O.O   ........I   ......O.O   ....I.I.I   ..I......
........I   .I..O..O.   .O..OO..O   ........I   .....O..O   ....II..I   .I.......
OOOOOOOOB   I...O...O   O...O...O   OOOOOOOOB   OOOOO...O   O...I...I   I...OOOOO
OO.OO...O   ...O....O   .O.....O.   .O.....O.   O.......O   OO.....I.   ...O....O
O.O.O...O   ..O.....O   ..O...O..   ..O...O..   O.......O   O.O...I..   ..O.....O
OO.OO...O   .O......O   ...O.O...   ...O.O...   O.......O   O..O.I...   .O......O
OOOOOOOOO   OOOOOOOOO   ....O....   ....O....   O.......O   OOOOB....   OOOOOOOOO
.........   .........   .........   .........   .O.....OO   .........   .........
.........   .........   .........   .........   ..O...O.O   .........   .........
.........   .........   .........   .........   ...O.O..O   .........   .........
.........   .........   .........   .........   ....O...O   .........   .........
```

```
.........    .........    .........    .........    .........    .........    .........
.........    .........    .........    .........    .........    .........    .........
.........    .........    .........    .........    .........    .........    .........
.........    .........    .........    .........    .........    .........    .........
........I    ........O    IIIIIIIII    IIIIB....    IIIIIIIII    ....IIIII    .........
.......II    .......O.    ........I    ....OO...    ........I    ........I    .........
......I.I    ......O..    ........I    ....O.O..    ........I    ........I    .........
.....I..I    .....O...    ........I    ....O..O.    ........I    ........I    .........
O...I...I    IIIIB....    OOOOO...I    OOOOO...O    OOOOO...I    OOOOO...B    O...IIIII
OO.....I.    ...OO....    O...O...I    O......O.    O....O..I    O.......O    OO.....I.
O.O...I..    ..O.O....    O...O...I    O.....O..    O.....O.I    O.......O    O.O...I..
O..O.I...    .O..O....    O...O...I    O....O...    O......OI    O.......O    O..O.I...
OOOOBOOOO    OOOOOOOOO    OOOOBIIII    OOOOO....    OOOOO...B    O...OOOOO    OOOOB....
.........    .........    .........    .........    .........    .........    .........
.........    .........    .........    .........    .........    .........    .........
.........    .........    .........    .........    .........    .........    .........
.........    .........    .........    .........    .........    .........    .........
```

```
.........    .........    .........    .........    .........    .........    .........
.........    .........    .........    .........    .........    .........    .........
.........    .........    .........    .........    .........    .........    .........
.........    .........    .........    .........    .........    .........    .........
I........    ....IIIII    IIIIIIIII    IIIIIIIII    IIIIIIIII    ....I....    IIIIIIIII
.I.......    ........I    ........I    .......I.    ........I    ....II...    .......I.
..I......    ........I    ........I    ......I..    ........I    ....I.I..    ......I..
...I.....    ........I    ........I    .....I...    ........I    ....I..I.    .....I...
OOOOBOOOO    OOOOO...I    OOOOO...B    OOOOB....    OOOOO...I    O...I...I    OOOOB...O
O..O.....    .O..O..I.    O........    .O...O...    O...OO..I    OO.....I.    O....O..O
O.O......    ..O.O.I..    O........    ..O...O..    O...O.O.I    O.O...I..    O.....O.O
OO.......    ...OOI...    O........    ...O...O.    O...O..OI    O..O.I...    O......OO
O........    ....BOOOO    OOOOO....    ....OOOOO    OOOOO...B    OOOOB....    OOOOO...O
.........    ........O    .........    .........    .........    .........    .........
.........    ........O    .........    .........    .........    .........    .........
.........    ........O    .........    .........    .........    .........    .........
.........    ........O    .........    .........    .........    .........    .........
```

```
.........    .........    .........    .........    .........    .........    .........
.........    .........    .........    .........    .........    .........    .........
.........    .........    .........    .........    .........    .........    .........
.........    .........    .........    .........    .........    .........    .........
IIIIB....    IIIIIIIII    IIIII....    ....I...O    ....IIIII    IIIIIIIII    ....I...O
...OO....    .......I.    .....I...    ...I.I.O.    ........I    ........I    ...I.I.O.
..O.O....    ......I..    ......I..    ..I...B..    ........I    ........I    ..I...B..
.O..O....    .....I...    .......I.    .I...O.O.    ........I    ........I    .I...O.I.
O...O....    OOOOB....    ....OOOOB    I...O...O    ....OOOOB    OOOOO...B    I...O...B
.O..OO...    O....O...    ...O...O.    ...O...O.    ....O..OO    O.......O    ...O....O
..O.O.O..    O.....O..    ..O...O..    ..O...O..    ....O.O.O    O.......O    ..O.....O
...OO..O.    O......O.    .O...O...    .O...O...    ....OO..O    O.......O    .O......O
....O...O    OOOOO...O    OOOOO....    OOOOO....    ....O...O    OOOOOOOOO    OOOOOOOOO
.........    .........    .........    .........    .........    .........    .........
.........    .........    .........    .........    .........    .........    .........
.........    .........    .........    .........    .........    .........    .........
.........    .........    .........    .........    .........    .........    .........
```

```
.........    .........    .........    .........    .........    .........    .........
.........    .........    .........    .........    .........    .........    .........
.........    .........    .........    .........    .........    .........    .........
.........    .........    .........    .........    .........    .........    .........
IIIII....    IIIIIIIII    IIIIIIIII    ....I....    I...I...O    IIIIIIIII    IIIII...O
....I....    ........I    ........I    .....I...    I..I.I.O.    ........I    ....I..OO
....I....    ........I    ........I    ......I..    I.I...B..    ........I    ....I.O.O
....I....    ........I    ........I    .......I.    II...O.I.    ........I    ....IO..O
OOOOB...O    O.......I    OOOOBIIII    OOOOO...I    I...O...B    ....O...I    OOOOB...O
O..OO..O.    OO.....OB    O...O....    .O.......    ...O....O    ...O.O..I    O...O..O.
O.O.O.O..    O.O...O.O    O...O....    ..O......    ..O.....O    ..O...O.I    O...O.O..
OO..OO...    O..O.O..O    O...O....    ...O.....    .O......O    .O.....OI    O...OO...
O...O....    OOOOOOOOO    OOOOO....    ....O....    OOOOOOOOO    OOOOO...B    OOOOO....
.........    .........    .........    .........    .........    .........    .........
.........    .........    .........    .........    .........    .........    .........
.........    .........    .........    .........    .........    .........    .........
.........    .........    .........    .........    .........    .........    .........
```

```
.........   .........   I........   .........   .........   .........   I........
.........   .........   II.......   .........   .........   .........   II.......
.........   .........   I.I......   .........   .........   .........   I.I......
.........   .........   I..I.....   .........   .........   .........   I..I.....
....I....   ....BOOOO   I...B....   ....IIIII   ....IIIII   IIIIBOOOO   I...IIIII
...I.I...   ...IO...O   ...OO....   ........I   ........I   ...O....O   ........I
..I...I..   ..I.O...O   ..O.O....   ........I   ........I   ..O.....O   ........I
.I.....I.   .I..O...O   .O..O....   ........I   ........I   .O......O   ........I
I...O...I   I...O...O   O...O....   OOOOOOOOB   ....BIIII   O...O...O   OOOOOOOOB
...O.O..I   ...O....O   O..O.....   O.......O   ...O.O...   O..OO..OO   O.......O
..O...O.I   ..O.....O   O.O......   O.......O   ..O...O..   O.O.O.O.O   O.......O
.O.....OI   .O......O   OO.......   O.......O   .O.....O.   OO..OO..O   O.......O
OOOOOOOOB   OOOOOOOOO   O........   OOOOOOOOO   OOOOOOOOO   O...O...O   O...OOOOO
.........   .........   .........   .........   .........   .........   O..O.....
.........   .........   .........   .........   .........   .........   O.O......
.........   .........   .........   .........   .........   .........   OO.......
.........   .........   .........   .........   .........   .........   O........

.........   .........   .........   .........   .........   .........   .........
.........   .........   .........   .........   .........   .........   .........
.........   .........   .........   .........   .........   .........   .........
.........   .........   .........   .........   .........   .........   .........
IIIIBOOOO   OIIII...O   IIIII....   IIIII....   I........   ....IIIII   ....I....
....O...O   ....I..OO   .....I...   .....I...   II.......   ...I....I   ...I.I...
....O...O   ....I.O.O   ......I..   ......I..   I.I......   ..I.....I   ..I...I..
....O...O   ....IO..O   .......I.   .......I.   I..I.....   .I......I   .I.....I.
OOOOO...O   OOOOB...O   OOOOO...I   OOOOO...B   I...B...O   I...O...B   I...OOOOB
O...O...O   O...O..OO   .O.O...I.   .O.....OO   ...O.O..O   ...O....O   ....O..O.
O...O...O   O...O.O.O   ..B...I..   ..O...O.O   ..O...O.O   ..O.....O   ....O.O..
O...O...O   O...OO..O   .O.I.I...   ...O.O..O   .O.....OO   .O......O   ....OO...
OOOOOOOOO   OOOOO...O   O...I....   ....O...O   OOOOOOOOO   OOOOOOOOO   ....O....
.........   .........   .........   .........   .........   .........   .........
.........   .........   .........   .........   .........   .........   .........
.........   .........   .........   .........   .........   .........   .........
.........   .........   .........   .........   .........   .........   .........

.........   .........   .........   .........   .........   ....I....   .........
.........   .........   .........   .........   .........   ....I....   .........
.........   .........   .........   .........   .........   ....I....   .........
.........   .........   .........   .........   .........   ....I....   .........
IIIII....   IIIIIIIII   ....I....   IIIII....   IIIIIIIII   ....IIIII   IIIIIIIII
.....I...   .I......I   ...II....   ....I....   ........I   ........I   ........I
......I..   ..I.....I   ..I.I....   ....I....   ........I   ........I   ........I
.......I.   ...I....I   .I..I....   ....I....   ........I   ........I   ........I
OOOOO...I   ....I...B   I...BOOOO   OOOOBOOOO   OOOOBIIII   OOOOO...B   OOOOBIIII
O....O.I.   .......OO   ...O...O.   O...O...O   O..O.....   O...O...O   O...OO...
O.....B..   ......O.O   ..O...O..   O...O...O   O.O......   O...O...O   O...O.O..
O....I.O.   .....O..O   .O...O...   O...O...O   OO.......   O...O...O   O...O..O.
OOOOB...O   ....OOOOO   OOOOO....   OOOOO...O   O........   OOOOOOOOO   OOOOO...O
.........   .........   .........   .........   .........   ....O....   .........
.........   .........   .........   .........   .........   ....O....   .........
.........   .........   .........   .........   .........   ....O....   .........
.........   .........   .........   .........   .........   ....O....   .........

.........   .........   .........   .........   .........   .........   .........
.........   .........   .........   .........   .........   .........   .........
.........   .........   .........   .........   .........   .........   .........
.........   .........   .........   .........   .........   .........   .........
....B....   IIIIIIIII   IIIII....   IIIIIIIII   .........   IIIIIIIII   IIIIIIIIB
...IOO...   .......I.   .....I...   ........I   .........   .......I.   .......IO
..I.O.O..   ......I..   ......I..   ........I   .........   ......I..   ......I.O
.I..O..O.   .....I...   .......I.   ........I   .........   .....I...   .....I..O
I...O...O   OOOOBOOOO   OOOOOOOOB   O...OOOOB   IIIIBOOOO   OOOOBOOOO   OOOOO...O
...O...OO   .O.....O.   O..O.....   O...O...O   ...O....O   O......O.   .O..O...O
..O...O.O   ..O...O..   O.O......   O...O...O   ..O.....O   O.....O..   ..O.O...O
.O...O..O   ...O.O...   OO.......   O...O...O   .O......O   O....O...   ...OO...O
OOOOO...O   ....O....   OOOOOOOOO   OOOOO...O   OOOOOOOOO   OOOOOOOOO   ....O...O
.........   .........   .........   .........   .........   .........   .........
.........   .........   .........   .........   .........   .........   .........
.........   .........   .........   .........   .........   .........   .........
.........   .........   .........   .........   .........   .........   .........
```

```
.........    .........    .........    .........    .........    .........    .........
.........    .........    .........    .........    .........    .........    .........
.........    .........    .........    .........    .........    .........    .........
.........    .........    .........    .........    .........    .........    .........
I........    ....IIIIB    ....IIIII    IIIIIIIIB    IIIII....    ....I....    IIIIIIIII
.I.......    ...I...OO    ...I....I    .......IO    ....I....    .....I...    ........I
..I......    ..I...O.O    ..I.....I    ......I.O    ....I....    ......I..    ........I
...I.....    .I...O..O    .I......I    .....I..O    ....I....    .......I.    ........I
OOOOB...O    I...O...O    I...OOOOB    OOOOO...O    OOOOBOOOO    ....OOOOB    OOOOOOOOB
.O...O.O.    .....O..O    ...O...O.    O...O...O    .O.....O.    ...O...O.    O.......O
..O...O..    ......O.O    ..O...O..    O...O...O    ..O...O..    ..O...O..    O.......O
...O.O.O.    .......OO    .O...O...    O...O...O    ...O.O...    .O...O...    O.......O
....O...O    ........O    OOOOO....    OOOOO...O    ....OOOOO    OOOOO....    O...O...O
.........    .........    .........    .........    .........    .........    .O..OO..O
.........    .........    .........    .........    .........    .........    ..O.O.O.O
.........    .........    .........    .........    .........    .........    ...OO..OO
.........    .........    .........    .........    .........    .........    ....O...O
```

```
.........    .........    ....IIIIB    .........    .........    .........    ....I....
.........    .........    ...I...OO    .........    .........    .........    ...I.I...
.........    .........    ..I...O.O    .........    .........    .........    ..I...I..
.........    .........    .I...O..O    .........    .........    .........    .I.....I.
I...I....    ....IIIIB    I...OOOOO    IIIIIIIIB    ........B    IIIII....    I...OOOOB
.I..II...    ...I...OO    .........    .......IO    .......IB    .....I...    ...O....O
..I.I.I..    ..I...O.O    .........    ......I.O    ......I.B    ......I..    ..O.....O
...II..I.    .I...O..O    .........    .....I..O    .....I..B    .......I.    .O......O
O...I...B    I...O...O    .........    BIIII...O    O...I...B    ....OOOOB    O.......O
OO.....IO    ....O..O.    .........    OO......O    OO.....IO    ...OO..I.    .O......O
O.O...I.O    ....O.O..    .........    O.O.....O    O.O...I.O    ..O.O.I..    ..O.....O
O..O.I..O    ....OO...    .........    O..O....O    O..O.I..O    .O..OI...    ...O....O
OOOOB...O    ....O....    .........    OOOOO...O    OOOOO...O    OOOOB....    ....OOOOO
.........    .........    .........    .........    .........    .........    .........
.........    .........    .........    .........    .........    .........    .........
.........    .........    .........    .........    .........    .........    .........
.........    .........    .........    .........    .........    .........    .........
```

```
.........    .........    I........    .........    .........    .........    .........
.........    .........    .I.......    .........    .........    .........    .........
.........    .........    ..I......    .........    .........    .........    .........
.........    .........    ...I.....    .........    .........    .........    .........
IIIIIIIII    I........    O...I....    ....I....    ....I....    IIIIIIIII    O...IIIIB
........I    .I.......    OO..I....    .....I...    ...II....    .......I.    OO.....OO
........I    ..I......    O.O.I....    ......I..    ..I.I....    ......I..    O.O...O.O
........I    ...I.....    O..OI....    .......I.    .I..I....    .....I...    O..O.O..O
OOOOOOOOB    OOOOB....    O...B....    OOOOO...I    I...B...O    OOOOBIIII    O...O...O
.O..OO..O    O..O.O...    .O..OO...    .O..O..I.    ...O.O..O    .O..OO...    .O.....OO
..O.O.O.O    O.O...O..    ..O.O.O..    ..O.O.I..    ..O...O.O    ..O.O.O..    ..O...O.O
...OO..OO    OO.....O.    ...OO..O.    ...OOI...    .O.....OO    ...OO..O.    ...O.O..O
....O...O    O...OOOOO    ....O...O    ....B....    OOOOOOOOO    ....O...O    ....O...O
.........    .........    .........    .........    .........    .........    .........
.........    .........    .........    .........    .........    .........    .........
.........    .........    .........    .........    .........    .........    .........
.........    .........    .........    .........    .........    .........    .........
```

```
....I....    I........    .........    .........    .........    .........    .........
...I.....    .I.......    .........    .........    .........    .........    .........
..I......    ..I......    .........    .........    .........    .........    .........
.I.......    ...I.....    .........    .........    .........    .........    .........
IIIIIIIII    O...B....    IIIII....    I........    ....IIIII    IIIIIIIII    .........
........I    OO..OO...    ....I....    .I.......    ........I    ........I    .........
........I    O.O.O.O..    ....I....    ..I......    ........I    ........I    .........
........I    O..OO..O.    ....I....    ...I.....    ........I    ........I    .........
OOOOOOOOB    O...O...O    OOOOB....    OOOOBOOOO    OOOOOOOOB    OOOOOOOOB    .........
.O......O    .O..O..OO    .O..OO...    O.......O    ...O...O.    ....O...O    .........
..O.....O    ..O.O.O.O    ..O.O.O..    O.......O    ..O...O..    ....O...O    .........
...O....O    ...OOO..O    ...OO..O.    O.......O    .O...O...    ....O...O    .........
....O...O    ....O...O    ....O...O    OOOOOOOOO    OOOOOOOOO    OOOOOOOOO    IIIIBOOOO
.........    .........    .........    .........    .........    .........    ...O...O.
.........    .........    .........    .........    .........    .........    ..O...O..
.........    .........    .........    .........    .........    .........    .O...O...
.........    .........    .........    .........    .........    .........    OOOOO....
```

```
.........    .........    .........    .........    .........    .........    .........
.........    .........    .........    .........    .........    .........    .........
.........    .........    .........    .........    .........    .........    .........
.........    .........    .........    .........    .........    .........    .........
....BOOOO    ....I....    IIIII....    ....IIIIB    ....IIIII    ....IIIII    IIIII....
...IO...O    ...I.I...    .....I...    ...I...OO    ...I...I.    ........I    ....I....
..I.O...O    ..I...I..    ......I..    ..I...O.O    ..I...I..    ........I    ....I....
.I..O...O    .I.....I.    .......I.    .I...O..O    .I...I...    ........I    ....I....
I...O...O    I...O...B    OOOOO...I    I...O...O    I...I...O    OOOOO...I    OOOOB....
...O...O.    ...O...O.    O..O...I.    ....O..OO    ...OO..O.    ....O...I    O...O....
..O...O..    ..O...O..    O.O...I..    ....O.O.O    ..O.O.O..    ....O...I    O...O....
.O...O...    .O...O...    OO...I...    ....OO..O    .O..OO...    ....O...I    O...O....
OOOOO....    OOOOO....    O...I....    ....O...O    OOOOO....    OOOOOIIII    OOOOO....
.........    .........    .........    .........    .........    .........    .........
.........    .........    .........    .........    .........    .........    .........
.........    .........    .........    .........    .........    .........    .........
.........    .........    .........    .........    .........    .........    .........
```

```
.........    .........    .........    .........    .........    .........    .........
.........    .........    .........    .........    .........    .........    .........
.........    .........    .........    .........    .........    .........    .........
.........    .........    .........    .........    .........    .........    .........
.........    IIIII....    IIIII....    I...IIIII    IIIIIIIII    ....IIIII    IIIII....
.........    .....I...    ....IO...    I..I....I    ........I    ...I....I    .....I...
.........    ......I..    ....I.O..    I.I.....I    ........I    ..I.....I    ......I..
.........    .......I.    ....I..O.    II......I    ........I    .I......I    .......I.
IIIIB....    OOOOO...I    OOOOB...O    I...OOOOB    OOOOO...I    I...OOOOB    OOOOO...I
...OO....    O......I.    O...O..O.    .....O..O    O...O...I    ...O....O    O......I.
..O.O....    O.....I..    O...O.O..    ......O.O    O...O...I    ..O.....O    O.....I..
.O..O....    O....I...    O...OO...    .......OO    O...O...I    .O......O    O....I...
OOOOOOOOO    OOOOB....    OOOOO....    ........O    OOOOO...I    OOOOOOOOO    OOOOBOOOO
.........    .........    .........    .........    .........    .........    .........
.........    .........    .........    .........    .........    .........    .........
.........    .........    .........    .........    .........    .........    .........
.........    .........    .........    .........    .........    .........    .........
```

```
.........    .........    .........    .........    .........    .........    .........
.........    .........    .........    .........    .........    .........    .........
.........    .........    .........    .........    .........    .........    .........
.........    .........    .........    .........    .........    .........    .........
....IIIII    IIIII...O    IIIIIIIBB    ....I....    ....IIIII    ....I....    ....IIIII
.......I.    ...I...OO    .......OO    .....I...    ........I    .....I...    ........I
......I..    ..I...O.O    ......O.O    ......I..    ........I    ......I..    ........I
.....I...    .I...O..O    .....O..O    .......I.    ........I    .......I.    ........I
OOOOB...O    IIIIB...O    OOOOO...O    OOOOOOOOB    OOOOO...I    OOOOOOOOB    OOOOO...I
.O.....OO    ...O....O    .O.....OO    .O......O    O...O...I    .O.....O.    O.......I
..O...O.O    ..O.....O    ..O...O.O    ..O.....O    O...O...I    ..O...O..    O.......I
...O.O..O    .O......O    ...O.O..O    ...O....O    O...O...I    ...O.O...    O.......I
....OOOOO    OOOOOOOOO    ....O...O    ....OOOOO    OOOOBIIII    ....O....    OOOOOOOOB
.........    .........    .........    .........    .........    .........    .........
.........    .........    .........    .........    .........    .........    .........
.........    .........    .........    .........    .........    .........    .........
.........    .........    .........    .........    .........    .........    .........
```

```
IIIIB....    .........    .........    .........    .........
....OO...    .........    .........    .........    .........
....O.O..    .........    .........    .........    .........
....O..O.    .........    .........    .........    .........
O...O...O    IIIII....    ....IIIII    ....I....    ....I...O
OO..O...O    .....I...    ...I....I    .....I...    ...I.I.O.
O.O.O...O    ......I..    ..I.....I    ......I..    ..I...B..
O..OO...O    .......I.    .I......I    .......I.    .I...O.I.
O...O...O    O...BIIII    I...O...B    OOOOO...I    I...O...I
O......OO    OO...O...    ...O...O.    .O...O..I    ...OOO...
O.....O.O    O.O...O..    ..O...O..    ..O...O.I    ..O.O.O..
O....O..O    O..O...O.    .O...O...    ...O...OI    .O..O..O.
OOOOO...O    OOOOO...O    OOOOO....    ....OOOOB    OOOOO...O
.........    .........    .........    .........    .........
.........    .........    .........    .........    .........
.........    .........    .........    .........    .........
.........    .........    .........    .........    .........
```

# Appendix C:

# Family Tree Relations

The following lists the relations used in the final family tree experiment. All six families were isomorphic.

11 Relations:

wife, husband, son, daughter, mother, father, brother, sister, parent-in-law, child-in-law, and sibling-in-law.

6 People:

Bello, Matope, Neema, Hadiya, Kissa, and Kwaku.

22 facts:

Bello is the son of Matope.
Bello is the son of Neema.
Hadiya is the daughter of Matope.
Hadiya is the daughter of Neema.
Matope is the father of Bello.
Matope is the father of Hadiya.
Neema is the mother of Bello.

Neema is the mother of Hadiya.
Bello is the brother of Hadiya.
Hadiya is the sister of Bello.
Kissa is the wife of Bello.
Neema is the wife of Matope.
Matope is the husband of Neema.
Bello is the husband of Kissa.
Hadiya is the wife of Kwaku.
Kwaku is the husband of Hadiya.
Matope is the parent-in-law of Kissa.
Kissa is the child-in-law of Matope.
Neema is the parent-in-law of Kwaku.
Kwaku is the child-in-law of Neema.
Kissa is the sibling-in-law of Kwaku.
Kwaku is the sibling-in-law of Kissa.